\documentclass[runningheads]{llncs}
\usepackage{graphicx}
\usepackage{comment}
\usepackage{amsmath,amssymb} %
\usepackage{color}
\usepackage{cite}
\usepackage{tabularx}
\usepackage{nicefrac}
\usepackage{multirow}
\usepackage{subcaption}
\usepackage{caption}
\usepackage{xcolor,colortbl}
\usepackage{booktabs}

\usepackage{dsfont}
\usepackage[pdftex,colorlinks=true,pdfstartview=FitV,linkcolor=black,citecolor=black,urlcolor=black,bookmarks=false]{hyperref}

\usepackage[width=122mm,left=12mm,paperwidth=146mm,height=193mm,top=12mm,paperheight=217mm]{geometry}

\definecolor{Yellow}{rgb}{1,1, 0.6}
\definecolor{Red}{rgb}{1, 0.6, 0.6}
\definecolor{Blue1}{rgb}{0, 0.6, 1}
\definecolor{green}{rgb}{0.2, 0.8, 0.2}
\definecolor{Red1}{rgb}{0.5, 0.2, 0.2}
\definecolor{PaleYellow}{rgb}{0.8,0.8, 0}
\definecolor{Purple}{rgb}{1.0, 0, 1.0}

\newcommand{\algoname}[0]{Du\textsuperscript{2}Net}
\newcommand{\etal}{et al.\ }

\hypersetup{final}

\begin{document}
\pagestyle{headings}
\mainmatter
\def\ECCVSubNumber{2263}  %

\title{\algoname: Learning Depth Estimation from Dual-Cameras and Dual-Pixels} %

\titlerunning{\algoname: Learning Depth Estimation from Dual-Cameras and Dual-Pixels}
\author{Yinda Zhang \and 
Neal Wadhwa \and
Sergio Orts-Escolano \and
Christian H{\"a}ne \and \\
Sean Fanello \and
Rahul Garg}
\authorrunning{Y. Zhang et al.}
\institute{Google Research}
\maketitle
\newcommand{\teasersubfigwidth}[0]{0.36\textwidth}
\begin{figure*}
\resizebox{\textwidth}{!}{
\begin{tabular}{ccccc}
\includegraphics[width=\teasersubfigwidth]{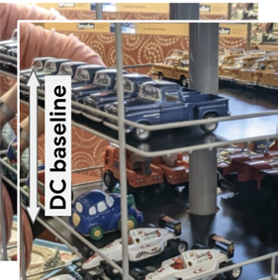} &
\includegraphics[width=\teasersubfigwidth]{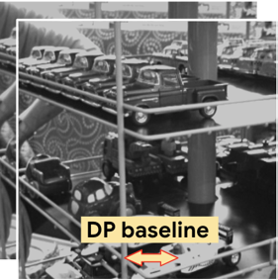} &
\includegraphics[width=\teasersubfigwidth]{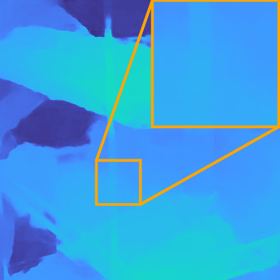} &
\includegraphics[width=\teasersubfigwidth]{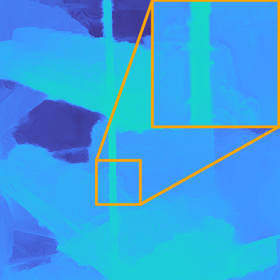} &
\includegraphics[width=\teasersubfigwidth]{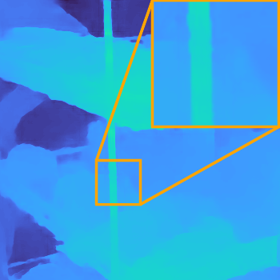} \\
\includegraphics[width=\teasersubfigwidth]{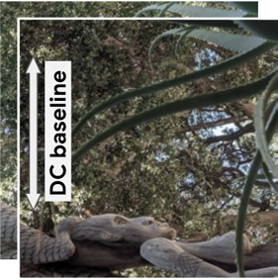} &
\includegraphics[width=\teasersubfigwidth]{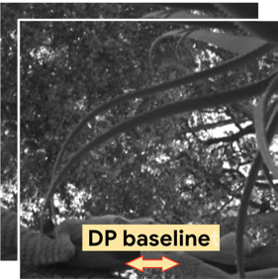} &
\includegraphics[width=\teasersubfigwidth]{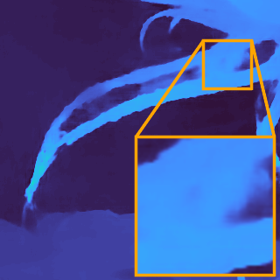} &
\includegraphics[width=\teasersubfigwidth]{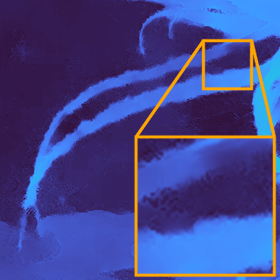} &
\includegraphics[width=\teasersubfigwidth]{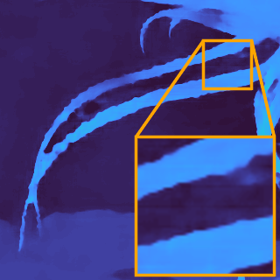} \\
(a) Dual-camera (DC) input & 
(b) Dual-pixel (DP) input & 
(c) Stereonet (DC input) \cite{khamis2018stereonet} & 
(d) DPNet (DP input) \cite{garg2019learning} & 
(e) \algoname~(Ours) \\
\end{tabular}}
    \centering
    \caption{\algoname \ combines dual-camera (DC) and dual-pixel (DP) images to produce edge-aware disparities with high precision even near occlusion boundaries. The large vertical DC baseline complements the small horizontal DP baseline to mitigate the aperture problem (top) and occlusions (bottom).}
    \label{fig:teaser}
    \vspace{-25pt}
\end{figure*}
\begin{abstract}
Computational stereo has reached a high level of accuracy, but
	degrades in the presence of occlusions, repeated textures, and
	correspondence errors along edges. 
We present a novel approach based
	on neural networks for depth estimation that combines stereo from dual
	cameras with stereo from a dual-pixel sensor, which is
	increasingly common on consumer cameras.  Our network uses a novel architecture to fuse these two sources of information and can overcome the above-mentioned limitations of pure binocular stereo matching.
Our method provides a dense depth map with sharp edges, which is crucial for computational photography applications like synthetic shallow-depth-of-field or 3D Photos.
Additionally, we avoid the inherent ambiguity due to the aperture problem in stereo cameras by designing the stereo baseline to be orthogonal to the dual-pixel baseline.
We present experiments and comparisons with state-of-the-art approaches to show that our method offers a substantial improvement over previous works.
\end{abstract}

\section{Introduction}
\label{sec:intro}
Despite their maturity, modern stereo depth estimation techniques still suffer from artifacts in occluded areas, around object boundaries and in regions containing edges parallel to the baseline (the so-called aperture problem). 
These errors are especially problematic for applications
requiring a depth map that is accurate near object boundaries, such as
synthetic shallow depth-of-field or 3D photos.

While these problems can be mitigated by using more than two cameras, a recent improvement to consumer camera sensors allows us to alleviate them without any extra hardware.
Specifically, camera manufacturers have added dual-pixel (DP) sensors to DSLR and smartphone cameras to assist with focusing. 
These sensors work by capturing two views of a scene through the camera's
single lens, thereby creating a tiny baseline binocular stereo pair  (Fig.~\ref{fig:dp_fig}).
Recent work has shown that it is possible to estimate depth from these dense dual-pixel sensors \cite{wadhwa2018,garg2019learning}.
Due to the tiny baseline, there are fewer occluded areas between the views, and as a result the depth from dual-pixels is more accurate near object boundaries than the depth from binocular stereo. However, the tiny baseline also means that the depth quality is worse than stereo at farther distances due to the quadratic increase in depth error in triangulation-based systems \cite{szeliski2010}.

In this work, we consider a dual-camera (DC) system where one camera has a dual-pixel sensor, a common setup on recently-released flagship smartphones. We propose a deep learning solution to estimate depth from both dual-pixels and dual-cameras. Because depth from dual-cameras and depth from dual-pixels have complementary errors, such a setup promises to have accurate depth at both near and far distances and around object boundaries. In addition, in our setup, the dual-pixel baseline is orthogonal to the dual-camera baseline.
This allows us to estimate depth even in regions where image texture is parallel to one of the two baselines (Fig.~\ref{fig:teaser}). This is usually difficult due to the well known aperture problem \cite{morgan1997aperture}.

One key problem that prevents the trivial solution of multi-view stereo matching from working is a fundamental affine ambiguity in the depth estimated from dual-pixels\cite{garg2019learning}. This is because disparity is related to inverse depth via an affine transformation that depends on the camera's focus distance, focal length and aperture size, which are often unknown or inaccurately recorded. 

To address this issue, we propose an end-to-end solution that uses two separate low resolution learned {\em confidence volumes}, i.e. the softmax of a negative cost volume, to compute disparity maps from dual-cameras and dual-pixels independently. We then fit an affine transformation between the two disparity maps and use it to resample the dual-pixels' \text{confidence volume}, so that it is in the same space as the dual-cameras' confidence volume. The two are then fused to estimate a low resolution disparity map. A final edge-aware refinement \cite{khamis2018stereonet} that leverages features computed from dual-pixels is then used to obtain the final high resolution disparity map.

To train and evaluate our approach, we capture a new dataset using a capture rig containing five synchronized Google Pixel 4 smartphones. Each phone has two cameras and each capture consists of ten RGB images (two per phone) and the corresponding dual-pixel data from one camera on each phone. We use multi-view stereo techniques to estimate ground truth disparity using all ten views.

Via extensive experiments and comparisons with state-of-the-art approaches, we show that our solution effectively leverages both dual-cameras and dual-pixels. We additionally show applications in computational photography where precise edges and dense disparity maps are the key for compelling results.

\section{Related Work}
\label{sec:related}
Stereo matching is a fundamental problem in computer vision and is often used in triangulation systems to estimate depth for various applications such as computational photography \cite{wadhwa2018}, autonomous driving \cite{Menze2015KITTI}, robotics \cite{fanello14}, augmented and virtual reality \cite{holoportation} and volumetric capture \cite{relightables}. Traditionally, stereo matching pipelines \cite{Scharstein2002} follow these main steps: matching cost computation, cost aggregation, and disparity optimization, often followed by a disparity refinement (post-processing) step. This problem has been studied for over four decades \cite{marr1976cooperative} and we refer the reader to \cite{Scharstein2002,Sinha2014,hamzah2016literature} for a survey of traditional techniques. 

Recent classical approaches aim at improving the disparity correspondence search by using either global \cite{besse2014pmbp,felzenszwalb2006efficient,klaus2006segment,kolmogorov2001computing} or local \cite{bleyer2011patchmatch,fanello17_hashmatch,sos} optimization schemes. These methods usually rely on hand-crafted descriptors \cite{bleyer2011patchmatch,sos} or learned shallow binary features \cite{fanello2017ultrastereo,fanello17_hashmatch} followed by sequential propagation steps \cite{bleyer2011patchmatch,fanello2017ultrastereo} or fast parallel approximated CRF inference \cite{sos,fanello17_hashmatch}. However, these methods cannot compete with recent deep learning-based methods that use end-to-end training \cite{kendall2017end,khamis2018stereonet,psmnet,EdgeStereo2018,iResNet2018,SegStereo2018,DeepPrunerICCV2019}. Such approaches were introduced by \cite{mayer2016large,ilg2017flownet}, who used encoder-decoder networks for the problems of disparity and flow estimation. 

Kendall et al.\ \cite{kendall2017end}, inspired by classical methods, employed a model architecture that constructs a full cost-volume with 3D convolutions as an intermediate stage and infers the final disparity through a soft-$\arg\min$ function. Khamis et al.\ \cite{khamis2018stereonet} extended this concept by using a learned edge aware refinement step as the final stage of the model to reduce computational cost. More recently, PSMNet \cite{psmnet} used a multi-scale pooling approach to improve the accuracy of the predicted disparities. Finally \cite{ganet}, inspired by \cite{hirschmuller2008stereo}, used a semi-global matching approach to replace the expensive 3D convolutions.

Other end-to-end approaches use multiple iterative refinements to converge to a final disparity solution. Gidaris et al.\ \cite{gidaris2017detect} propose a generic architecture for labeling problems, such as depth estimation, that is trained end-to-end to predict and refine the output. Pang et al.\ \cite{pang2017cascade} propose a cascaded approach to learn the depth residual from an initial estimate. Despite this progress, stereo depth estimation systems still suffer from limited precision in occlusion boundaries, imprecise edges, errors in areas with repeated textures, and the aperture problem. The aperture problem can be addressed with two orthogonal dual camera pairs \cite{meier2017real}. Occlusions can be reduced by using trinocular stereo \cite{mulligan2002trinocular}. Both of these approaches require additional hardware and more complex calibration. In this work, we combine bincocular stereo with a dual-pixel sensor, a hardware available in most modern smartphone and DSLR cameras where they are used for autofocus.

Recently, a handful of techniques have been proposed to recover depth from a single camera using dual-pixels \cite{wadhwa2018,garg2019learning}. Dual-pixels are essentially a two-view light field \cite{ng2005lightcamera}, providing two slightly different views of the scene. These two views can be approximated as a stereo pair except for a fundamental ambiguity identified by \cite{garg2019learning} discussed in the introduction.
In addition to depth estimation, dual-pixels have been used for dereflection \cite{Punnappurath_2019_CVPR}.

\section{Dual-Pixel Sensors}
\label{sec:dp}

\begin{figure}[t]
    \centering
    \includegraphics[width=\textwidth]{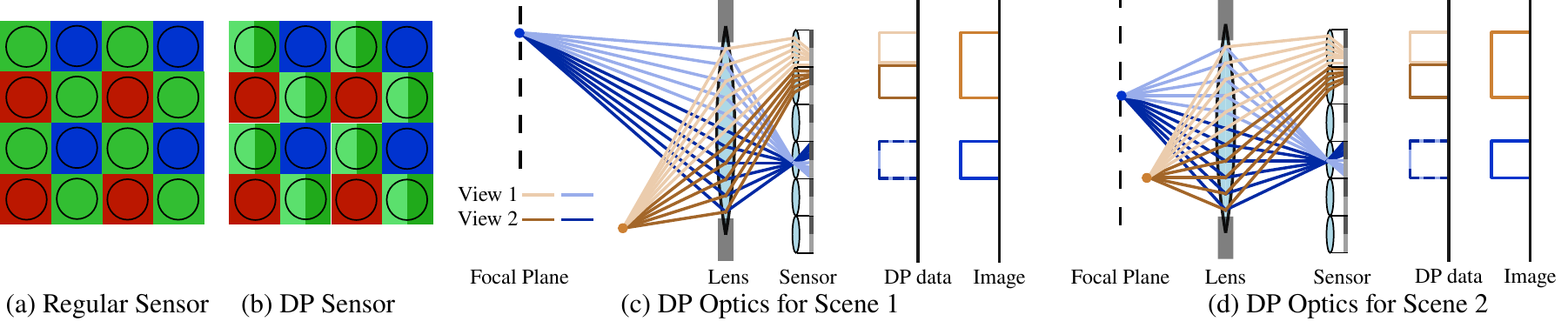}
    \caption{In a regular Bayer sensor, each pixel has a microlens on top to collect more light (a). Dual-pixel sensors split some of the pixels underneath the microlens into two halves; the green pixels in (b). The two dual-pixel views get their light from different halves of the aperture, resulting in a slight depth-dependent disparity between the views (c). Different scenes can produce the same dual-pixel images if the focus distance changes ((c) vs. (d)). This is a fundamental ambiguity of dual-pixel sensors. (Reproduced with permission from Garg \etal \cite{garg2019learning}.)}
    \label{fig:dp_fig}
    \vspace{-10pt}
\end{figure}

Dual-pixel sensors work by splitting each pixel in half, such that the left half integrates light over the right half of the aperture and the right half integrates light over the left half of the aperture. Because the two half pixels see light from different halves of the aperture, they form a kind of ``stereo pair'', whose centers of projection are in the centers of each half aperture. Since the two half pixel images account for all the light going through the aperture, when they are added together, the full normal image is recovered.

These sensors are becoming increasingly common in smartphone and DSLR cameras because they assist in auto-focus. The reason for this is that the zero-disparity distance corresponds exactly to the image being in-focus (e.g. the blue point in Fig.~\ref{fig:dp_fig}(c)) and disparity is exactly proportional to how much the lens needs to be moved to make the image in-focus. This property also implies that unlike rectified stereo image pairs, the range of disparities can be both negative and positive for DP data.

In addition to this dependence on focus distance, dual-pixels have a number of other key differences from stereo cameras. On the positive side, the dual-pixel views are perfectly synchronized and have the same white balance, exposure and focus, making matching easier. In addition, they are perfectly rectified. This means that the baseline is perfectly horizontal for a sensor whose dual-pixels are split horizontally as in Fig.~\ref{fig:dp_fig}(b). Another advantage of this perfect rectification is that, dual-pixel sensors are not affected by rolling shutter or optical image stabilization \cite{Diverdi2016}, which shifts the principal point and center of projection of a camera. While we need to calibrate for this with stereo cameras, it does not cause problems for dual-pixel images.

Like in a stereo pair, the small baseline of dual-pixel images means that depth estimation at large distances is difficult. Please see the supplementary material for a visualization of the tiny parallax between the DP images. However, it also means there are fewer occlusions and it is possible to get accurate depth near occlusion boundaries in the image. This suggests that a system that combines dual-cameras and dual-pixels could recover depth at short distances and in occluded areas from dual-pixels and depth at larger distances from dual-cameras.

Another difference between dual-pixels and traditional stereo cameras is the interaction between defocus and disparity. Specifically, the amount of defocus is exactly proportional to the disparity between the views. This means that a learned model that makes use of dual-pixels could make use of defocus as well to resolve ambiguities that typically fool matching-based approaches, such as repeated textures.

Finally, we elaborate on the affine ambiguity of depth predictions discussed in the introduction (Fig.~\ref{fig:dp_fig}(c-d)). This happens because the mapping between disparity and depth depends on focus distance which is often inaccurate or unknown in cheap smartphone camera modules. Garg \etal \cite{garg2019learning} used the paraxial and thin-lens approximation to show that $D_{DP}(x, y) = \alpha +\nicefrac{\beta}{Z(x, y)}$, where $Z(x, y)$ is the depth for pixel $x,y$; $D_{DP}(x,y)$ is the dual-pixel disparity at $(x, y)$, and $\alpha$ and $\beta$ are constants that depend on the aperture, point spread function, and the focus distance of the lens. Because these can be difficult to determine, inverse depth can be estimated only up to an unknown affine transform. 

If there is a second camera in addition to the camera with dual-pixels, such as in our setup, the stereo disparity $D_{DC}$ of the dual-cameras is $\nicefrac{bf}{Z}$ \cite{szeliski2010} where $b$ is the baseline and $f$ is the focal length. From this, it follows that $D_{DC}$ and $D_{DP}$ are also related via an affine transform 

\begin{equation}
D_{DC}(x, y) = \alpha' +\beta'D_{DP}(x, y)
\end{equation}

We use this observation in our network architecture to effectively integrate stereo and dual-pixel cues.

\section{Fusing Dual-Pixels and Dual-Cameras}
\label{sec:method}

\begin{figure}[t]
    \centering
    \includegraphics[width=\textwidth]{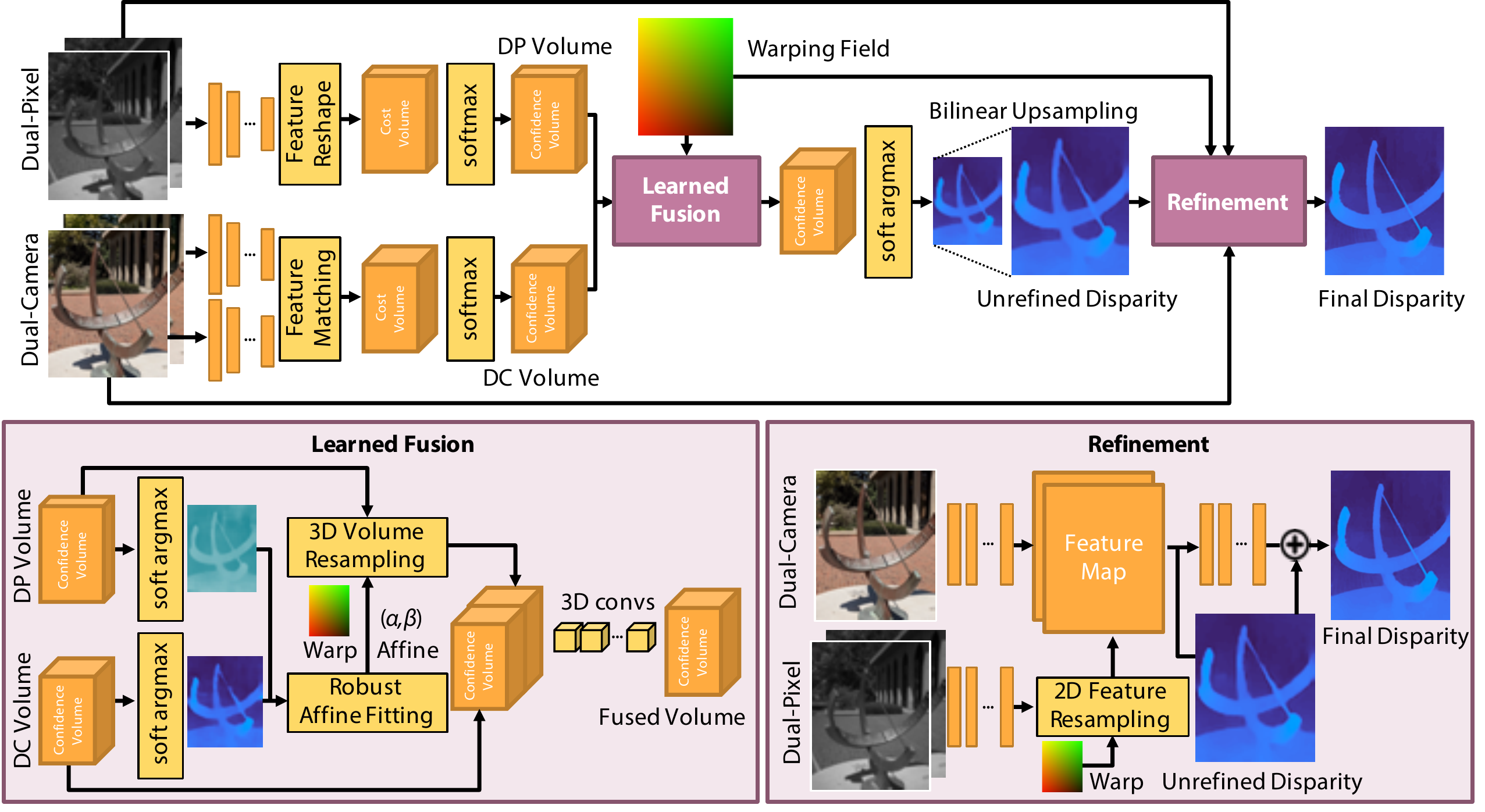}
    \caption{Overview of \algoname. Top: two disparity maps are separately inferred from the dual-camera and the dual-pixel branches. An affine transformation is fit between them and used to resample the dual-pixel confidence volume. It is then fused with the dual-camera volume and they are together used to infer the unrefined disparity. An edge-aware refinement step uses the dual-pixel features to predict the final disparity. Bottom: details of the volume fusion and refinement step. See text for more details.}
    \label{fig:arch}
    \vspace{-10pt}
\end{figure}

We describe our deep learning model to predict disparity from both dual-camera and dual-pixel data (Fig. \ref{fig:arch}).
The input to our system is a pair of rectified dual-camera (DC) images, $I_l$ and $I_r$ corresponding to the left and the right cameras, and a pair of dual-pixel (DP) images from the right camera sensor, $I^{DP}_{t}$ and $I^{DP}_{b}$ corresponding to the top and bottom half-pixels on the sensor.

At capture time, the DP images are perfectly aligned with the right DC image. However, after stereo rectification, the left and right dual-camera images are respectively warped by spatial homography transformations $\mathbf{W}_{l}(x,y)$ and $\mathbf{W}_{r}(x,y)$, which remaps every pixel to new coordinates in both images. As a result, the right image is no longer aligned with the dual-pixel images. In addition, as explained in Section \ref{sec:dp}, the two disparity maps coming from DC and DP respectively, are related via an affine transformation. Our method takes both of these issues into account when fusing information from the two sources.

Our model uses two building blocks from other state-of-the-art stereo matching architectures. Specifically, a cost volume \cite{kendall2017end,khamis2018stereonet,psmnet} and refinement stages \cite{khamis2018stereonet,pang2017cascade}. Our main contributions are a method to fuse the \text{confidence volumes} (the softmax of the negative cost volume) computed from dual-cameras and dual-pixels, and to show the effectiveness of dual-pixels for refinement. Note that the proposed scheme can be used to give any stereo matching method that uses a cost-volume or that has a refinement stage, the benefits of the additional information in dual-pixels.

Our model consists of three stages, (a) extracting features and building cost volumes from DP and DC inputs independently, (b) building a fused confidence volume by fusing the DP and DC confidence volumes while accounting for the aforementioned spatial warp and affine ambiguity, and (c) a refinement stage that refines the coarse disparity from the fused confidence volume using features computed from the DP and DC images. We will now explain these in detail.

\subsection{Feature Extraction and Cost Volumes}

\vspace{2mm}\noindent
\paragraph{Dual-Camera Cost Volume.}
Inspired by \cite{khamis2018stereonet}, we create features that are $1/8$ of the spatial resolution in each axis of the original DC images. We do this by running three 2D convolutions with stride 2 followed by six residual blocks \cite{resnet}. Each convolutional layer uses a kernel of size $3\times3$ with 32 channels, followed by leaky ReLU activations. The resulting feature map from the left image is warped to the right feature map using multiple disparity hypotheses $d\in [0,16]$. For each hypothesis, we calculate the distance between the two feature maps. Unlike \cite{khamis2018stereonet}, we use the $\ell_1$ distance instead of subtracting the features since this increases the stability of training.

\vspace{2mm}\noindent
\paragraph{Dual-Pixel Cost Volume.}
Due to the tiny baseline ($<\sim 1$ mm), the disparities between DP views are usually in the range $-4$ to $4$ pixels.
Creating a cost volume with hypotheses that are one pixel apart (at reduced resolution) leads to unstable training. This is because all the possible disparities correspond to the same hypothesis.
Instead, we concatenate the two DP images and implicitly produce the cost volume.
To do so, we feed the DP images into a 2D network with six residual blocks. Each layer consists of convolutions of size $3\times3$ with 32 channels followed by leaky ReLU activations.
A 2D convolution is attached to the end to
produce a feature map with $N_d$ channels, where $N_d=17$ is the number of desired disparity hypotheses ($0.5$ pixels per sample from $-4$ to $4$). We then reshape the feature map and convert it to a 3D volume by expanding the final dimension.

\subsection{Fused Confidence Volume}
It is not straightforward to merge the DP and DC cost volumes due to the rectification warp $\mathbf{W}_r$ between the DP and DC images and the affine ambiguity in DP disparity (Sec~\ref{sec:dp}). In addition, the costs in the two volumes may be scaled differently since they are predicted from different network layers.

We first normalize the two cost volumes to the range $[0,1]$ by applying softmax to the negative cost volume along the disparity dimension. We call the resulting tensors confidence volumes. To handle the affine ambiguity, we first predict disparity maps $D_{DP}$ and $D_{DC}$ from the two confidence volumes using a soft--$\arg\max$ operator \cite{kendall2017end}, and fit an affine transformation between the two by solving a Tikhonov-regularized least squares problem that biases the solution to be close to $\alpha = 0, \beta = 1$:
\begin{equation}
    \hat{\alpha},\hat{\beta}=\underset{\alpha,\beta}{\mathit{argmin}}\left\Vert (\alpha + \beta\cdot D_{DP})-D_{DC}\right\Vert ^{2} + \gamma \left\Vert \beta - 1 \right\Vert^2 + \gamma \left\Vert \alpha \right\Vert^2,
\end{equation}
where $\gamma=0.1$ is a regularization constant.
Then, we use the known rectification warp $\mathbf{W}_r(x,y) = [W_{r}^{x}(x,y), W_{r}^{y}(x,y)]$ and the estimated affine transformation to warp the DP confidence volume into the DC space:
\begin{equation}
    C^{DP}_{\mathit{warp}}(x,y,z)=C^{DP}\left(W_{r}^{x}(x,y), W_{r}^{y}(x,y),(z-\hat{\alpha})/\hat{\beta}\right),
\end{equation}
where $C^{DP}$ is the confidence volume built from DP images. The 3D warping uses differentiable bilinear interpolation and zero padding.

The warped DP confidence volume $C^{DP}_{\mathit{warp}}$ is now normalized and aligned with the DC confidence volume. We stack these together to form a 4D tensor that is fused into a 3D confidence volume by a shallow network consisting of three layers of 3D convolutions with leaky ReLU activations and a softmax at the end. Finally, the fused volume is converted into a disparity map $D_{\mathit{unref}}$ using a soft--$\arg\max$.

\subsection{Disparity Refinement}
The next step is to refine the low resolution disparity $D_{\mathit{unref}}$. Khamis \etal \cite{khamis2018stereonet} use the RGB image as the guide image to upsample the disparity while applying a learned residual to improve edges and minimize the final error. A straightforward extension of \cite{khamis2018stereonet} is to warp the DP images using the rectification warp $\mathbf{W}_r$ and use them along with the RGB image as the guide image. However, we find that this yields inferior results compared to our method, presumably because warping and resampling makes it harder to extract disparity cues from DP images.

Instead, we extract features from the input DP images and then warp them using $\mathbf{W}_r$. This way, the model can easily extract disparity cues from the perfectly rectified DP image pair. The warped features are concatenated with the features extracted from the right RGB image and the unrefined disparity $D_{\mathit{unref}}$.
These are fed into six residual blocks, with $3\times3$ convolutions followed by a leaky ReLU activation, to predict a residual $R$. The final output is set to $D_{\mathit{ref}}=D_{\mathit{unref}}+R$.

\subsection{Loss Function}
To train our network we use a weighted Huber loss \cite{huber1992robust}:
\begin{equation}
    L(D) = \frac{\sum_\mathbf{p}\mathcal{H}\left(D(\mathbf{p})-D^{gt}(\mathbf{p}),\delta\right) \cdot C^{gt}(\mathbf{p})}{\sum_\mathbf{p} C^{gt}(\mathbf{p})},
\end{equation}
where $D^{gt}$ is the ground truth disparity, $C^{gt}$ is the per-pixel confidence of the ground truth, and $\delta$ is the switching point between the quadratic and the linear function, which is set to $1$ for disparity in range $[0, 128]$.
The overall loss is a weighted sum of four terms:
\begin{equation}
    L_{total} = \lambda_{DP}L(\hat{\alpha} + \hat{\beta}\cdot D_{DP})+\lambda_{DC}L(D_{DC})+\lambda_{\mathit{unref}}L(D_{\mathit{unref}})+\lambda_{\mathit{ref}}L(D_{\mathit{ref}}),
\end{equation}
where $\lambda_{DC}$ is set to $10$ and the other weights are set to $1$. %

\section{Evaluation}
In this section, we perform extensive experiments to evaluate our model. We conduct an ablation study to show the effectiveness of our design choices. We also compare to other stereo and dual-pixel methods. We focus our experiments on thin structures, edges and occlusion boundaries, to show the effectiveness of the complementary information coming from DP and DC data. For quantitative evaluations we report MAE, RMSE and the \textit{bad} $\delta$ metric, i.e. the percentage of pixels with disparity error greater than $\delta$. These are weighted by the ground-truth confidence. See the supplementary material for details.

\subsection{Data Collection}
\label{sec:data}
We collect a new data set using the Google Pixel 4 smartphone, which has a dual camera system consisting of a main camera with a dual-pixel sensor and a regular telephoto camera. We refer to the main camera as the right camera and the telephoto camera as the left camera. We use a data acquisition set up similar to \cite{garg2019learning}, i.e., a capture rig consisting of 5 phones (Fig.~\ref{fig:rig}) synchronized with \cite{SoftwareSync2019}. Structure from motion \cite{Hartley2003} and multi-view stereo techniques are used to generate depth maps. Similar to \cite{garg2019learning}, we also compute a per-pixel confidence for the depth by checking for depth coherence with neighboring views. The center phone in the rig is used for training and evaluation since its depth quality is higher than other views especially for occluded regions.

\newcommand{\datawidth}{0.16\textwidth}{}
\newcommand{\fakecaption}[1]{\raisebox{0.5 em}{\scriptsize #1}}
\begin{figure}[h]
    \centering
    \begin{tabular}{@{}c@{\,\,}c@{\,\,}c@{\,\,}c@{\,\,}c@{\,\,}c@{}}
    \includegraphics[width=\datawidth]{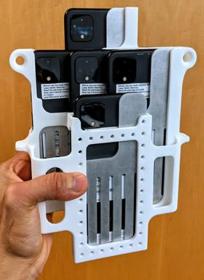} &
\includegraphics[width=\datawidth]{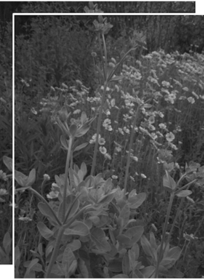} &
\includegraphics[width=\datawidth]{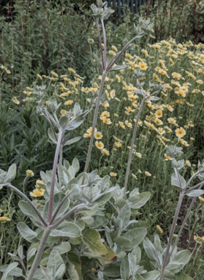} &
\includegraphics[width=\datawidth]{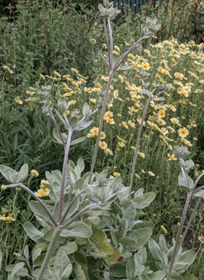} &
\includegraphics[width=\datawidth]{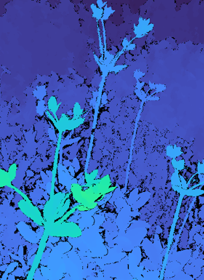} &
\includegraphics[width=\datawidth]{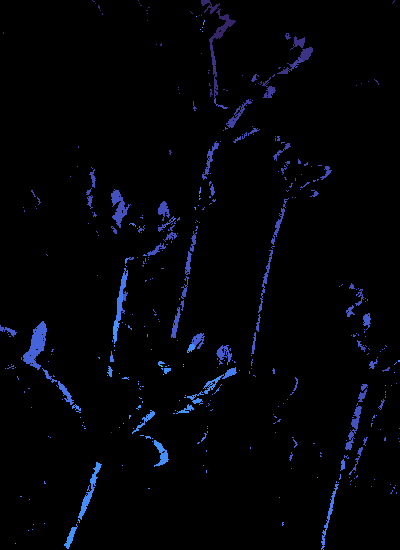}
\\
    \fakecaption{(a) Capture Rig} & \fakecaption{(b) DP Views} & \fakecaption{(c) Right View} & \fakecaption{(d) Left View} & \fakecaption{(e) GT} & \fakecaption{(f) GT (Occ.)}\\
    \addlinespace[-1.2ex]
    \end{tabular}
     \caption{Our capture rig (a) similar to \cite{garg2019learning} but with phones that can capture both dual-pixel (b) and dual-camera (c, d) data. 
    The left and right views are rectified, and the ground truth disparity (e) corresponding to the right view is computed using multi-view stereo techniques on all 10 views captured by the rig. Low confidence depth samples are rendered in black. The multitude of views ensures that we have good quality depth in regions that are occluded in the left view. (f) shows the GT depth masked to regions that are occluded in the left view.}
    \label{fig:rig}
    \vspace{-10pt}
\end{figure}

We rectify the stereo images from the center phone using estimated camera poses \cite{Fusiello00}. The estimated depth is converted to disparity and then rectified along with confidences to yield $D^{gt}_l, C^{gt}_l$ and $D^{gt}_r, C^{gt}_r$ for the left and right images respectively. As described in Section \ref{sec:method}, dual-pixel images from the main (right) camera are not rectified or warped. Instead, we store the warp map $\mathbf{W}_r$ that is needed to warp and align the DP images with the rectified right camera image (see Section \ref{sec:method}).
In addition, to evaluate the quality of the estimated disparity in regions that are visible in only one of the cameras in the stereo pair, we also compute $C^{\mathit{occ}}_i$ for $i\in l, r$, i.e., a per-pixel confidence indicating that the ground-truth disparity is correct but the pixel is occluded in the other camera.
In total, we collect 3308 training examples and 1077 testing examples. Please refer to supplementary materials for details of the calculation of $C^{\mathit{occ}}_i$, data collection and ground truth calculation.

\subsection{Training Scheme}
We use Tensorflow \cite{tensorflow2015-whitepaper} to implement the network and train using Adam \cite{kingma:adam} for 2 million iterations with a batch size of 1. The learning rate is set to $3 \times 10^{-5}$ and then reduced to $3 \times 10^{-6}$ after 1.5 million iterations. Training takes roughly $16$ hours using $8$ Tesla V100 GPUs. Inputs to the network $I_l$, $I_r$, and $\mathbf{W_r}$ are resized to match the resolution of the predicted and the ground truth disparity, i.e., $448\times560$. DP images $I^{DP}_t$ and $I^{DP}_b$ are of size $1000\times1250$.

\subsection{Ablation Study}

We evaluate the effect of each component of the model. In particular we focus on the impact of dual-pixels on the fused volume and the refinement stage. We provide quantitative comparisons (Tab. \ref{tab:ablation-real}) and qualitative comparisons (Fig.~\ref{fig:ablation}).
\begin{table}[t]
    \centering
\resizebox{\linewidth}{!}{
\begin{tabular}{c||cccc}
\hline
\multicolumn{5}{c}{DP in Confidence Volume}\tabularnewline
\hline 
Conf. Volume & MAE & RMSE & $\delta\!>\!2$  & $\delta\!>\!3$ \tabularnewline
\hline 
DC & 1.023 & 2.502 & 10.65 & 6.32\tabularnewline
DP+DC (2D) & 0.969 & 2.423 & 9.72 & 5.79\tabularnewline
DP+DC (C) & 0.964 & 2.372 & 9.79 & 5.80\tabularnewline
DP+DC (Ours) & \textbf{0.889} & \textbf{2.263} & \textbf{8.78} & \textbf{5.18}\tabularnewline
\hline 
\end{tabular}

\begin{tabular}{c||cccc}
\hline
\multicolumn{5}{c}{DP in Refinement}\tabularnewline
\hline 
Refinement & MAE & RMSE & $\delta\!>\!2$  & $\delta\!>\!3$ \tabularnewline
\hline 
RGB & 0.838 & 2.197 & 7.74 & 4.55\tabularnewline
DP (I) & 0.835 & 2.173 & 7.75 & 4.54\tabularnewline
DP & 0.829 & 2.184 & 7.51 & 4.45\tabularnewline
RGB+DP (Ours)  & \textbf{0.802} & \textbf{2.147} & \textbf{7.17} & \textbf{4.25}\tabularnewline
\hline 
\end{tabular}
}
\vspace{1mm}
    \caption{Ablation Study. Left: We compare different ways of fusing DP with the DC confidence volume. `(2D)' indicates fusion of the 2D disparity maps extracted from the two confidence volumes. `(C)' indicates fusing cost volumes instead of confidence volumes. Right: We compare the different ways of using DP to refine the best unrefined disparity from the left and show evaluation on the final disparity. `(I)' indicates that input DP images are warped before computing features for refinement.}
    \label{tab:ablation-real}
\vspace{-20pt}
\end{table}

\newcommand{\resultswidth}{0.16\textwidth}{}
\begin{figure}[h]
    \centering
    \begin{tabular}{@{}c@{\,\,}c@{\,\,}c@{\,\,}c@{\,\,}c@{\,\,}c@{}}
    \includegraphics[width=\resultswidth]{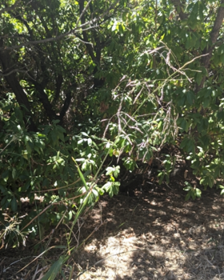} &
\includegraphics[width=\resultswidth]{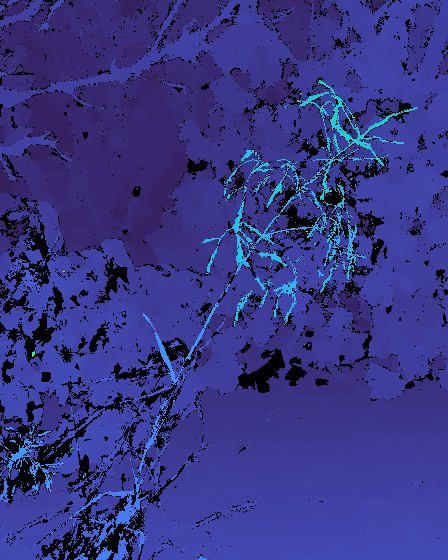} &
\includegraphics[width=\resultswidth]{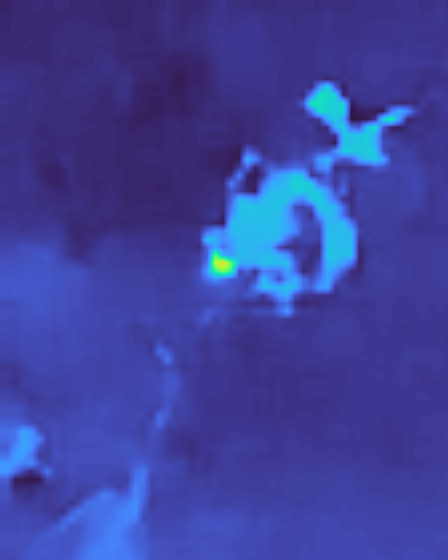} &
\includegraphics[width=\resultswidth]{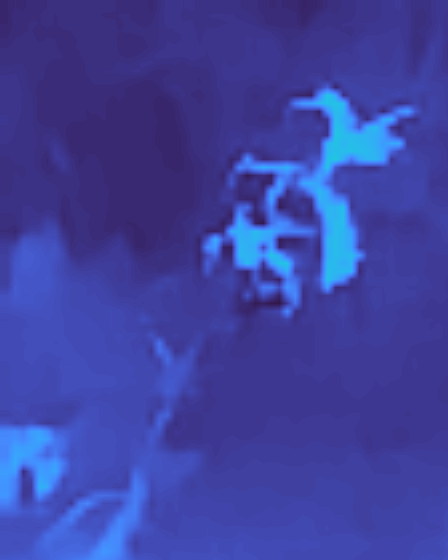} &
\includegraphics[width=\resultswidth]{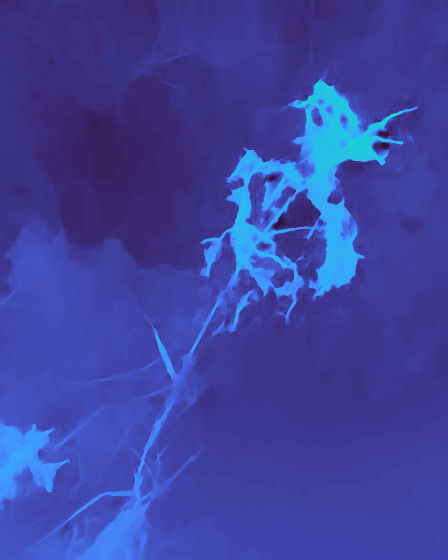} &
\includegraphics[width=\resultswidth]{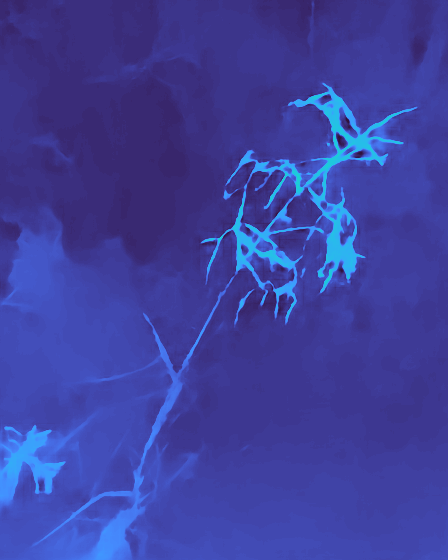}
\\
\includegraphics[width=\resultswidth]{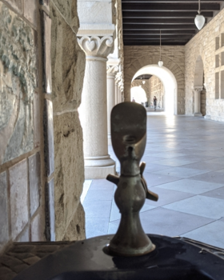} &
\includegraphics[width=\resultswidth]{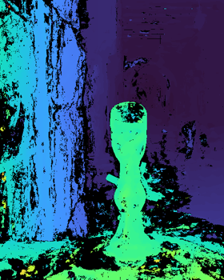} &
\includegraphics[width=\resultswidth]{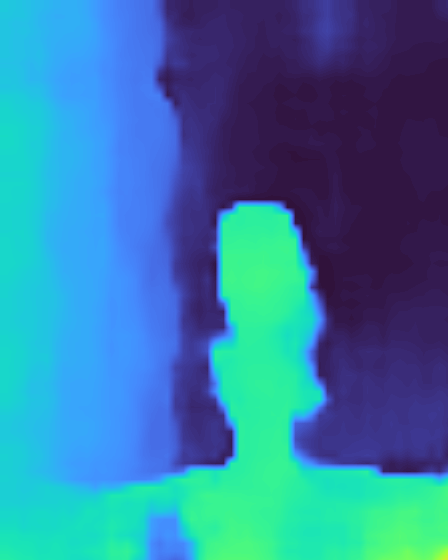} &
\includegraphics[width=\resultswidth]{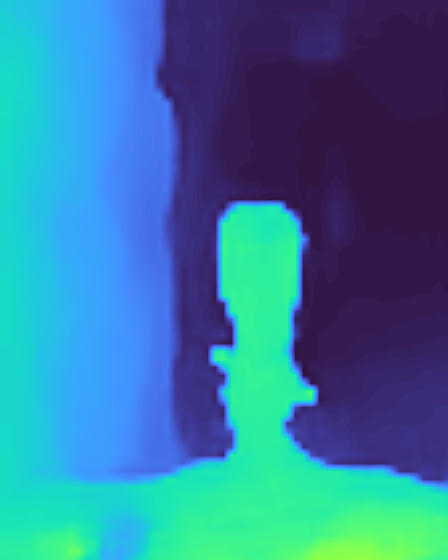} &
\includegraphics[width=\resultswidth]{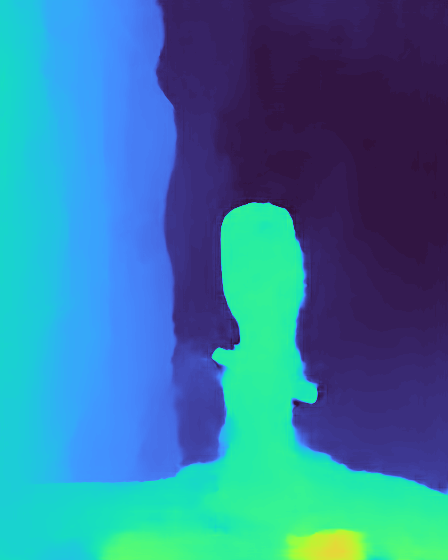} &
\includegraphics[width=\resultswidth]{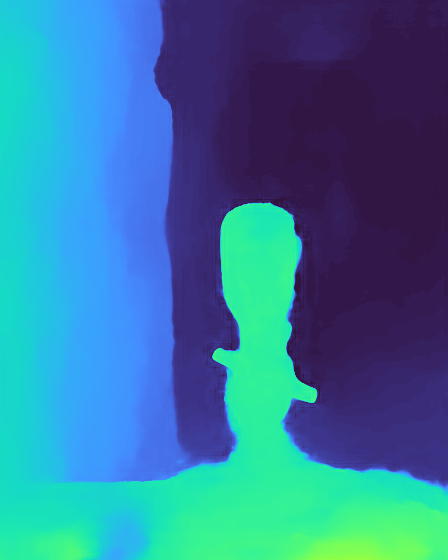}
\\
    \fakecaption{(a) \,Image} & \fakecaption{(b)\,GT} & \fakecaption{(c)\,$D_{\mathit{unref}}$} & \fakecaption{(d)\,$D_{\mathit{unref}}$} & \fakecaption{(e)\,$D_{\mathit{ref}}$} & \fakecaption{(f)\,$D_{\mathit{ref}}$}\\
    \addlinespace[-1.2ex]
    \fakecaption{} & \fakecaption{} & \fakecaption{DC, RGB+DP} & \fakecaption{\algoname} & \fakecaption{DP+DC,RGB} & \fakecaption{\algoname}\\
    \addlinespace[-2.2ex]
    \end{tabular}
    \caption{Ablations of our method. The right camera image (a), ground truth disparity (b) with low confidence disparity in black, $D_{\mathit{unref}}$ (c) from an ablation where only the DC input is used for the confidence volume, $D_{\mathit{unref}}$ (d) from \algoname, $D_{\mathit{ref}}$ (e) from an ablation where only the RGB image is used for refinement, and $D_{\mathit{ref}}$ (f) from \algoname. DP input is useful for both the confidence volume and refinement stages to recover accurate depth for fine structures and occluded regions.}
    \label{fig:ablation}
    \vspace{-15pt}
\end{figure}

\paragraph{Dual-pixels in the confidence volume.}
Tab. \ref{tab:ablation-real} (left) shows the error of the unrefined disparity $D_{\mathit{unref}}$ using different fusing strategies for the cost volume.
Our method of merging the DP and DC volumes (DP+DC) significantly outperforms using only the DC cost volume.
We also compare to fusing the 2D disparity maps instead of the the 3D confidence volumes. Specifically, we concatenate $D_{DP}$ and $D_{DC}$ after the affine transformation and use a 2D neural network with six residual blocks to predict $D_{\mathit{unref}}$. This is worse than our method according to all metrics (DP+DC (2D) vs DP+DC). Finally, we evaluate fusing cost volumes instead of confidence volumes. This, DP+DC (C), is also inferior to DP+DC.

Qualitative comparisons are shown in Fig. \ref{fig:ablation} (c) and (d). Compared to a DC only cost volume (c), fusing DP into the cost volume (d) adds more details to the unrefined disparity and prevents errors at object boundaries. This is critical for getting high quality disparities since the refinement is only able to make local adjustments to the disparity and cannot fix large errors.

\paragraph{Dual-pixels in refinement.}
We show that our method of using dual-pixels in the refinement stage is better than several baselines (Tab. \ref{tab:ablation-real} (right)). 
We use the fused DP + DC confidence volume for all cases and extract RGB and DP features for refinement using networks with the same capacity for a fair comparison.
Using DP for refinement is better than just using the right RGB image. However, results are best when both are used (RGB+DP).
Notably, if the DP is warped before feature extraction (instead of after), performance (DP (I) in Tab. \ref{tab:ablation-real}) is not better than using only the RGB image.
This suggests that DP cues are not effective after warping, and it is important to extract features before warping.

Qualitative comparisons are shown in Fig. \ref{fig:ablation} (e) and (f).
While both methods use DP and DC to compute the unrefined disparity, (e) uses only the right RGB image for refinement, and (f) uses RGB and DP for refinement.
Even though DP increases the quality of the unrefined disparity, using it during refinement further improves depth quality at thin structures and near object boundaries.
This indicates that it is important to use DP for both the cost volume and the refinement to achieve the best performance.

\subsection{Comparison to State-of-the-art Methods}

We compare to other stereo and dual-pixel depth estimation approaches (Tab.~\ref{tab:compare-others}). All methods are trained on our data using code provided by the authors. We modified the original loss functions to use the confidence maps $C^{gt}_r$ from our dataset in the same way that these are used in our method (as a per pixel weight). Additionally, for the stereo methods we compare to, we set the maximum disparity range to $128$. The baseline methods were trained until convergence on the test set error. We used the hyper-parameters and training optimization strategies (including annealing schedules for learning rates) provided in the original implementations.

\begin{table}[t]
    \centering
    
\begin{tabular}{c|c||cccc||cccc}
\hline 
\multirow{2}{*}{Method} & \multirow{2}{*}{Input} & \multicolumn{4}{c||}{All Pixels} & \multicolumn{4}{c}{Occluded Pixels}\tabularnewline
\cline{3-10} \cline{4-9} \cline{5-9} \cline{6-9} \cline{7-9} \cline{8-9} \cline{9-9} 
 & & MAE & RMSE & $\delta\!>\!2$ & $\delta\!>\!3$ & MAE & RMSE & $\delta\!>\!2$ & $\delta\!>\!3$\tabularnewline
\hline 
GA-Net \cite{ganet} & DC & 1.001 & 2.425 & 9.31 & 6.09 & 6.068 & 8.386 & 69.81 & 60.07\tabularnewline
PSM-Net \cite{psmnet} & DC & 0.815 & 2.289 & 7.98 & 4.97 & 2.799 & 5.188 & 34.73 & 26.78\tabularnewline
StereoNet \cite{khamis2018stereonet}& DC & 0.935 & 2.432 & 9.07 & 5.41 & 3.123 & 5.632 & 38.13 & 28.49\tabularnewline
\algoname (ours) & DC + DP & \textbf{0.802} & \textbf{2.147} & \textbf{7.17} &  \textbf{4.25} & \textbf{2.396} & \textbf{4.543} & \textbf{30.62} & \textbf{21.91} \tabularnewline
\hline 
DPNet{*} \cite{garg2019learning}& DP & 1.090 & 1.989 & 12.60 & 5.80 & 2.594 & \textbf{4.307} & 38.14 & 26.18\tabularnewline
\algoname{*} (ours)  & DC + DP & \textbf{0.746} & \textbf{1.825} & \textbf{6.63} & \textbf{3.63} & \textbf{2.352} & 4.373 & \textbf{32.65} & \textbf{21.77}\tabularnewline
\hline 
\end{tabular}
    \caption{Quantitative comparisons to the state-of-the-art. Note how the proposed approach substantially outperforms the StereoNet baseline \cite{khamis2018stereonet}, and GA-Net \cite{garg2019learning}. The method is on par with the more sophisticated PSM-Net \cite{psmnet} and it outperforms all the competitors in occluded regions. `*' indicates a final affine transformation applied to the output disparity and the best results for these methods are highlighted separately (see text for details).}
    \label{tab:compare-others}
    \vspace{-20pt}
\end{table}

For stereo baselines, we compare to StereoNet \cite{khamis2018stereonet} which is similar to our model with only the DC input, PSMNet \cite{psmnet} and GANet \cite{ganet}.
PSMNet \cite{psmnet} uses multi-scale feature extraction and cost volumes, and GANet \cite{ganet} uses a sophisticated semi-global aggregation \cite{ganet}.
In Tab.~\ref{tab:compare-others}, we report quantitative results. 
Our model uses a low-resolution cost volume and a refinement stage, with a run-time comparable with StereoNet, while achieving accuracy that is on par with or higher than more computationally expensive models, such as PSMNet (which uses \textit{25} 3D convolutional layers as opposed to our 8). 

For DP input baselines, DPNet \cite{garg2019learning} predicts disparity \textit{up to an unknown affine transformation}. To handle this, like \cite{garg2019learning}, we find the best fit (according to MSE) affine transformation between the prediction and the ground truth and transform the prediction to compute the metrics. For a fair comparison, we apply the same post processing to our method (\algoname{*} in Tab.~\ref{tab:compare-others}), showing it consistently outperforms DPNet. 

As mentioned in Sec.~\ref{sec:data}, we also compute an occlusion mask  $C^{\mathit{occ}}_r$ and evaluate the methods only on these pixels.
The results are reported in Tab. \ref{tab:compare-others} under ``Occluded pixels''.
Our method outperforms the other competitors by a substantial margin in those areas showing the advantage of small baseline DP data.

\begin{figure}[t]
    \centering
    \begin{tabular}{@{}c@{\,\,}c@{\,\,}c@{\,\,}c@{\,\,}c@{\,\,}c@{}}
\includegraphics[width=\resultswidth]{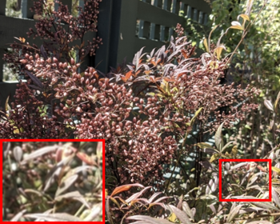} &
\includegraphics[width=\resultswidth]{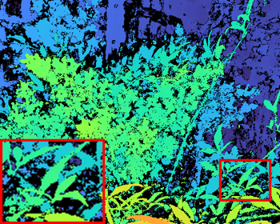} &
\includegraphics[width=\resultswidth]{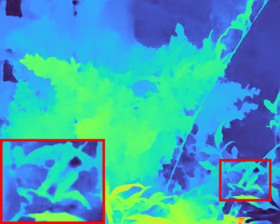} &
\includegraphics[width=\resultswidth]{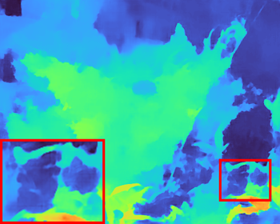} &
\includegraphics[width=\resultswidth]{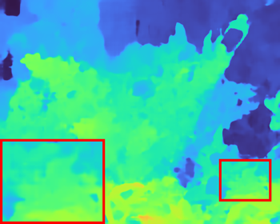} &
\includegraphics[width=\resultswidth]{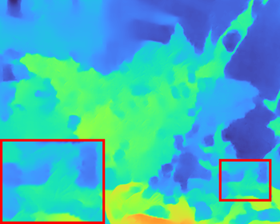}
\\
\includegraphics[width=\resultswidth]{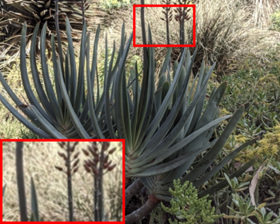} &
\includegraphics[width=\resultswidth]{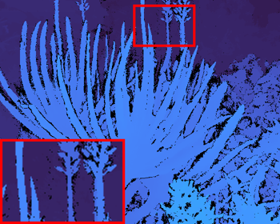} &
\includegraphics[width=\resultswidth]{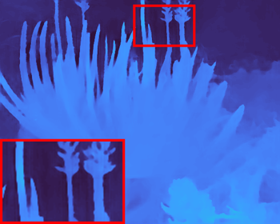} &
\includegraphics[width=\resultswidth]{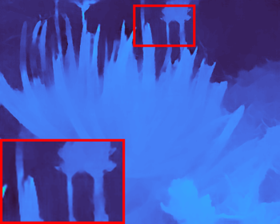} &
\includegraphics[width=\resultswidth]{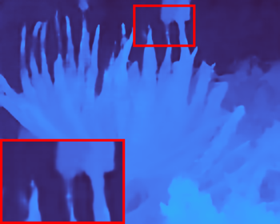} &
\includegraphics[width=\resultswidth]{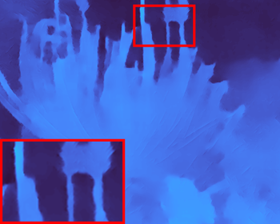}
\\
\includegraphics[width=\resultswidth]{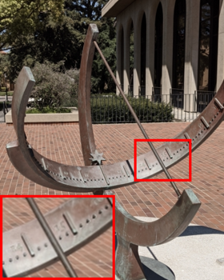} &
\includegraphics[width=\resultswidth]{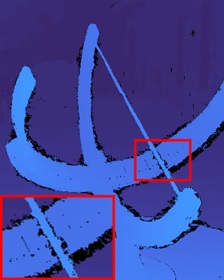} &
\includegraphics[width=\resultswidth]{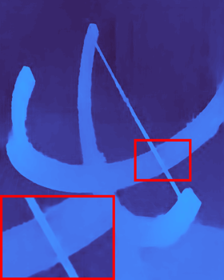} &
\includegraphics[width=\resultswidth]{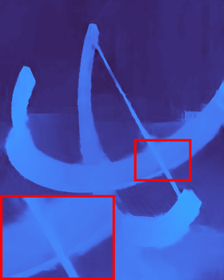} &
\includegraphics[width=\resultswidth]{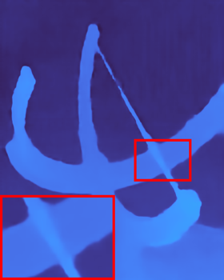} &
\includegraphics[width=\resultswidth]{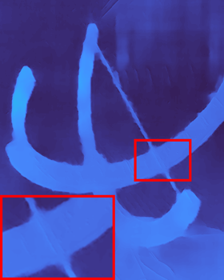}
\\
\includegraphics[width=\resultswidth]{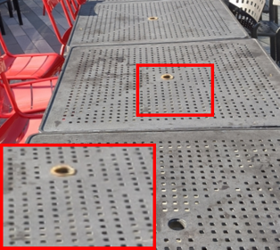} &
\includegraphics[width=\resultswidth]{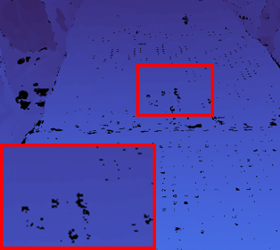} &
\includegraphics[width=\resultswidth]{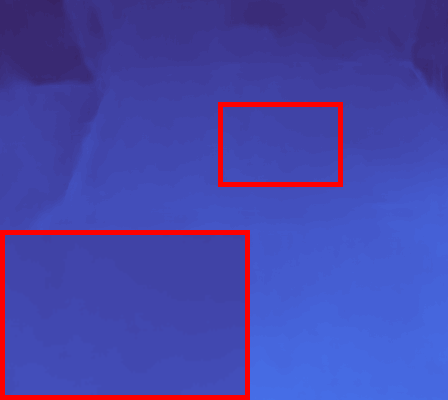} &
\includegraphics[width=\resultswidth]{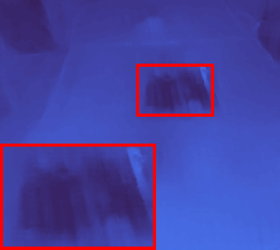} &
\includegraphics[width=\resultswidth]{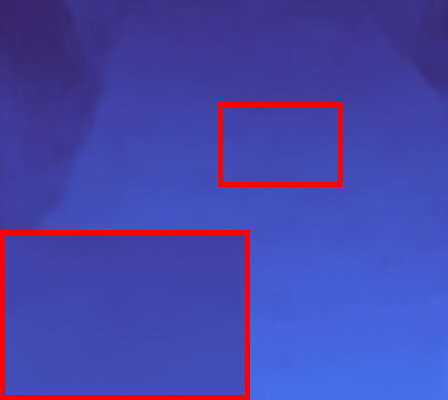} &
\includegraphics[width=\resultswidth]{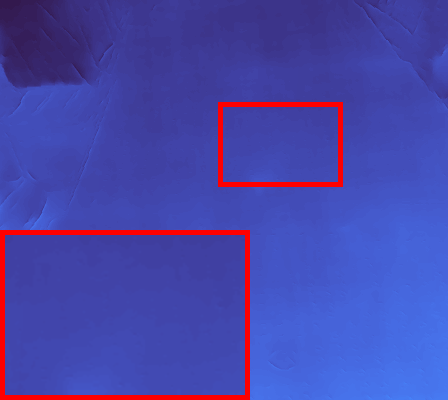}
\\
    \fakecaption{(a) \,Image} & \fakecaption{(b)\,GT} & \fakecaption{(c)\,Ours} & \fakecaption{(d)\,StereoNet} & \fakecaption{(e)\,PSMNet} & \fakecaption{(f)\,DPNet}\\
    \addlinespace[-2.2ex]
    \end{tabular}
    \caption{Qualitative comparison to the state-of-the-art. Right camera image (a) from our test set, ground truth disparity with low confidence disparity in black (b),  and results from our method (c), stereo only methods StereoNet \cite{khamis2018stereonet} (d) and  PSMNet \cite{psmnet} (e), and DP only method DPNet \cite{garg2019learning} (f).  Stereo only methods fail for vertical structures due to the aperture problem, e.g., in the second image. They also fail in regions with fine structures and occlusions, e.g., in the first three images. StereoNet fails on the last image potentially due to repeated texture. DPNet's accuracy falls quickly with distance due to the small dual-pixel baseline. Our method overcomes these problems by fusing the two cues.}
    \label{fig:gallery}
    \vspace{-10pt}
\end{figure}

Qualitative comparisons are provided in Fig. ~\ref{fig:gallery}. Note how we better capture fine details and small structures, while correctly inferring disparity near occlusion boundaries. The orthogonal baselines of the dual-pixels and dual-cameras also helps mitigate the aperture problem and issues due to repeated textures. Additional results are available in the supplementary material.

\subsection{Applications in Computational Photography}

Predicting accurate disparities, hence depth, is crucial for many applications in computational photography. These applications usually require accurate depth for fine structures and near occlusion boundaries. Fig.~\ref{fig:bokeh} and \ref{fig:3d_photo} show how our more accurate depth leads to fewer artifacts when used to produce synthetic shallow depth-of-field images and 3D photos \cite{Instant3D2018} respectively.

\newcommand{\appresultswidth}{0.19\textwidth}{}
\newcommand{\appfakecaption}[1]{\raisebox{0.5 em}{\scriptsize #1}}
\begin{figure}[t]
    \centering
    \begin{tabular}{@{}c@{\,\,}c@{\,\,}c@{\,\,}c@{\,\,}c@{}}
\includegraphics[width=\appresultswidth]{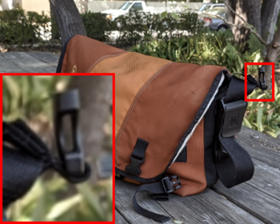} &
\includegraphics[width=\appresultswidth]{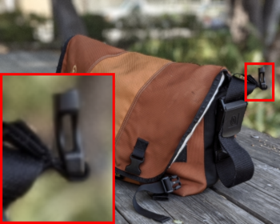} &
\includegraphics[width=\appresultswidth]{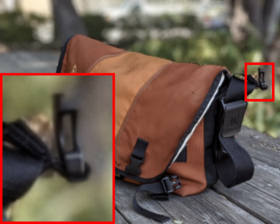} &
\includegraphics[width=\appresultswidth]{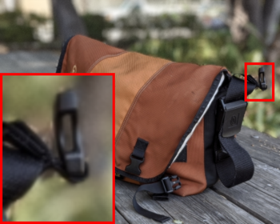} &
\includegraphics[width=\appresultswidth]{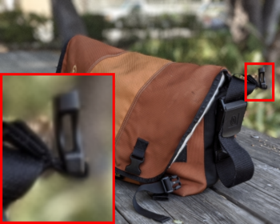} \\
\includegraphics[width=\appresultswidth]{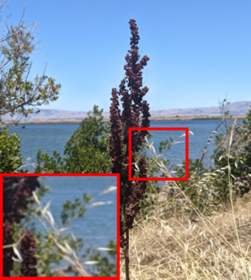} &
\includegraphics[width=\appresultswidth]{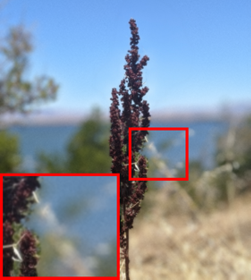} &
\includegraphics[width=\appresultswidth]{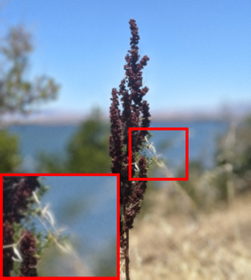} &
\includegraphics[width=\appresultswidth]{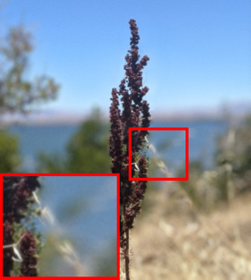} &
\includegraphics[width=\appresultswidth]{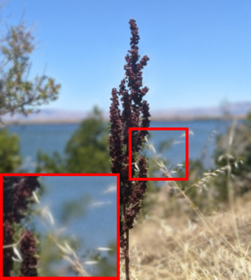} \\
    \appfakecaption{(a) \,Image} & \appfakecaption{(c)\,Ours} & \appfakecaption{(d)\,StereoNet\cite{khamis2018stereonet}} & \appfakecaption{(e)\,PSMNet\cite{psmnet}} & \appfakecaption{(f)\,DPNet\cite{garg2019learning}}\\
    \end{tabular}
    \caption{Synthetic shallow depth-of-field results for different methods. %
    Top: Accurate depth near occlusion boundaries is critical for avoiding artifacts near the subject boundary. Bottom: DPNet \cite{garg2019learning} is unable to resolve the small depth difference between the flower and the twigs in the background. As a result, parts of the background are incorrectly sharp.
    }
    \label{fig:bokeh}
    \vspace{-10pt}
\end{figure}

\begin{figure}
    \centering
    \begin{tabular}{@{}c@{\,\,}c@{\,\,}c@{\,\,}c@{\,\,}c@{}}
\includegraphics[height=1.9cm,width=\appresultswidth]{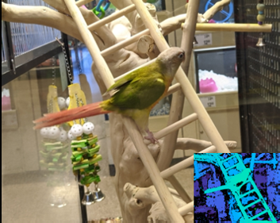} &
\includegraphics[height=1.9cm,width=\appresultswidth]{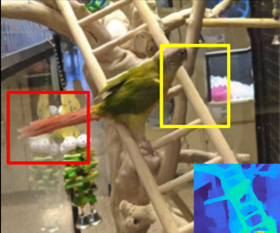} &
\includegraphics[height=1.9cm,width=\appresultswidth]{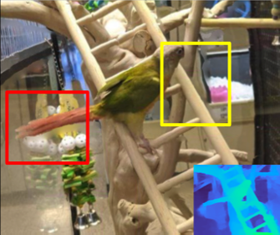} &
\includegraphics[height=1.9cm,width=\appresultswidth]{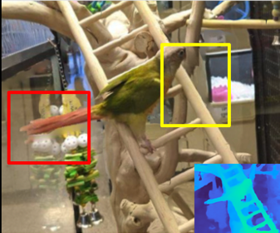} &
\includegraphics[height=1.9cm,width=\appresultswidth]{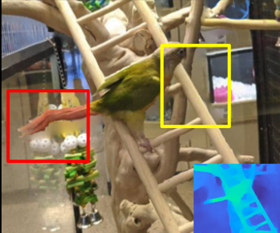} \\
\includegraphics[height=1.9cm,width=\appresultswidth]{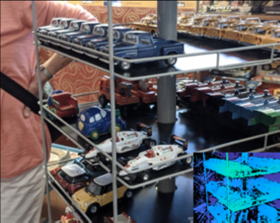} &
\includegraphics[height=1.9cm,width=\appresultswidth]{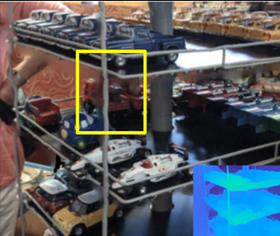} &
\includegraphics[height=1.9cm,width=\appresultswidth]{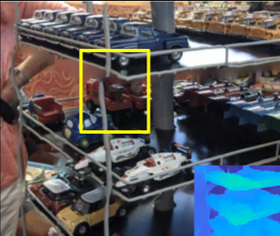} &
\includegraphics[height=1.9cm,width=\appresultswidth]{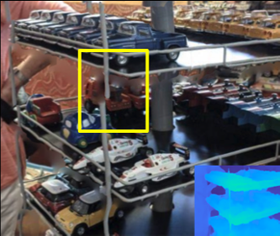} &
\includegraphics[height=1.9cm,width=\appresultswidth]{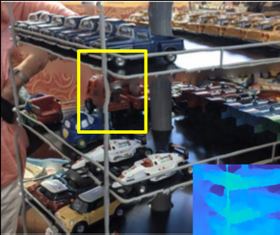}\\
    \appfakecaption{(a) \,Image} & \appfakecaption{(c)\,Ours} & \appfakecaption{(d)\,StereoNet\cite{khamis2018stereonet}} & \appfakecaption{(e)\,PSMNet\cite{psmnet}} & \appfakecaption{(f)\,DPNet\cite{garg2019learning}}\\
    \end{tabular}
    \caption{3D photos results \cite{Instant3D2018}. Novel views of the scene are rendered by warping the image according to the estimated depth to new camera positions. Depth errors lead to unnatural distortion of rigid scene structures in the novel views.}
    \label{fig:3d_photo}
    \vspace{-10pt}
\end{figure}

\section{Discussion}
We presented the first method to combine dual-camera and dual-pixel data. The inherent affine ambiguity of disparity computed from dual-pixel images prevents a straightforward integration of the two modalities. Therefore, we proposed a novel solution that resamples the confidence volume computed from dual-pixels and concatenates it with the dual-camera volume. A refinement stage leverages dual-pixels to infer the final disparity map. We show the effectiveness of the proposed solution with experiments, comparisons to the state-of-the-art and applications. While the orthogonality of the baselines allows us to avoid the aperture problem, our method doesn't work on textureless regions. Perhaps this could be handled by combining information from additional modalities like active depth sensors. Another interesting direction for future work would be to consider dual-camera pairs where both cameras have dual-pixels. 

\clearpage
\bibliographystyle{splncs04}
\bibliography{references}

\begin{thebibliography}{10}
\providecommand{\url}[1]{\texttt{#1}}
\providecommand{\urlprefix}{URL }
\providecommand{\doi}[1]{https://doi.org/#1}

\bibitem{tensorflow2015-whitepaper}
Abadi, M., Agarwal, A., Barham, P., Brevdo, E., et~al.: {TensorFlow}:
  Large-scale machine learning on heterogeneous systems (2015),
  \url{https://www.tensorflow.org/}

\bibitem{SoftwareSync2019}
Ansari, S., Wadhwa, N., Garg, R., Chen, J.: Wireless software synchronization
  of multiple distributed cameras. In: ICCP (2019)

\bibitem{besse2014pmbp}
Besse, F., Rother, C., Fitzgibbon, A., Kautz, J.: {PMBP:} patchmatch belief
  propagation for correspondence field estimation. IJCV  (2014)

\bibitem{bleyer2011patchmatch}
Bleyer, M., Rhemann, C., Rother, C.: Patchmatch stereo-stereo matching with
  slanted support windows. In: BMVC (2011)

\bibitem{psmnet}
Chang, J., Chen, Y.: Pyramid stereo matching network. In: CVPR (2018)

\bibitem{Diverdi2016}
DiVerdi, S., Barron, J.T.: Geometric calibration for mobile, stereo, autofocus
  cameras. In: WACV (2016)

\bibitem{DeepPrunerICCV2019}
Duggal, S., Wang, S., Ma, W.C., Hu, R., Urtasun, R.: {DeepPruner:} learning
  efficient stereo matching via differentiable patchmatch. In: ICCV (2019)

\bibitem{fanello17_hashmatch}
Fanello, S.R., Valentin, J., Kowdle, A., Rhemann, C., Tankovich, V., Ciliberto,
  C., Davidson, P., Izadi, S.: Low compute and fully parallel computer vision
  with hashmatch. In: ICCV (2017)

\bibitem{fanello2017ultrastereo}
Fanello, S.R., Valentin, J., Rhemann, C., Kowdle, A., Tankovich, V., Davidson,
  P., Izadi, S.: {UltraStereo:} efficient learning-based matching for active
  stereo systems. In: CVPR (2017)

\bibitem{fanello14}
Fanello, S., Pattacini, U., Gori, I., Tikhanoff, V., Randazzo, M., Roncone, A.,
  Odone, F., Metta, G.: 3d stereo estimation and fully automated learning of
  eye-hand coordination in humanoid robots. In: Humanoids (2014)

\bibitem{felzenszwalb2006efficient}
Felzenszwalb, P.F., Huttenlocher, D.P.: Efficient belief propagation for early
  vision. IJCV  (2006)

\bibitem{Fusiello00}
Fusiello, A., Trucco, E., Verri, A.: A compact algorithm for rectification of
  stereo pairs. Machine Vision and Applications  (2000)

\bibitem{garg2019learning}
Garg, R., Wadhwa, N., Ansari, S., Barron, J.T.: Learning single camera depth
  estimation using dual-pixels. In: ICCV (2019)

\bibitem{gidaris2017detect}
Gidaris, S., Komodakis, N.: Detect, replace, refine: Deep structured prediction
  for pixel wise labeling. In: CVPR (2017)

\bibitem{relightables}
Guo, K., Lincoln, P., Davidson, P., Busch, J., Yu, X., Whalen, M., Harvey, G.,
  Orts-Escolano, S., Pandey, R., Dourgarian, J., Tang, D., Tkach, A., Kowdle,
  A., Cooper, E., Dou, M., Fanello, S., Fyffe, G., Rhemann, C., Taylor, J.,
  Debevec, P., Izadi, S.: The relightables: Volumetric performance capture of
  humans with realistic relighting. TOG  (2019)

\bibitem{hamzah2016literature}
Hamzah, R.A., Ibrahim, H.: Literature survey on stereo vision disparity map
  algorithms. Journal of Sensors  (2016)

\bibitem{Hartley2003}
Hartley, R., Zisserman, A.: Multiple View Geometry in Computer Vision.
  Cambridge University Press (2003)

\bibitem{resnet}
{He}, K., {Zhang}, X., {Ren}, S., {Sun}, J.: Deep residual learning for image
  recognition. In: CVPR (2016)

\bibitem{Instant3D2018}
Hedman, P., Kopf, J.: {Instant 3{D} Photography}. SIGGRAPH  (2018)

\bibitem{hirschmuller2008stereo}
Hirschmuller, H.: Stereo processing by semiglobal matching and mutual
  information. TPAMI  (2008)

\bibitem{huber1992robust}
Huber, P.J.: Robust estimation of a location parameter. In: Breakthroughs in
  statistics, pp. 492--518. Springer (1992)

\bibitem{ilg2017flownet}
Ilg, E., Mayer, N., Saikia, T., Keuper, M., Dosovitskiy, A., Brox, T.: Flownet
  2.0: Evolution of optical flow estimation with deep networks. In: CVPR (2017)

\bibitem{kendall2017end}
Kendall, A., Martirosyan, H., Dasgupta, S., Henry, P., Kennedy, R., Bachrach,
  A., Bry, A.: End-to-end learning of geometry and context for deep stereo
  regression. In: CVPR (2017)

\bibitem{khamis2018stereonet}
Khamis, S., Fanello, S., Rhemann, C., Kowdle, A., Valentin, J., Izadi, S.:
  Stereonet: Guided hierarchical refinement for real-time edge-aware depth
  prediction. In: ECCV (2018)

\bibitem{kingma:adam}
Kingma, D.P., Ba, J.: Adam: A method for stochastic optimization. In: ICLR
  (2015)

\bibitem{klaus2006segment}
Klaus, A., Sormann, M., Karner, K.: Segment-based stereo matching using belief
  propagation and a self-adapting dissimilarity measure. In: ICPR (2006)

\bibitem{kolmogorov2001computing}
Kolmogorov, V., Zabih, R.: Computing visual correspondence with occlusions
  using graph cuts. In: ICCV (2001)

\bibitem{iResNet2018}
Liang, Z., Feng, Y., Guo, Y., Liu, H., Chen, W., Qiao, L., Zhou, L., Zhang, J.:
  Learning for disparity estimation through feature constancy. In: CVPR (2018)

\bibitem{marr1976cooperative}
Marr, D., Poggio, T.: Cooperative computation of stereo disparity. Science
  (1976)

\bibitem{mayer2016large}
Mayer, N., Ilg, E., Hausser, P., Fischer, P., Cremers, D., Dosovitskiy, A.,
  Brox, T.: A large dataset to train convolutional networks for disparity,
  optical flow, and scene flow estimation. In: CVPR (2016)

\bibitem{meier2017real}
Meier, L., Honegger, D., Vilhjalmsson, V., Pollefeys, M.: Real-time stereo
  matching failure prediction and resolution using orthogonal stereo setups.
  In: ICRA (2017)

\bibitem{Menze2015KITTI}
Menze, M., Geiger, A.: Object scene flow for autonomous vehicles. In: CVPR
  (2015)

\bibitem{morgan1997aperture}
Morgan, M., Castet, E.: The aperture problem in stereopsis. Vision research
  (1997)

\bibitem{mulligan2002trinocular}
Mulligan, J., Isler, V., Daniilidis, K.: Trinocular stereo: A real-time
  algorithm and its evaluation. IJCV  (2002)

\bibitem{ng2005lightcamera}
Ng, R., Levoy, M., Br{\'e}dif, M., Duval, G., Horowitz, M., Hanrahan, P.: Light
  field photography with a hand-held plenoptic camera. Tech. rep., Stanford
  University (2005)

\bibitem{holoportation}
Orts-Escolano, S., Rhemann, C., Fanello, S., Chang, W., Kowdle, A., Degtyarev,
  Y., Kim, D., Davidson, P.L., Khamis, S., Dou, M., Tankovich, V., Loop, C.,
  Cai, Q., Chou, P.A., Mennicken, S., Valentin, J., Pradeep, V., Wang, S.,
  Kang, S.B., Kohli, P., Lutchyn, Y., Keskin, C., Izadi, S.: Holoportation:
  Virtual 3d teleportation in real-time. In: UIST (2016)

\bibitem{pang2017cascade}
Pang, J., Sun, W., Ren, J., Yang, C., Yan, Q.: Cascade residual learning: A
  two-stage convolutional neural network for stereo matching. In: ICCV Workshop
  (2017)

\bibitem{Punnappurath_2019_CVPR}
Punnappurath, A., Brown, M.S.: Reflection removal using a dual-pixel sensor.
  In: CVPR (2019)

\bibitem{Richardt2010}
Richardt, C., Orr, D., Davies, I., Criminisi, A., Dodgson, N.A.: Real-time
  spatiotemporal stereo matching using the dual-cross-bilateral grid. In: ECCV
  (2010)

\bibitem{Scharstein2002}
Scharstein, D., Szeliski, R.: A taxonomy and evaluation of dense two-frame
  stereo correspondence algorithms. IJCV  (2002)

\bibitem{Sinha2014}
Sinha, S.N., Scharstein, D., Szeliski, R.: Efficient high-resolution stereo
  matching using local plane sweeps. In: CVPR (2014)

\bibitem{EdgeStereo2018}
Song, X., Zhao, X., Hu, H., Fang, L.: {EdgeStereo:} {A} context integrated
  residual pyramid network for stereo matching. In: ACCV (2018)

\bibitem{szeliski2010}
Szeliski, R.: Computer Vision: Algorithms and Applications. Springer-Verlag,
  Berlin, Heidelberg, 1st edn. (2010)

\bibitem{sos}
Tankovich, V., Schoenberg, M., Fanello, S.R., Kowdle, A., Rhemann, C.,
  Dzitsiuk, M., Schmidt, M., Valentin, J., Izadi, S.: Sos: Stereo matching in
  {O(1)} with slanted support windows. In: IROS (2018)

\bibitem{wadhwa2018}
Wadhwa, N., Garg, R., Jacobs, D.E., Feldman, B.E., Kanazawa, N., Carroll, R.,
  Movshovitz{-}Attias, Y., Barron, J.T., Pritch, Y., Levoy, M.: Synthetic
  depth-of-field with a single-camera mobile phone. In: SIGGRAPH (2018)

\bibitem{SegStereo2018}
Yang, G., Zhao, H., Shi, J., Deng, Z., Jia, J.: {SegStereo:} exploiting
  semantic information for disparity estimation. In: ECCV (2018)

\bibitem{ganet}
Zhang, F., Prisacariu, V.A., Yang, R., Torr, P.H.S.: {GA-Net:} guided
  aggregation net for end-to-end stereo matching. In: CVPR (2019)

\end{thebibliography}

\clearpage
\renewcommand\thesection{\Alph{section}}

\setcounter{section}{0}

\begin{center}
\vspace{10mm}
\textbf{\Large Supplementary Materials}
\end{center}

In this supplementary material, we provide more information for data collection, implementation details, and quantitative and qualitative results. 

\section{Data Collection}
In this section, we provide information about our data capture rig and how we obtain the ground truth disparity, confidence, and the occlusion mask.

\subsection{Data Capture}

\newcommand{\viewsresultswidth}{0.19\textwidth}{}
\begin{figure}[h]
    \centering
    \begin{tabular}{@{}c@{\,\,}c@{\,\,}c@{\,\,}c@{\,\,}c@{}}
    \includegraphics[width=\viewsresultswidth]{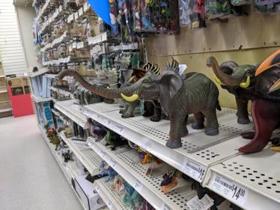} &
    \includegraphics[width=\viewsresultswidth]{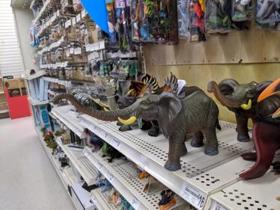} &
    \includegraphics[width=\viewsresultswidth]{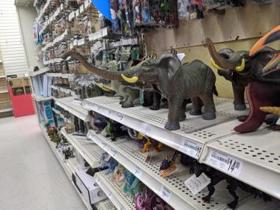} &
    \includegraphics[width=\viewsresultswidth]{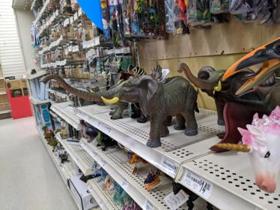} &
    \includegraphics[width=\viewsresultswidth]{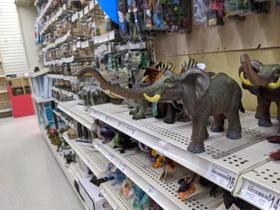}\\
    \includegraphics[width=\viewsresultswidth]{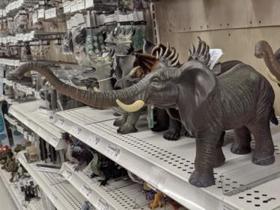} &
    \includegraphics[width=\viewsresultswidth]{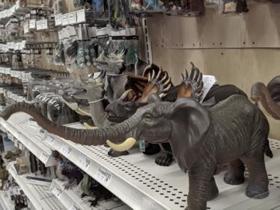} &
    \includegraphics[width=\viewsresultswidth]{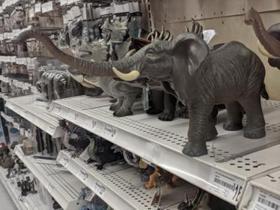} &
    \includegraphics[width=\viewsresultswidth]{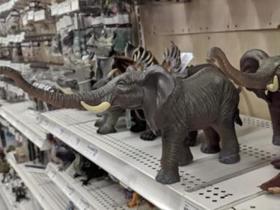} &
    \includegraphics[width=\viewsresultswidth]{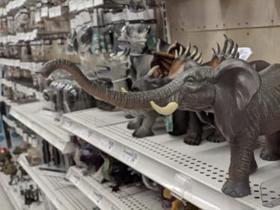}\\
     \fakecaption{View 1} & \fakecaption{View 2} & \fakecaption{View 3} & \fakecaption{View 4} & \fakecaption{View 5}\\
    \end{tabular}
    \caption{Example capture from our data collection rig. Top: Views from the main cameras of the five phones on the rig. Bottom: Views from the telephoto cameras. All ten views are used to compute ground truth depth using multi-view stereo techniques.}
    \label{fig:gt_views}
    \vspace{-10pt}
\end{figure}

As shown in Fig. 4(a) in the main paper, our capture rig consists of five Google Pixel 4 phones. Each phone captures a stereo pair (and dual-pixel data), giving us ten views of the same scene (Fig.~\ref{fig:gt_views}). The five phones are synchronized using \cite{SoftwareSync2019} allowing us to capture dynamic scenes, e.g., plants moving in the wind. Our dataset is captured both indoors and outdoors, and contains both man made and natural scenes.

\newcommand{\gtresultswidth}{0.2\textwidth}{}
\begin{figure}[hp]
    \centering
    \begin{tabular}{@{}c@{\,\,}c@{\,\,}c@{\,\,}c@{}}
\includegraphics[width=\gtresultswidth]{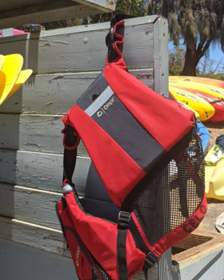} &
\includegraphics[width=\gtresultswidth]{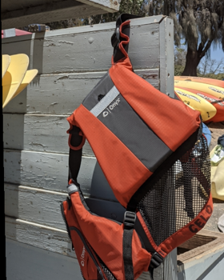} &
\includegraphics[width=\gtresultswidth]{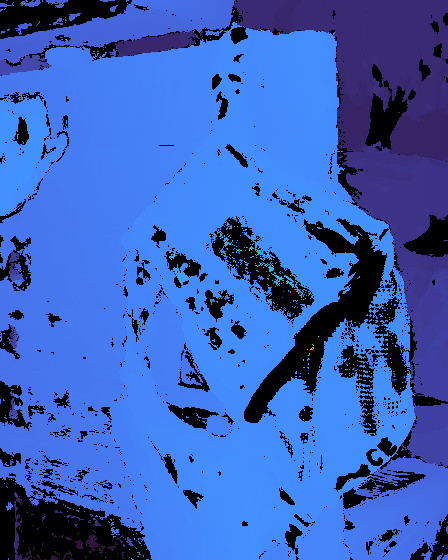} &
\includegraphics[width=\gtresultswidth]{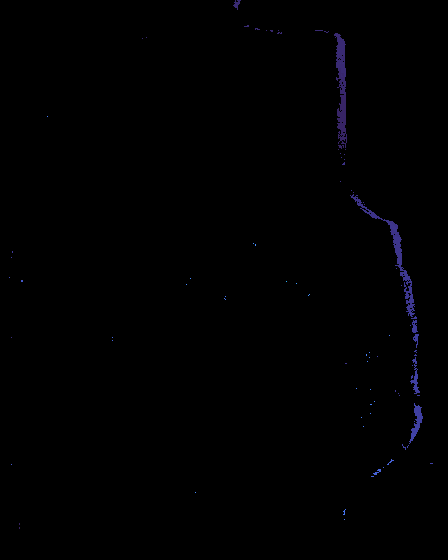}
\\
\includegraphics[width=\gtresultswidth]{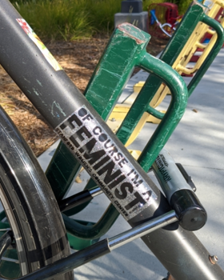} &
\includegraphics[width=\gtresultswidth]{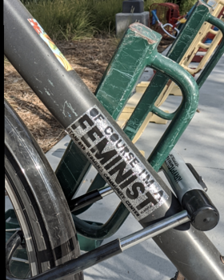} &
\includegraphics[width=\gtresultswidth]{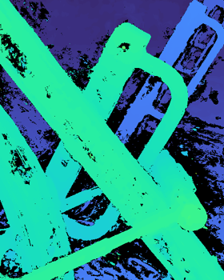} &
\includegraphics[width=\gtresultswidth]{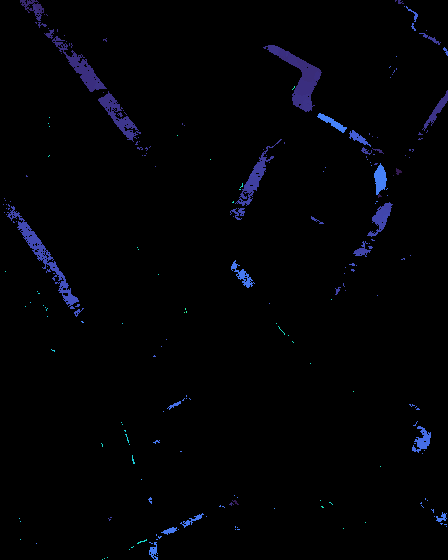}
\\
\includegraphics[width=\gtresultswidth]{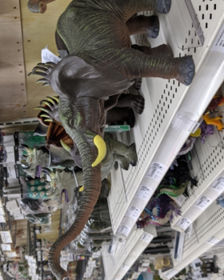} &
\includegraphics[width=\gtresultswidth]{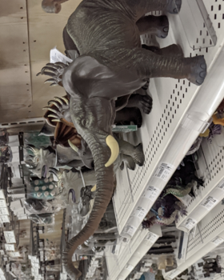} &
\includegraphics[width=\gtresultswidth]{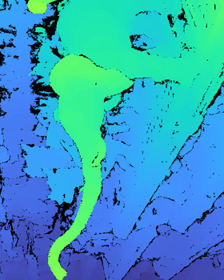} &
\includegraphics[width=\gtresultswidth]{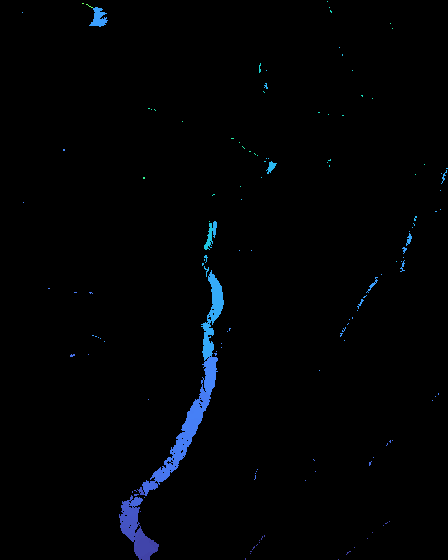}
\\
\includegraphics[width=\gtresultswidth]{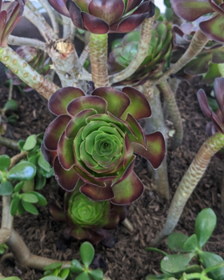} &
\includegraphics[width=\gtresultswidth]{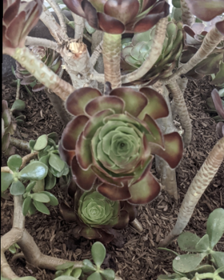} &
\includegraphics[width=\gtresultswidth]{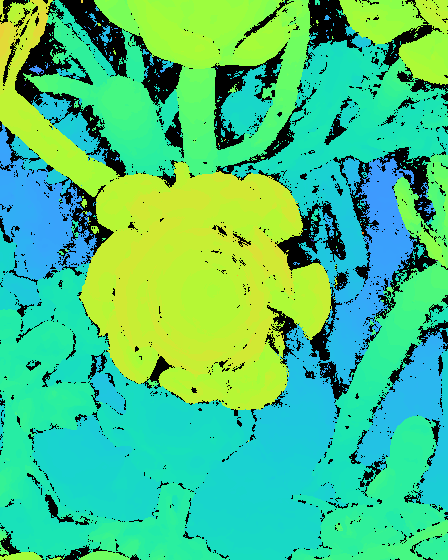} &
\includegraphics[width=\gtresultswidth]{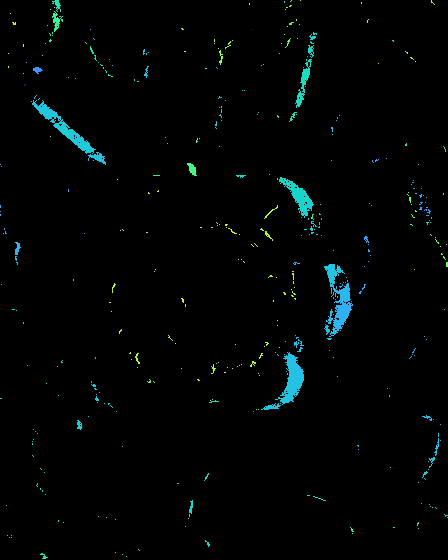}
\\
\fakecaption{(a) \,Right Image} & \fakecaption{(b)\,Left Image} & \fakecaption{(c)\,GT} & \fakecaption{(d)\,GT(Occ.)}\\
    \end{tabular}
    \caption{Examples of collected data. The right view and left view (a,b) of the binocular stereo pair, the ground truth disparity (c), and the ground truth disparity for occluded pixels (d). Low confidence disparity is rendered in black in (c) and (d).}
    \label{fig:gt}
    \vspace{-10pt}
\end{figure}

\subsection{Computing Ground Truth Disparity $D^{gt}$ and Confidence $C^{gt}$}
\label{sec:gt_conf}

We now describe how we compute $D^{gt}$ and $C^{gt}$ given a capture from the rig.

Since the rig may not be perfectly rigid, we first compute camera poses, i.e., intrinsics and extrinsics, using structure from motion \cite{Hartley2003}. For computing ground truth depth using multi-view stereo, we use a method similar to \cite{garg2019learning} that is designed to give accurate depth for fine structures while avoiding edge fattening artifacts. We describe it in more detail below.

All ten RGB images are resized to $756\times 1008$. For each view, we use a plane sweep algorithm, with $256$ planes sampled using inverse perspective sampling between $0.2$m  and $100$m, and take the minimum of a filtered cost volume as each pixel's depth. To compute the cost volume, for each pixel, we compute the sum of absolute differences for each of the warped neighbors and then bilaterally filter the cost volume using the grayscale reference image as the guide image thus avoiding edge fattening artifacts \cite{Richardt2010}. We use a spatial sigma of $3$ pixels and a range sigma of $12.5$ for the bilateral filter.

Following \cite{garg2019learning}, we also estimate per-pixel confidence for depth, i.e., a scalar in the range $[0,1]$. Specifically, we check for depth coherence across views by checking for left / right consistency \cite{bleyer2011patchmatch}. We first compute consistency with each of the $9$ neighboring images using the consistency measure in \cite{garg2019learning}. Then, under the assumption that a pixel must be visible in at least two other cameras for its depth to be reliable, we take the product of the largest two consistency values for each pixel to compute our final confidence.

Even though we capture data from all five phones, we only use the data from the center camera for training and testing since it's likely to have the most accurate depth. We use the estimated camera poses for the center phone to rectify the stereo pair \cite{Fusiello00}. Specifically, we compute $\mathbf{W_l}$ and $\mathbf{W_r}$, i.e., warp maps corresponding to the left and and the right cameras that are applied to the RGB images, depth maps and confidences for the left and the right images. The camera poses are also used to convert depth into disparity between the recitifed pair. See Fig.~\ref{fig:gt} for examples of rectified RGB images and the corresponding ground truth disparity.

Further, since the telephoto camera has a smaller field of view than the main camera, we apply a center crop of size $448 \times 560$ to all rectified images to restrict ourselves to the area of overlap.

\subsection{Computing Occlusion Confidence $C^{\mathit{occ}}$}
\label{sec:gt_occ}

As mentioned in the main paper, we also compute $C^{\mathit{occ}}$, i.e., a per-pixel confidence  where the ground truth disparity is accurate but the pixel is occluded in the other camera. This allows us to evaluate and compare the methods in regions that are occluded in one of the stereo views.

To compute it, we first estimate the set of pixels in the right image that are in field of view of the left image but are occluded by an occluder. This can be estimated as:

\begin{equation}
\mathit{Occ}_r  =  \left\{
     (x, y) \quad s.t. \quad
     \begin{aligned}
       & 0 \leq x' < W, \\
       & |D^{gt}_r(x, y) + D^{gt}_l(x', y)| > \Delta \\
       & |D^{gt}_l(x', y) + D^{gt}_r(x' + D^{gt}_l(x', y), y)| \leq \Delta  \\
       & |D^{gt}_l(x', y)| > |D^{gt}_r(x, y)| 
     \end{aligned}
  \right\}
\end{equation}

where $x'=x + D^{gt}_r(x, y)$, $W$ is the width of the rectified image, and $\Delta$ is set to $1$-pixel disparity. The first condition enforces that the pixel is in field of view of the other (left) camera; the second condition ensures that the pixel is not visible in the other camera by checking for failed left-right consistency check; the final two conditions check that the pixel is occluded by an object that is visible in both the views (consistency check succeeds) and is in front of the occluded pixel. Finally, for pixels that are in the set $\mathit{Occ}_r$ we set $C^{\mathit{occ}}_r(x, y)$ to be the product of the confidences of the pixel and the occluder, i.e., 

\begin{equation}
C^{\mathit{occ}}_r(x, y)=
\begin{cases}
  C^{gt}_r(x, y)\cdot C^{gt}_l(x', y), & \text{if} \quad (x, y) \in \mathit{Occ}_r \\
  0, & \text{otherwise}
\end{cases}
\end{equation}

A few examples are shown in Fig.~\ref{fig:gt}. Our conservative criterion for occlusion confidence ensures that we have few false positives.

\section{Implementation Details}
In this section, we provide more details about confidence volume fusion, network architecture, and evaluation with affine fitting.

\subsection{Cost Volume and Confidence Volume}
In Sec. 4.2 of the main paper, we fuse confidence volume instead of the cost volume. Here we give more explanation and motivation.

The commonly used cost volume \cite{kendall2017end} is a 3D volume with two dimensions for the image space $(H,W)$ and one dimension for the disparity space $R=[0,1,\cdots,d_{max}]$. Each voxel $(x,y,d)$ is a floating number indicating the feature distance if the pixel $(x,y)$ in one view is matched under the given disparity $d$ with the other view. The distance can be computed using various distance metrics, such as $\ell_1$ or $\ell_2$.

A soft--$\arg\min$, which is introduced in Eq. 1 in \cite{kendall2017end}, is used to convert the cost volume into a disparity map.
Specifically, a confidence volume is calculated as the softmax on the negative cost volume along the disparity dimension, and output disparity is the sum of disparity hypotheses weighted by the confidence.
The operator to convert cost volume to confidence volume can also be written as:
\begin{equation}
\operatorname{Confidence}(x,y,d) = \frac{e^{-\operatorname{Cost}(x,y,d)/t}}{\sum_{d\in R} e^{-\operatorname{Cost}(x,y,d)/t}},
    \label{eq:conf}
\end{equation}
where $t$ controls the sharpness of the softmax and is set to $0.5$ in our implementation. The voxels along the disparity dimension for each $(x,y)$ forms a probability distribution, i.e., $\sum_{d\in R} \operatorname{Confidence}(x,y,d)=1$, indicating the likelihood of each disparity proposal in $R$ being correct (i.e. confidence).
Therefore, the output disparity is
\begin{equation}
    D(x,y) = \sum_{d\in R}\operatorname{Confidence}(x,y,d)\cdot d.
\end{equation}

\begin{figure*}[h]
    \centering
    \includegraphics[width=\textwidth]{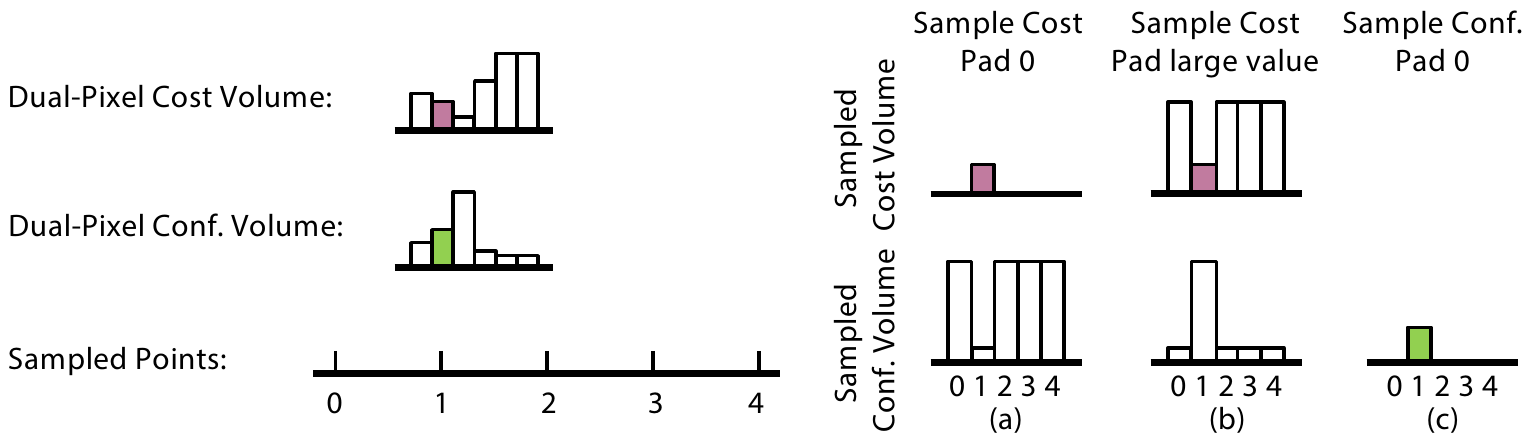}
    \caption{Explanation of Volume Sampling. See text for details.}
    \label{fig:cost-vs-conf}
\end{figure*}

We now explain why it is better to sample confidence volume (Eq. \ref{eq:conf}) instead of the cost volume.
Besides normalizing the scales of the two cost volumes, they may produce dramatically different results in some cases.

Fig. \ref{fig:cost-vs-conf} illustrates one such case. For simplicity, we drop the image dimension $(H,W)$ and only visualize the disparity dimension $R$ for one pixel.
On the left, we show an example of cost volume learned from DP inputs and the corresponding confidence volume. Note that the confidence is inversely related with the cost.
Since the DC inputs covers larger range of disparity compared to DP, the warping process in Eq. 3 of the main paper usually samples many points out of the DP disparity range.
In Fig. \ref{fig:cost-vs-conf}, we demonstrate the case where only one sample falls in the valid range of the DP disparity -- sampling point with disparity $1$.
If we sample cost volume and pad 0 for samples out of range, we obtain a cost volume shown in (a) on the right. The corresponding confidence volume (according to Eq. \ref{eq:conf}) indicates that $1$ is a bad disparity hypothesis and all the others are equally good, which is very inconsistent with the information provided in the original DP cost volume.
In the second case (b), we sample cost volume but pad with a large value. Now, the disparity hypothesis $1$ becomes the best hypothesis (unlike (a)) but the confidence in $1$ is much higher than the original confidence in DP confidence volume.
In contrast, if we sample the confidence volume and pad 0 as shown in (c), the produced confidence volume maintains exactly the same confidence from DP volume for disparity $1$, while the others are set to zero.

\subsection{Network Architecture}
\begin{figure}[hp]
\vspace{-0.5cm}
    \centering
    \includegraphics[height=19cm]{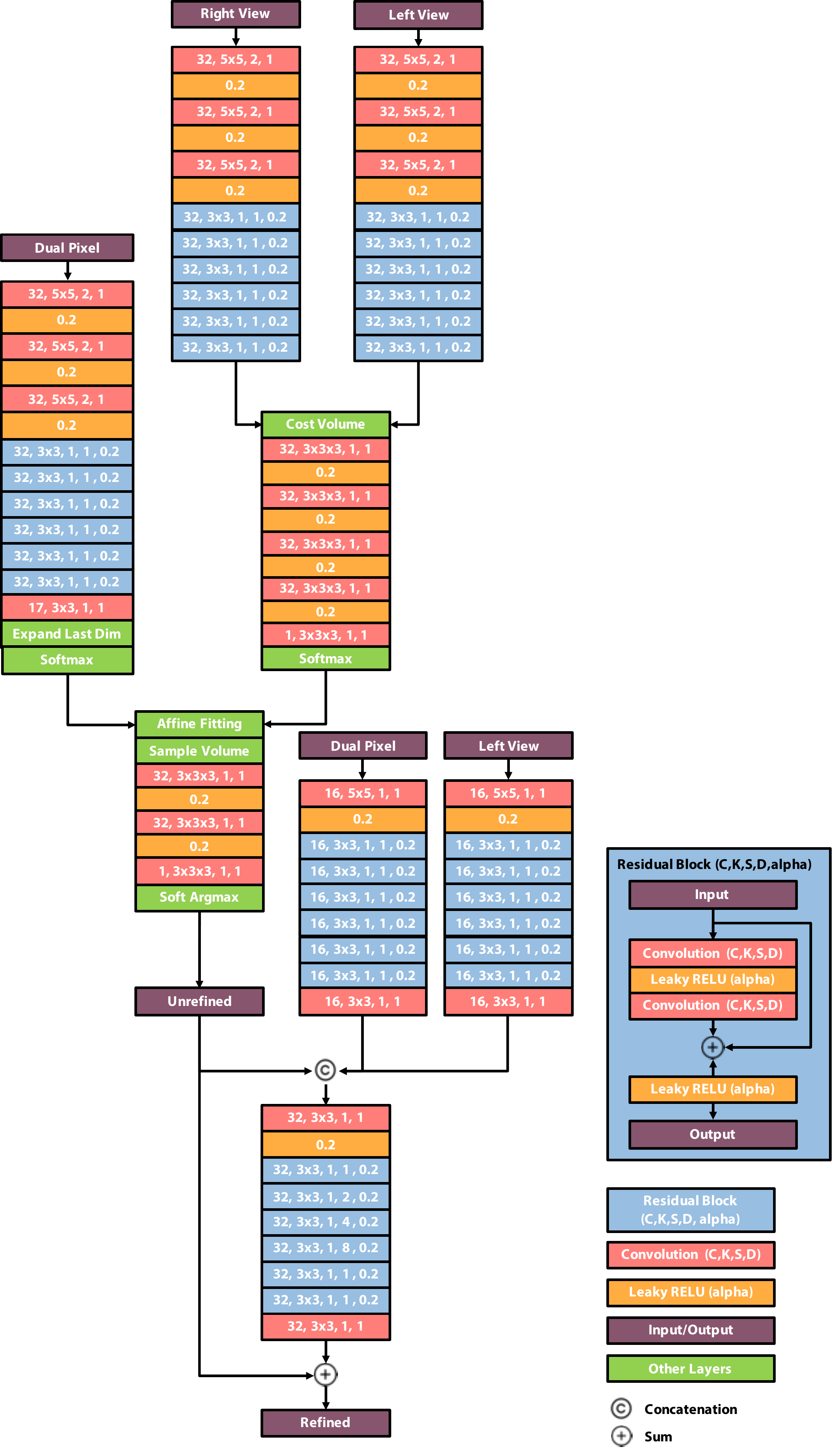}
\vspace{-0.5cm}
    \caption{Network Architecture. The numbers on convolution layers represent number of channels, size of filter, stride, and dilation respectively. The number on the leaky ReLU layer represents the slope for the negative input.}
    \label{fig:network}
\end{figure}

We provide detailed network architecture in Fig. \ref{fig:network}.

\subsection{Evaluation with Affine Fitting}
In the Sec. 5.4 of the main paper, we compared to dual-pixel based depth estimation solution DPNet \cite{garg2019learning}. Since depth from DP can only be predicted up to an unknown affine transformation, Garg \etal \cite{garg2019learning} first estimate the affine transformation by solving a weighted least squares problem using the ground truth:

\begin{equation}
    \hat{\alpha},\hat{\beta}=\underset{\alpha,\beta}{\operatorname{argmin}}\left\Vert C^{gt}\cdot((\alpha + \beta\cdot D_{raw})-D^{gt})\right\Vert ^{2},
\end{equation}
where $D_{raw}$ is the network output. $D_{\mathit{fit}}=\hat{\alpha}+\hat{\beta} \cdot D_{raw}$ is then used for computing the metrics.
Even though \algoname \ produces disparity in absolute scale and is free from this ambiguity, we apply the same post-processing when comparing to the DPNet for fairness (\algoname$^*$ in Tab. 2 of the main paper).
\section{More Experiment Results}
In this section, we provide more quantitative and qualitative evaluations.

\subsection{Weighted Metrics}
Our ground truth comes with a confidence mask (Sec. \ref{sec:gt_conf}), and we use it to calculate weighted evaluation metrics.
\begin{align}
    \operatorname{MAE} &= \frac{\sum_p|D(p)-D^{gt}(p)|\cdot C^{gt}(p)}{\sum_pC^{gt}(p)}, \\
   \operatorname{RMSE} &= \sqrt{\frac{\sum_p(D(p)-D^{gt}(p))^2 \cdot C^{gt}(p)}{\sum_pC^{gt}(p)}}, \\
 \delta>\epsilon &= \frac{\sum_p\mathds{1}(|D(p)-D^{gt}(p)|>\epsilon) \cdot C^{gt}(p)}{\sum_pC^{gt}(p)},
\end{align}
where $D$ is the predicted disparity, $D^{gt}$ is the ground truth disparity, $C^{gt}$ is the confidence map, $\mathds{1}$ is an indicator function which equals $1$ if the condition is true and $0$ otherwise, and $p$ is a pixel in the image.

\subsection{More Quantitative Ablation Study}
In the Sec. 5.3 of the main paper, we showed the quantitative evaluation of our method under different ablations on all pixels $C^{gt}$. We perform the same comparison on the occluded pixels using $C^{occ}$ (Sec. \ref{sec:gt_occ}) to show the performance in occluded regions.

Tab. \ref{tab:ablation-volume} shows the evaluation of the unrefined disparity (i.e. the output of the fused volume) under different fusion strategies.
Consistent with the conclusion drawn from all pixels, our method outperforms all the others on the occluded regions.

Tab. \ref{tab:ablation-refine} shows the evaluation of the refined disparity (i.e. the output of the refinement) under different settings.
Refinement with DP consistently outperform the case without DP on all the metrics.
Using only DP is better than using both under some metrics, which is reasonable since RGB may not be very helpful to recover details in the occluded region and may even hurt the valuable information encoded in DP.

\begin{table}[t]
    \centering
\resizebox{\linewidth}{!}{
\begin{tabular}{c||ccccc||ccccc}
\hline 
\multirow{2}{*}{Conf. Volume} & \multicolumn{5}{c||}{All Pixels} & \multicolumn{5}{c}{Occluded Pixels}\tabularnewline
\cline{2-11} \cline{3-11} \cline{4-11} \cline{5-11} \cline{6-11} \cline{7-11} \cline{8-11} \cline{9-11} \cline{10-11} \cline{11-11} 
 & MAE & RMSE & $\delta\!>\!1.25$ & $\delta\!>\!2$ & $\delta\!>\!3$ & MAE & RMSE &$\delta\!>\!1.25$ & $\delta\!>\!2$ & $\delta\!>\!3$\tabularnewline
\hline 
DC & 1.023 & 2.502 & 18.74 & 10.65 & 6.32 & 3.956 & 6.230 & 66.33 & 52.82 & 40.16\tabularnewline
DP+DC (2D) & 0.969 & 2.423 & 17.37 & 9.72 & 5.79 & 3.718 & 5.834 & 65.74 & 51.55 & 38.51\tabularnewline
DP+DC (C) & 0.964 & 2.372 & 17.51 & 9.79 & 5.80 & 3.671 & 5.768 & 65.49 & 51.63 & 38.62\tabularnewline
DP+DC (Ours) & \textbf{0.902} & \textbf{2.252} & \textbf{16.16} & \textbf{8.96} & \textbf{5.26} & \textbf{3.526} & \textbf{5.523} & \textbf{64.98} & \textbf{50.82} & \textbf{37.41}\tabularnewline
\hline 
\end{tabular}
}
\vspace{1mm}
    \caption{Ablation study on volume fusion. We compare different ways of fusing DP with the DC confidence volume. `(2D)' indicates fusion of the 2D disparity maps extracted from the two confidence volumes. `(C)' indicates fusing cost volumes instead of confidence volumes.}
    \label{tab:ablation-volume}
\vspace{-20pt}
\end{table}

\begin{table}[h]
    \centering
\resizebox{\linewidth}{!}{
\begin{tabular}{c||ccccc||ccccc}
\hline 
\multirow{2}{*}{Refinement} & \multicolumn{5}{c||}{All Pixels} & \multicolumn{5}{c}{Occluded Pixels}\tabularnewline
\cline{2-11} \cline{3-11} \cline{4-11} \cline{5-11} \cline{6-11} \cline{7-11} \cline{8-11} \cline{9-11} \cline{10-11} \cline{11-11} 
 & MAE & RMSE & $\delta\!>\!1.25$ & $\delta\!>\!2$ & $\delta\!>\!3$ & MAE & RMSE &$\delta\!>\!1.25$ & $\delta\!>\!2$ & $\delta\!>\!3$\tabularnewline
\hline 
RGB & 0.838 & 2.197 & 14.17 & 7.74 & 4.55 & 2.627 & 4.866 & 46.17 & 33.70 & 24.18\tabularnewline
DP (I) & 0.835 & 2.173 & 14.25 & 7.75 & 4.54 & 2.518 & 4.600 & 46.07 & 33.25 & 23.47\tabularnewline
DP & 0.829 & 2.184 & 13.84 & 7.51 & 4.45 & 2.481 & 4.619 & \textbf{44.87} & \textbf{31.99} & 22.54\tabularnewline
RGB+DP (Ours) & \textbf{0.817} & \textbf{2.141} & \textbf{13.64} & \textbf{7.33} & \textbf{4.35} & \textbf{2.469} & \textbf{4.564} & 44.94 & 32.09 & \textbf{22.47}\tabularnewline
\hline 
\end{tabular}
}
\vspace{1mm}
    \caption{Ablation study on refinement. We compare the different ways of using DP to refine the best unrefined disparity from the left and show evaluation on the final disparity. `(I)' indicates that input DP images are warped before computing features for refinement.}
    \label{tab:ablation-refine}
\vspace{-20pt}
\end{table}

\subsection{More Qualitative Ablation Study}
We show more qualitative comparison in Fig. \ref{fig:supp_ablation}. Our method fusing DP into the cost volume (d) significantly improves the quality of the unrefined disparity compared to the case using only DC (c). 
Based on this improved unrefined disparity with less error (d), refinement using both DP and DC (f) can further improve the object boundary and thin structures compared to the case using only DC (e).

\newcommand{\ablationresultswidth}{0.16\textwidth}{}
\begin{figure}[hp]
    \centering
    \begin{tabular}{@{}c@{\,\,}c@{\,\,}c@{\,\,}c@{\,\,}c@{\,\,}c@{}}
\includegraphics[width=\ablationresultswidth]{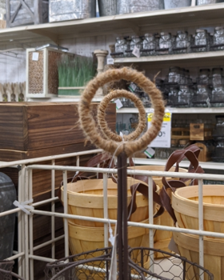} &
\includegraphics[width=\ablationresultswidth]{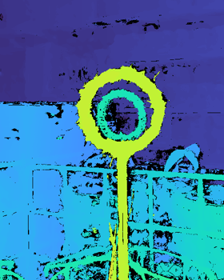} &
\includegraphics[width=\ablationresultswidth]{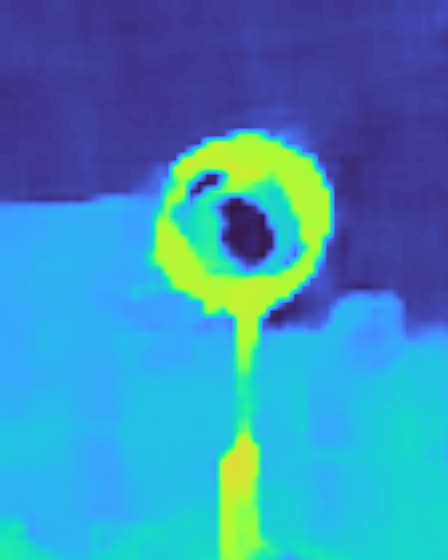} &
\includegraphics[width=\ablationresultswidth]{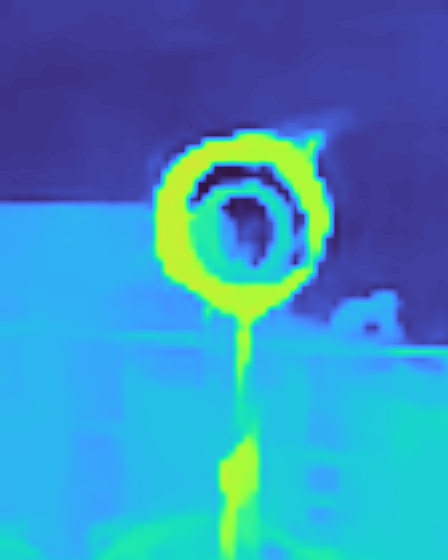} &
\includegraphics[width=\ablationresultswidth]{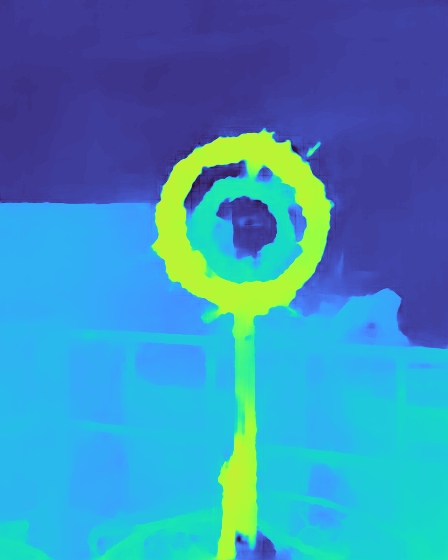} &
\includegraphics[width=\ablationresultswidth]{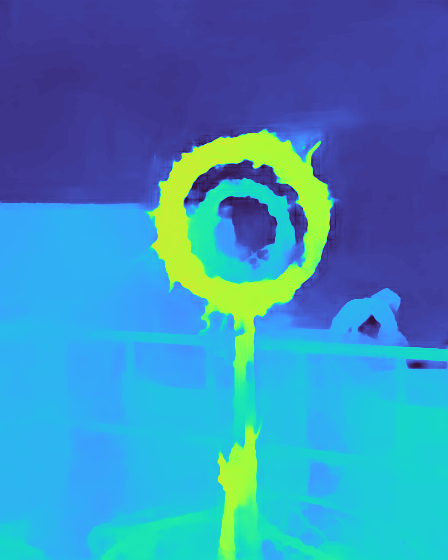}
\\
\includegraphics[width=\ablationresultswidth]{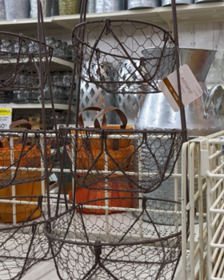} &
\includegraphics[width=\ablationresultswidth]{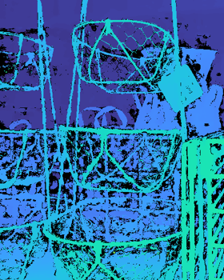} &
\includegraphics[width=\ablationresultswidth]{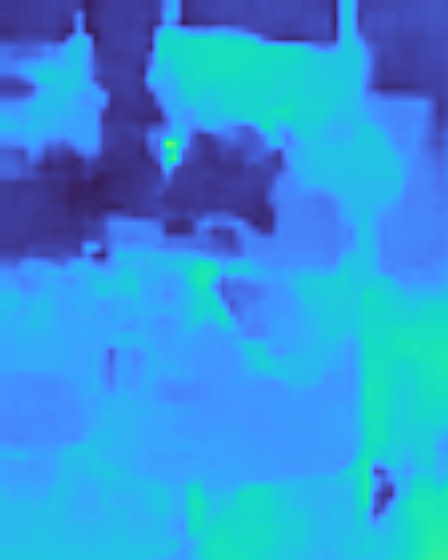} &
\includegraphics[width=\ablationresultswidth]{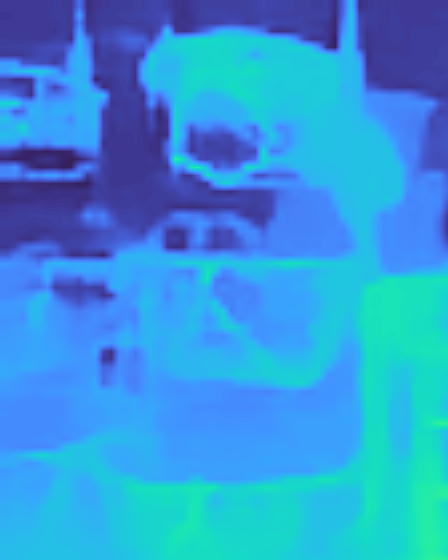} &
\includegraphics[width=\ablationresultswidth]{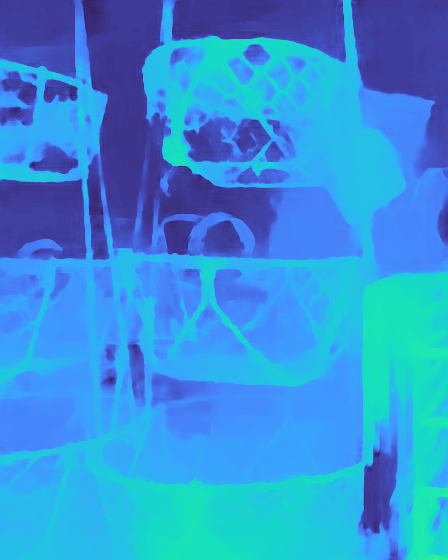} &
\includegraphics[width=\ablationresultswidth]{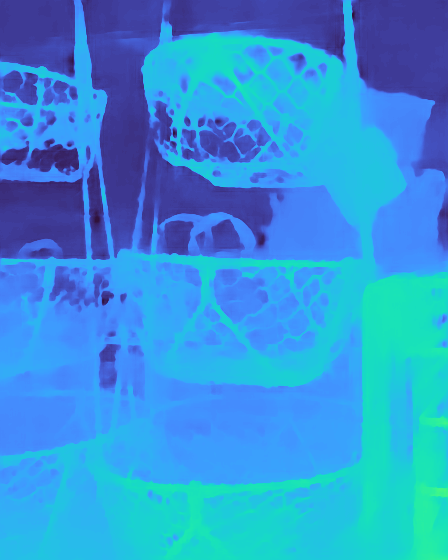}
\\
\includegraphics[width=\ablationresultswidth]{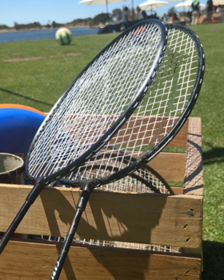} &
\includegraphics[width=\ablationresultswidth]{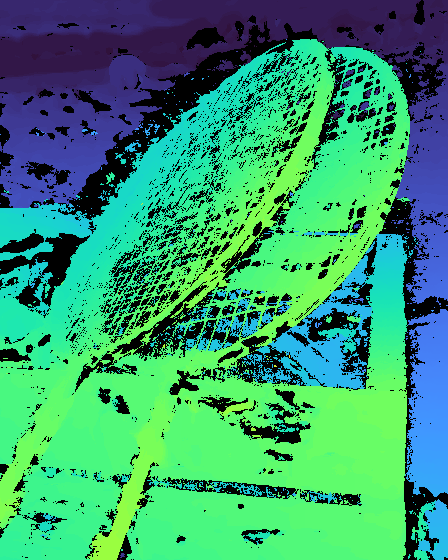} &
\includegraphics[width=\ablationresultswidth]{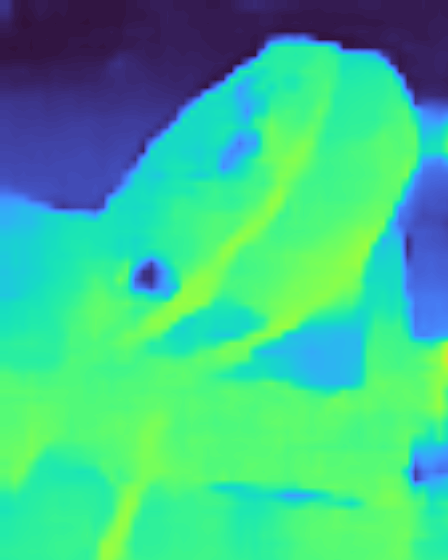} &
\includegraphics[width=\ablationresultswidth]{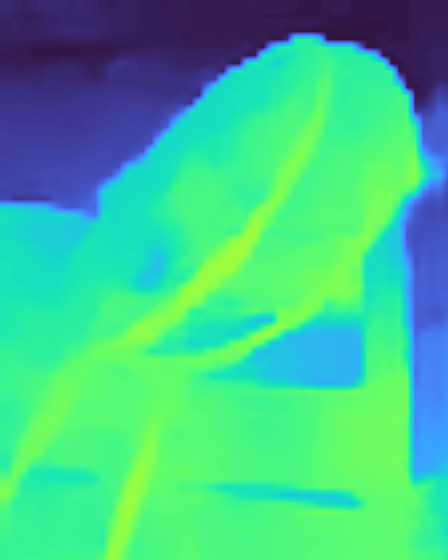} &
\includegraphics[width=\ablationresultswidth]{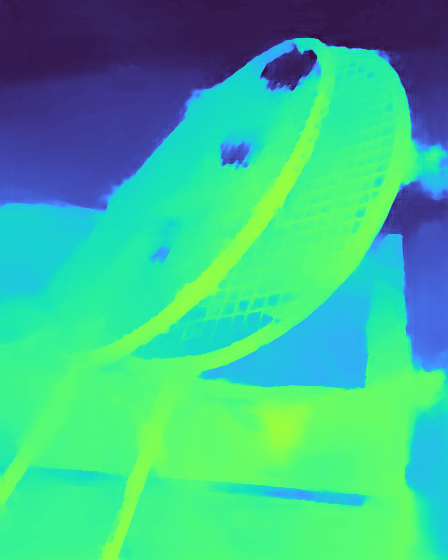} &
\includegraphics[width=\ablationresultswidth]{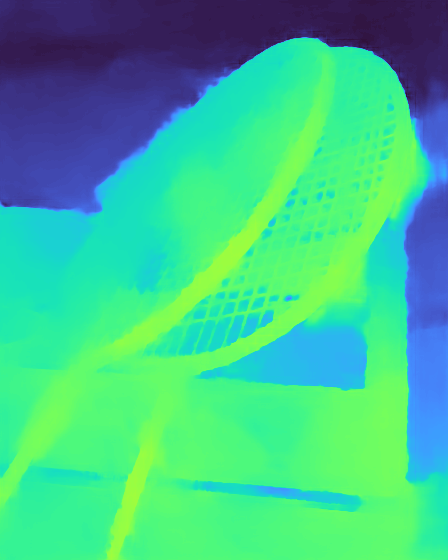}
\\
\includegraphics[width=\ablationresultswidth]{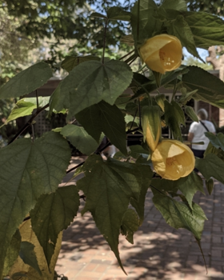} &
\includegraphics[width=\ablationresultswidth]{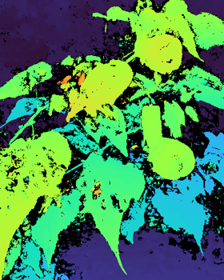} &
\includegraphics[width=\ablationresultswidth]{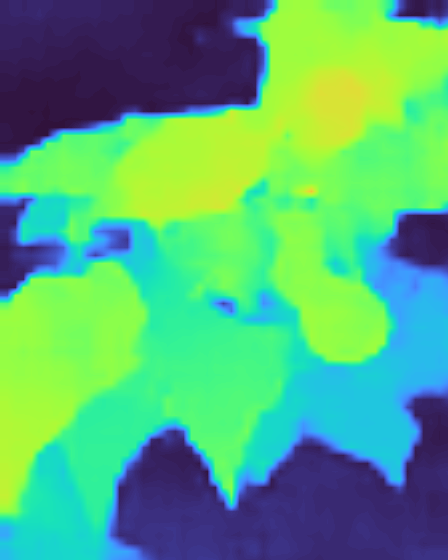} &
\includegraphics[width=\ablationresultswidth]{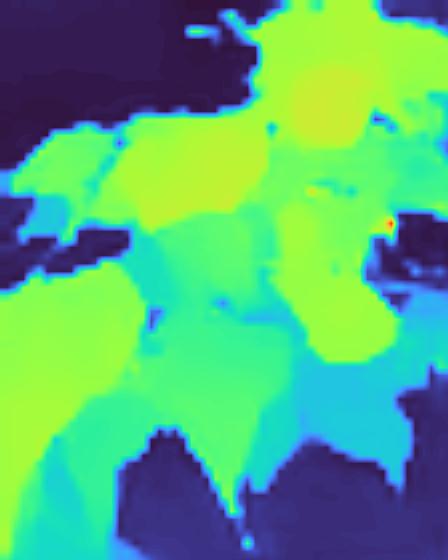} &
\includegraphics[width=\ablationresultswidth]{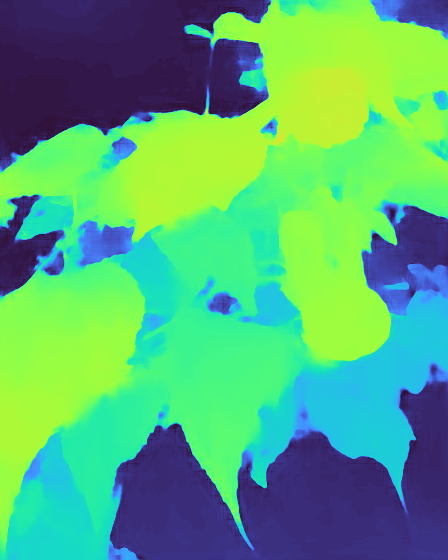} &
\includegraphics[width=\ablationresultswidth]{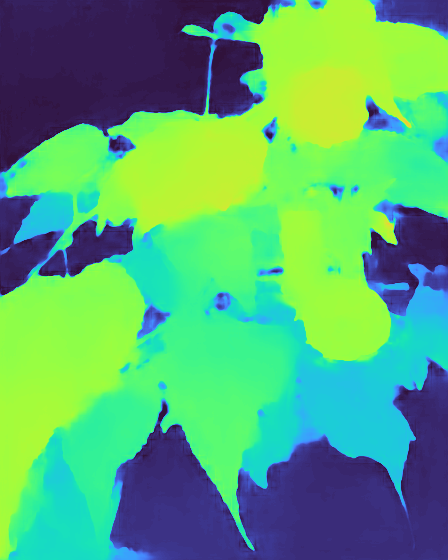}
\\
\includegraphics[width=\ablationresultswidth]{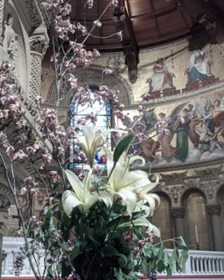} &
\includegraphics[width=\ablationresultswidth]{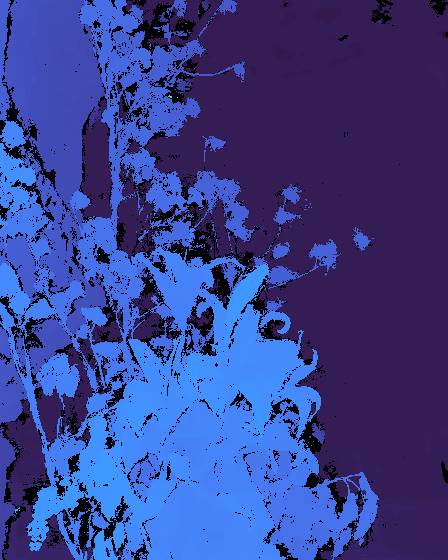} &
\includegraphics[width=\ablationresultswidth]{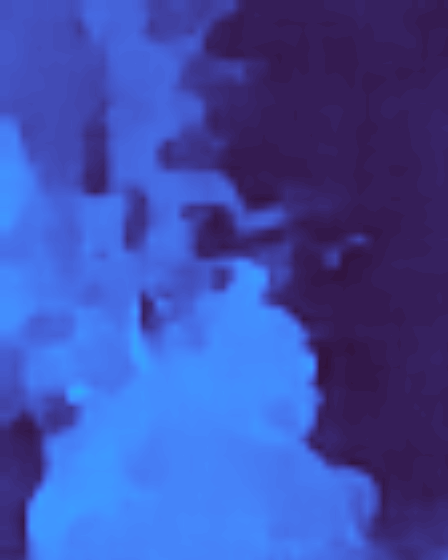} &
\includegraphics[width=\ablationresultswidth]{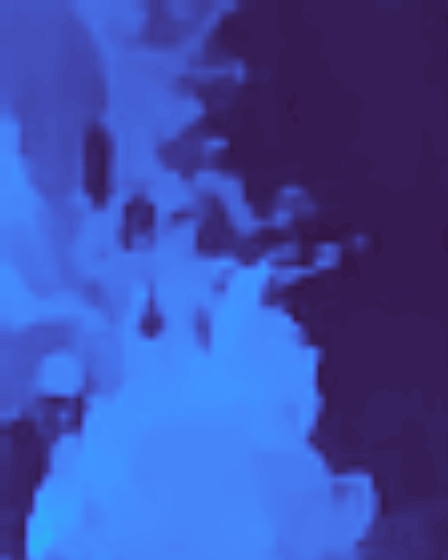} &
\includegraphics[width=\ablationresultswidth]{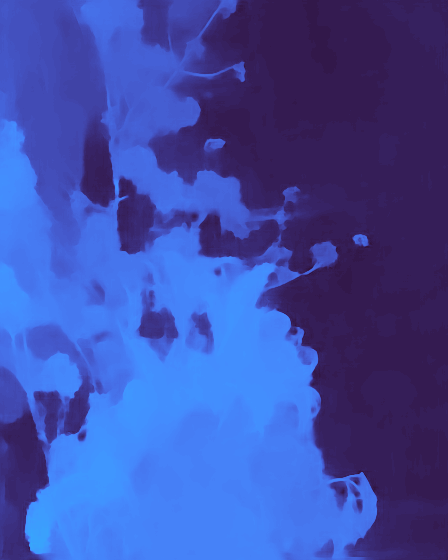} &
\includegraphics[width=\ablationresultswidth]{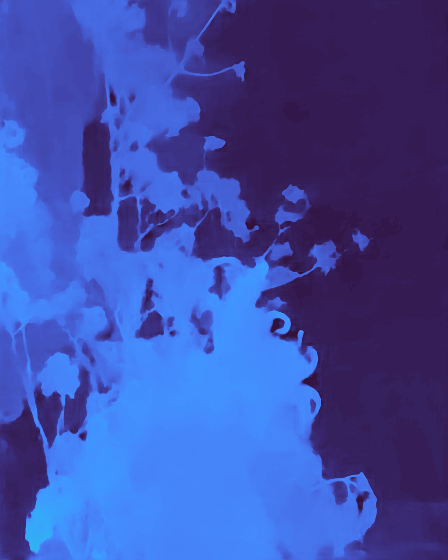}
\\
    \fakecaption{(a) \,Image} & \fakecaption{(b)\,GT} & \fakecaption{(c)\,$D_{\mathit{unref}}$} & \fakecaption{(d)\,$D_{\mathit{unref}}$} & \fakecaption{(e)\,$D_{\mathit{ref}}$} & \fakecaption{(f)\,$D_{\mathit{ref}}$}\\
    \addlinespace[-1.2ex]
    \fakecaption{} & \fakecaption{} & \fakecaption{DC, RGB+DP} & \fakecaption{\algoname} & \fakecaption{DP+DC,RGB} & \fakecaption{\algoname}\\
    \addlinespace[-2.2ex]
    \end{tabular}
    \caption{Ablations of our method. The right camera image (a), ground truth disparity (b) with low confidence disparity in black, $D_{\mathit{unref}}$ (c) from an ablation where only the DC input is used for the confidence volume, $D_{\mathit{unref}}$ (d) from \algoname, $D_{\mathit{ref}}$ (e) from an ablation where only the RGB image is used for refinement, and $D_{\mathit{ref}}$ (f) from \algoname. DP input is useful for both the confidence volume and refinement stages to recover accurate depth for fine structures and occluded regions.}
    \label{fig:supp_ablation}
    \vspace{-15pt}
\end{figure}

\subsection{More Qualitative Comparison to SOTA}
We show more qualitative comparison to state-of-the-art stereo and DP based approaches in Fig. \ref{fig:gallery_1}, \ref{fig:gallery_2}, and \ref{fig:gallery_dp_worse}. Compared to other stereo based approaches \cite{khamis2018stereonet,psmnet} that only take DC as the input, our method performs better at object boundary and thin structures. Compared to dual-pixel only approach \cite{garg2019learning}, our method produces significantly better depth for distant areas in the background while maintaining the foreground details (Fig. \ref{fig:gallery_dp_worse}). 

\newcommand{\sotaresultswidth}{0.16\textwidth}{}
\begin{figure}[tp]
    \centering
    \begin{tabular}{@{}c@{\,\,}c@{\,\,}c@{\,\,}c@{\,\,}c@{\,\,}c@{}}
\includegraphics[width=\sotaresultswidth]{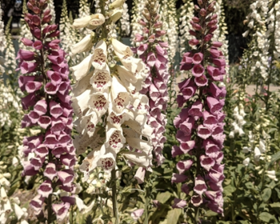} &
\includegraphics[width=\sotaresultswidth]{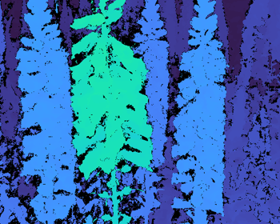} &
\includegraphics[width=\sotaresultswidth]{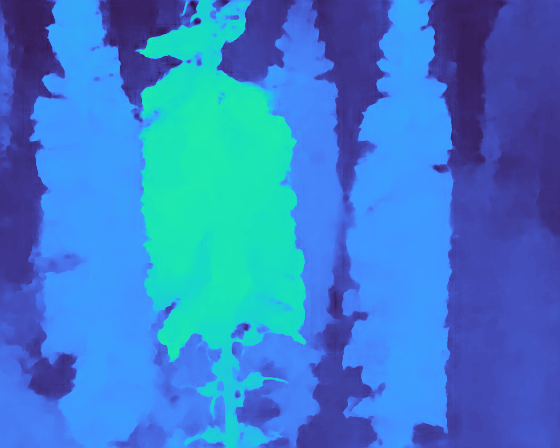} &
\includegraphics[width=\sotaresultswidth]{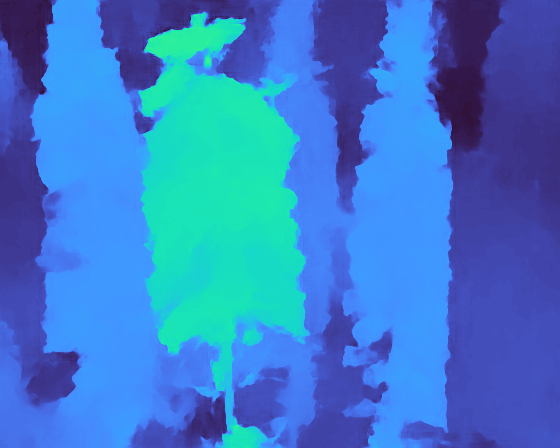} &
\includegraphics[width=\sotaresultswidth]{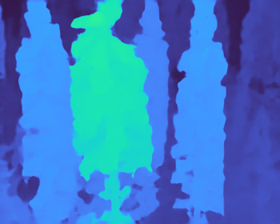} &
\includegraphics[width=\sotaresultswidth]{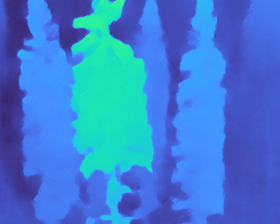}
\\
\includegraphics[width=\sotaresultswidth]{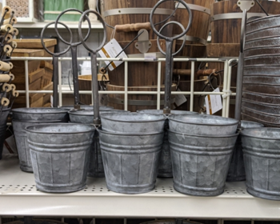} &
\includegraphics[width=\sotaresultswidth]{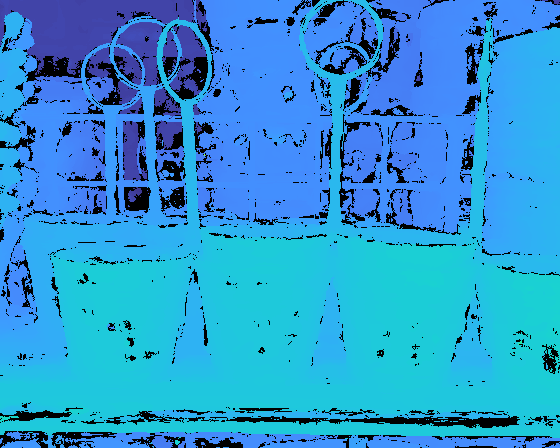} &
\includegraphics[width=\sotaresultswidth]{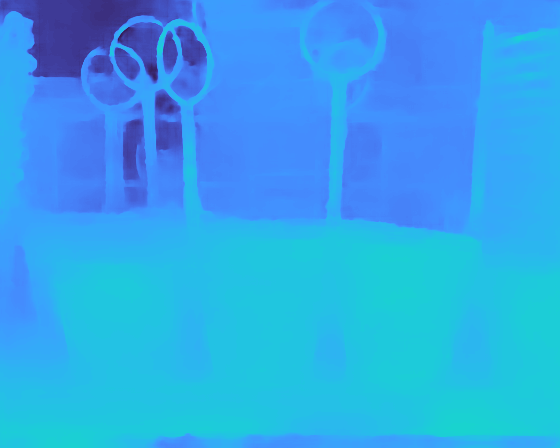} &
\includegraphics[width=\sotaresultswidth]{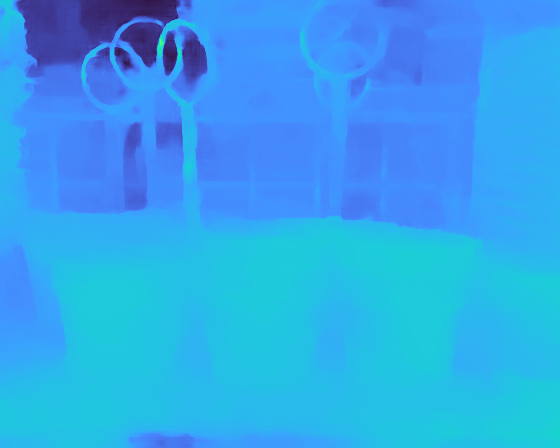} &
\includegraphics[width=\sotaresultswidth]{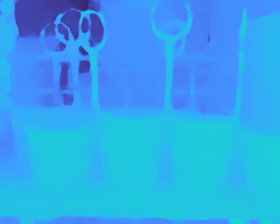} &
\includegraphics[width=\sotaresultswidth]{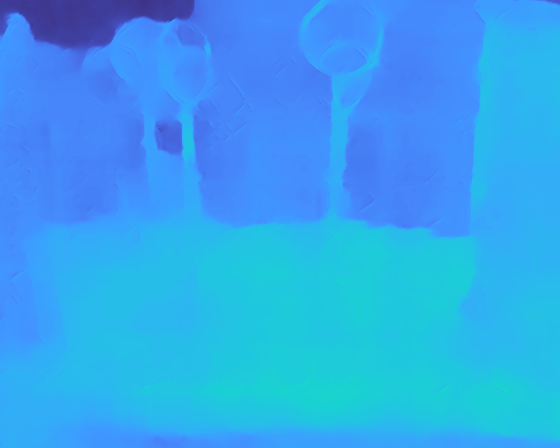}
\\
\includegraphics[width=\sotaresultswidth]{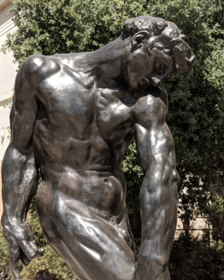} &
\includegraphics[width=\sotaresultswidth]{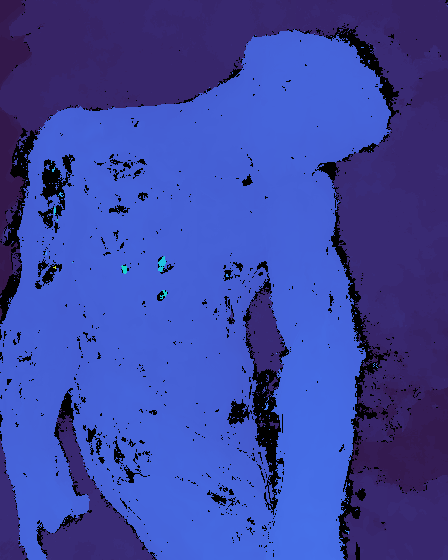} &
\includegraphics[width=\sotaresultswidth]{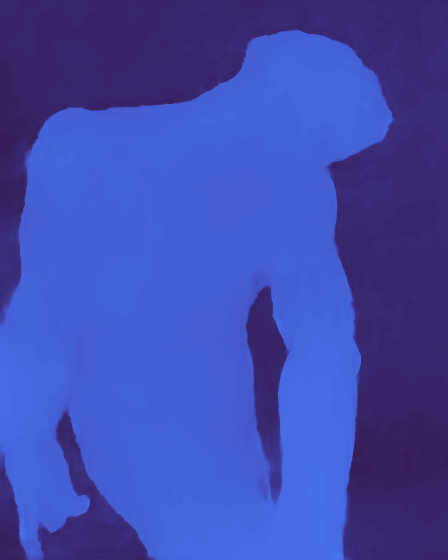} &
\includegraphics[width=\sotaresultswidth]{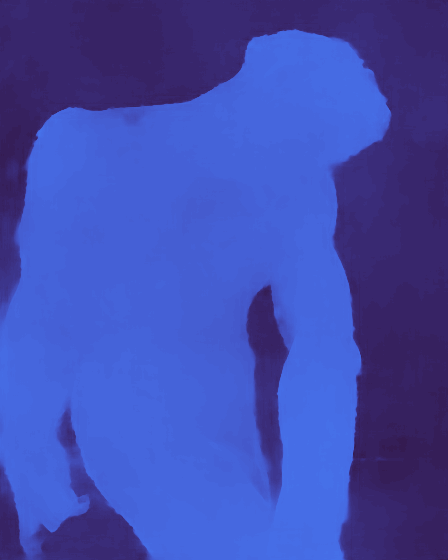} &
\includegraphics[width=\sotaresultswidth]{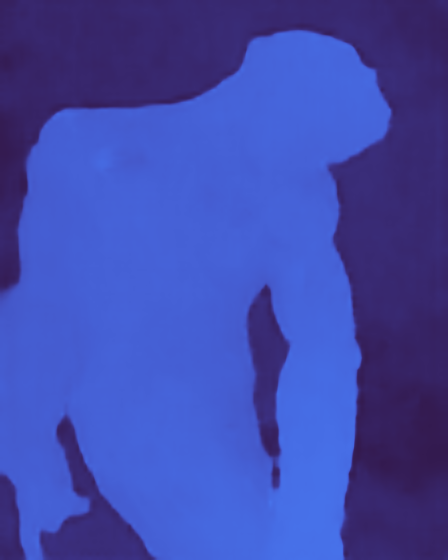} &
\includegraphics[width=\sotaresultswidth]{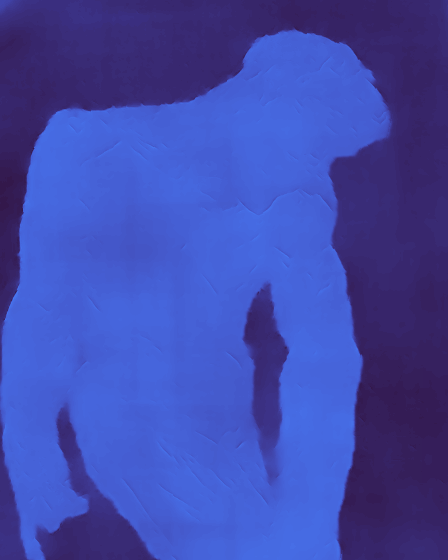}
\\
\includegraphics[width=\sotaresultswidth]{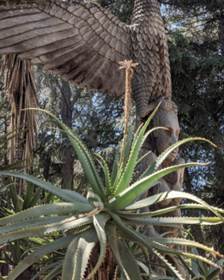} &
\includegraphics[width=\sotaresultswidth]{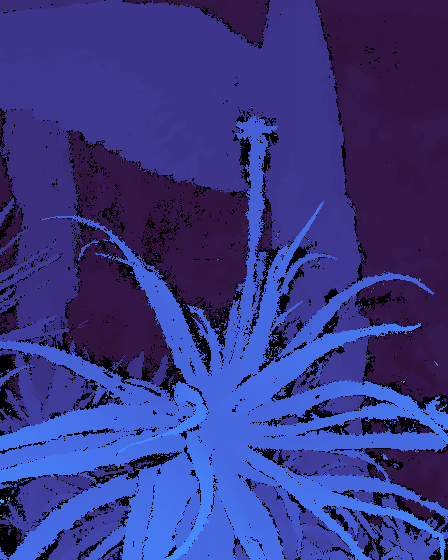} &
\includegraphics[width=\sotaresultswidth]{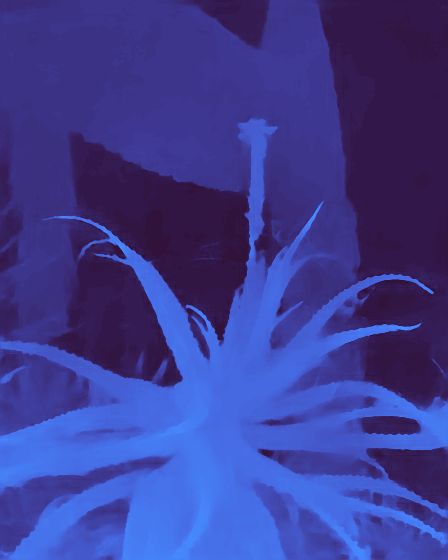} &
\includegraphics[width=\sotaresultswidth]{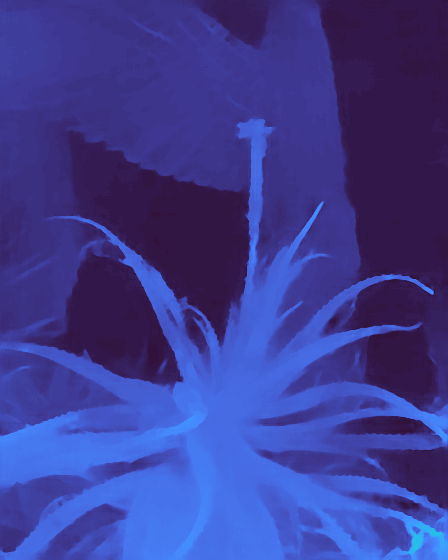} &
\includegraphics[width=\sotaresultswidth]{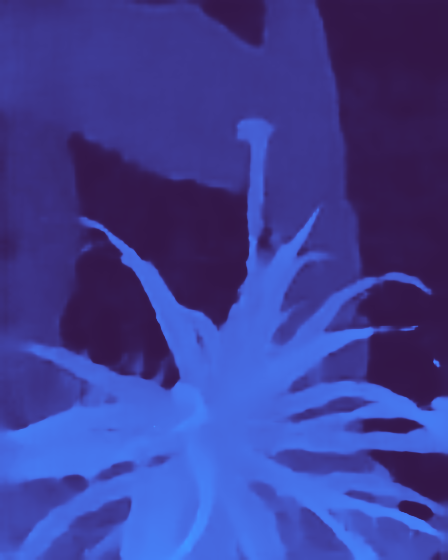} &
\includegraphics[width=\sotaresultswidth]{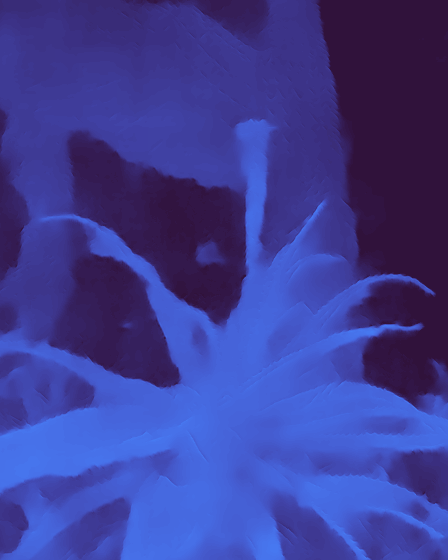}
\\
\includegraphics[width=\sotaresultswidth]{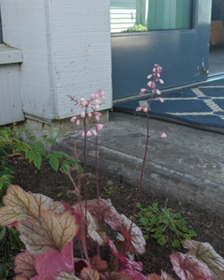} &
\includegraphics[width=\sotaresultswidth]{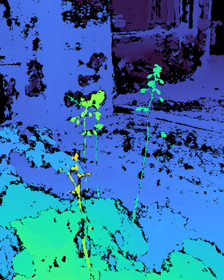} &
\includegraphics[width=\sotaresultswidth]{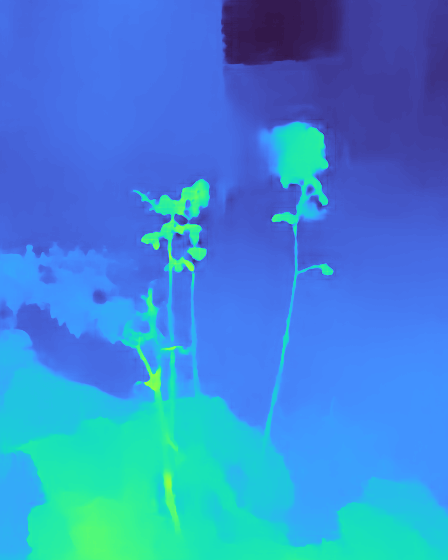} &
\includegraphics[width=\sotaresultswidth]{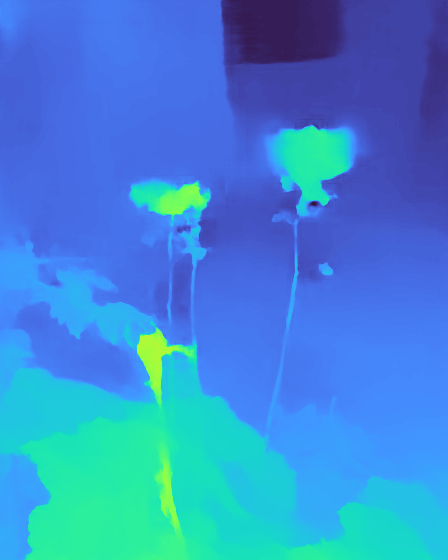} &
\includegraphics[width=\sotaresultswidth]{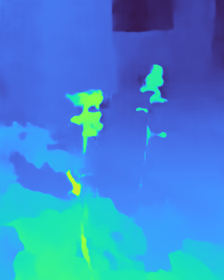} &
\includegraphics[width=\sotaresultswidth]{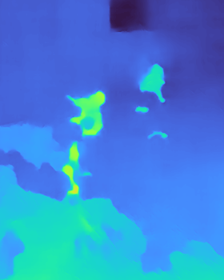}
\\
\includegraphics[width=\sotaresultswidth]{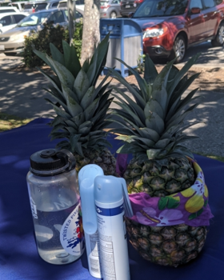} &
\includegraphics[width=\sotaresultswidth]{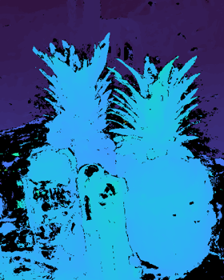} &
\includegraphics[width=\sotaresultswidth]{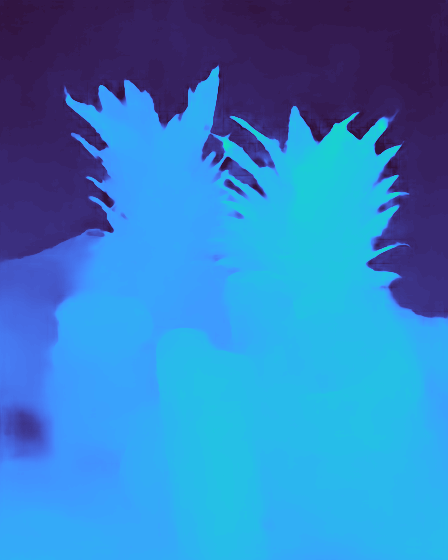} &
\includegraphics[width=\sotaresultswidth]{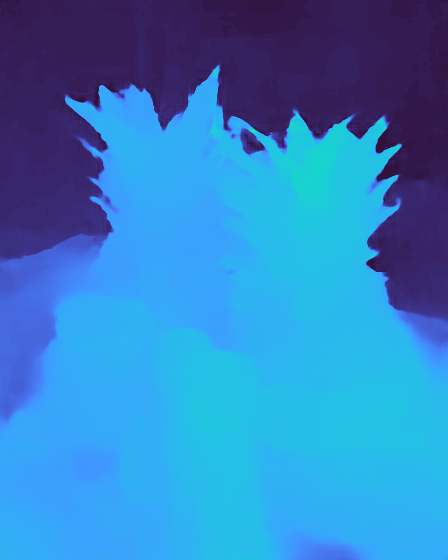} &
\includegraphics[width=\sotaresultswidth]{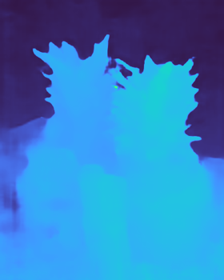} &
\includegraphics[width=\sotaresultswidth]{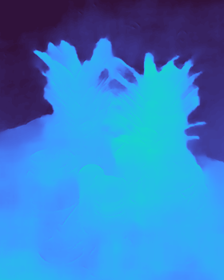}
\\
    \fakecaption{(a) \,Image} & \fakecaption{(b)\,GT} & \fakecaption{(c)\,Ours} & \fakecaption{(d)\,StereoNet} & \fakecaption{(e)\,PSMNet} & \fakecaption{(f)\,DPNet}\\
    \end{tabular}
    \caption{Qualitative comparison to state-of-the-art stereo and DP based methods.}
    \label{fig:gallery_1}
    \vspace{-10pt}
\end{figure}

\begin{figure}[tp]
    \centering
    \begin{tabular}{@{}c@{\,\,}c@{\,\,}c@{\,\,}c@{\,\,}c@{\,\,}c@{}}
\includegraphics[width=\sotaresultswidth]{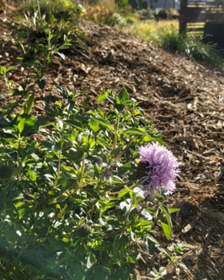} &
\includegraphics[width=\sotaresultswidth]{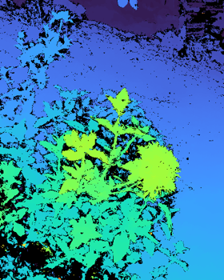} &
\includegraphics[width=\sotaresultswidth]{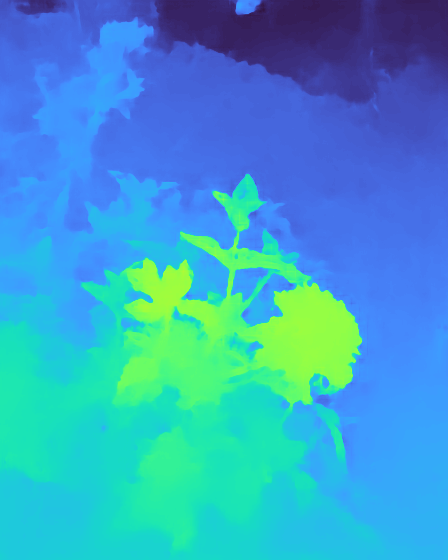} &
\includegraphics[width=\sotaresultswidth]{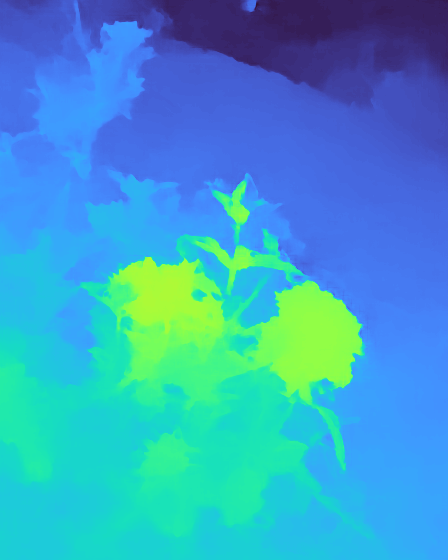} &
\includegraphics[width=\sotaresultswidth]{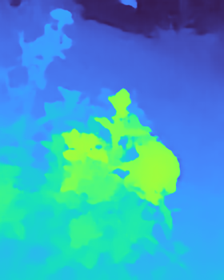} &
\includegraphics[width=\sotaresultswidth]{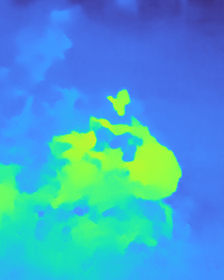}
\\
\includegraphics[width=\sotaresultswidth]{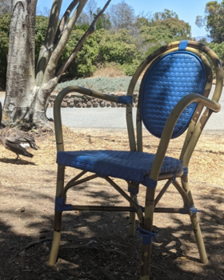} &
\includegraphics[width=\sotaresultswidth]{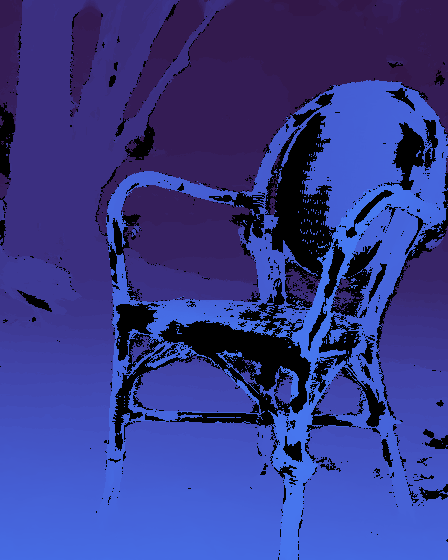} &
\includegraphics[width=\sotaresultswidth]{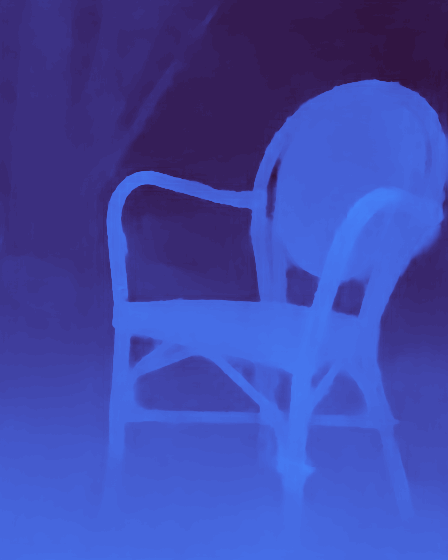} &
\includegraphics[width=\sotaresultswidth]{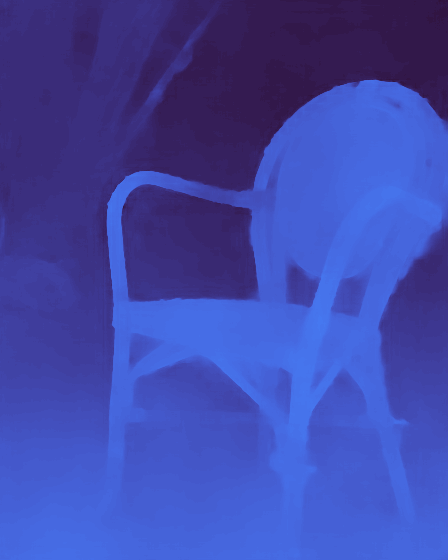} &
\includegraphics[width=\sotaresultswidth]{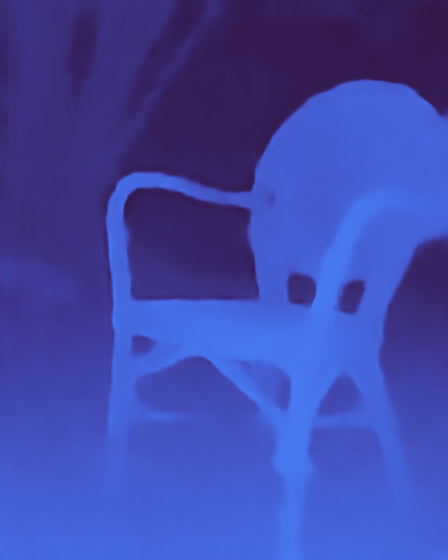} &
\includegraphics[width=\sotaresultswidth]{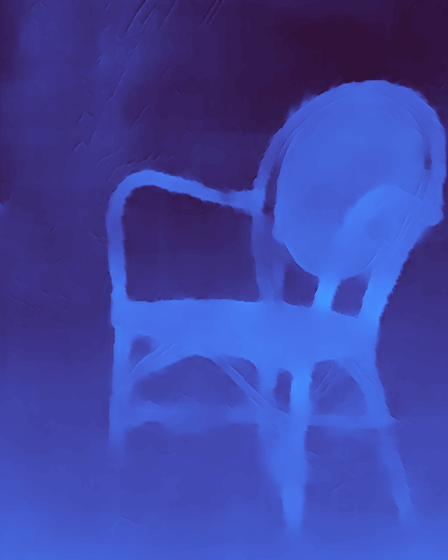}
\\
\includegraphics[width=\sotaresultswidth]{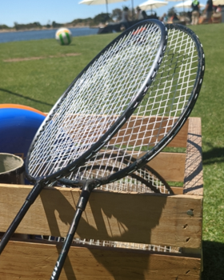} &
\includegraphics[width=\sotaresultswidth]{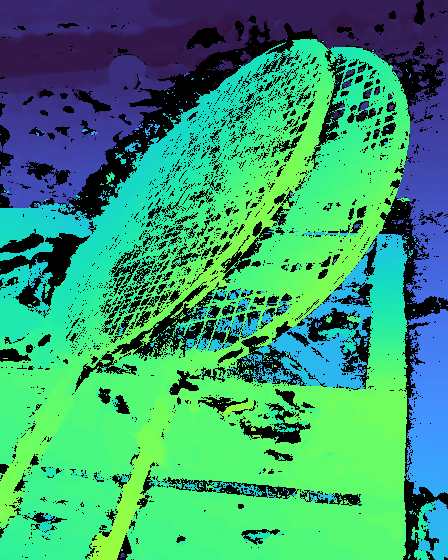} &
\includegraphics[width=\sotaresultswidth]{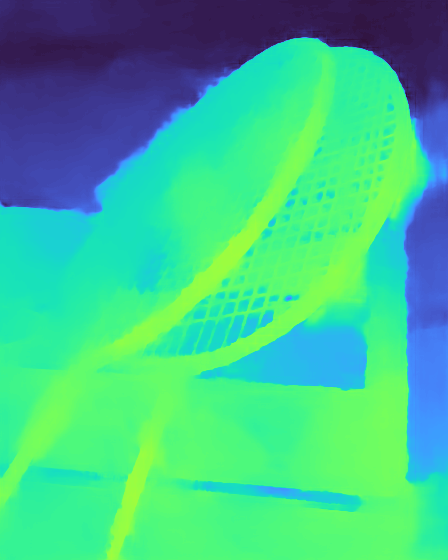} &
\includegraphics[width=\sotaresultswidth]{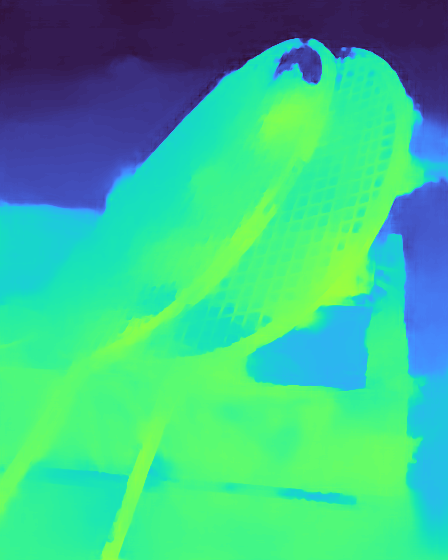} &
\includegraphics[width=\sotaresultswidth]{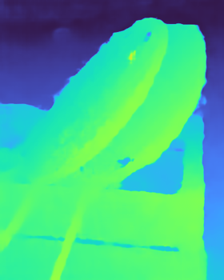} &
\includegraphics[width=\sotaresultswidth]{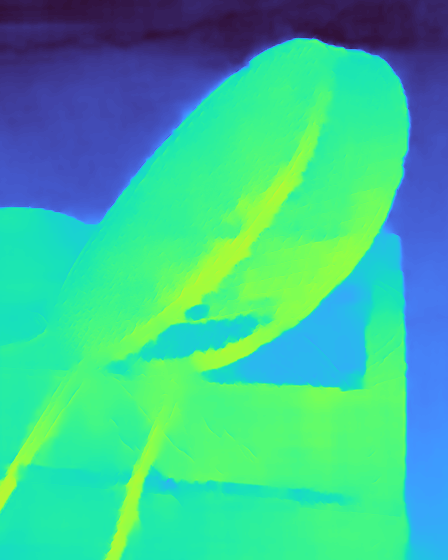}
\\
\includegraphics[width=\sotaresultswidth]{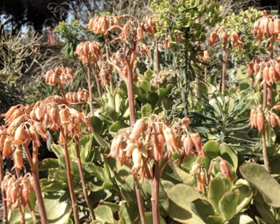} &
\includegraphics[width=\sotaresultswidth]{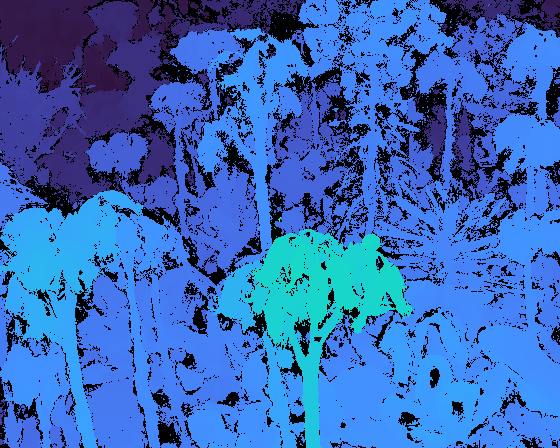} &
\includegraphics[width=\sotaresultswidth]{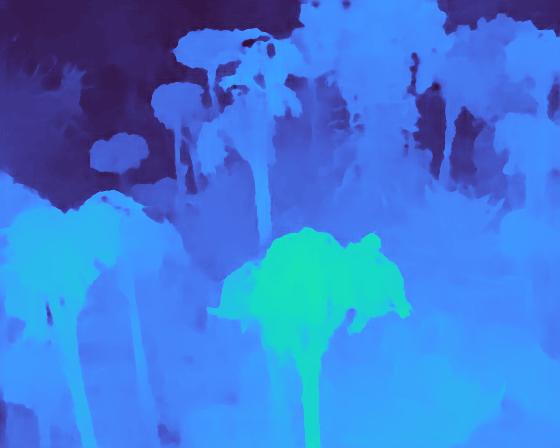} &
\includegraphics[width=\sotaresultswidth]{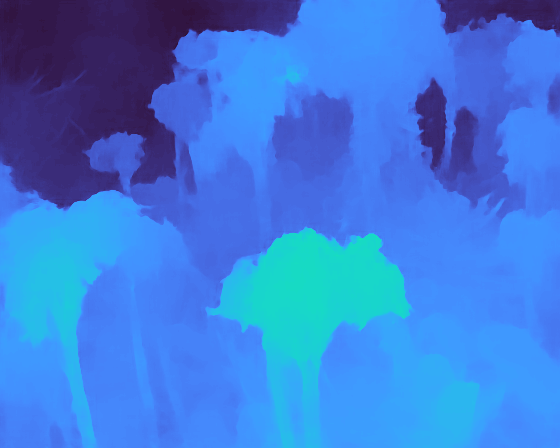} &
\includegraphics[width=\sotaresultswidth]{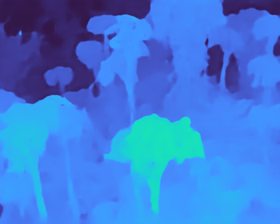} &
\includegraphics[width=\sotaresultswidth]{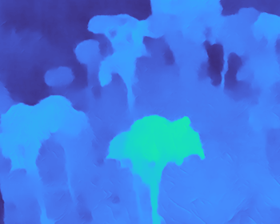}
\\
\includegraphics[width=\sotaresultswidth]{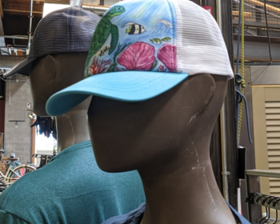} &
\includegraphics[width=\sotaresultswidth]{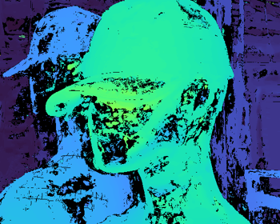} &
\includegraphics[width=\sotaresultswidth]{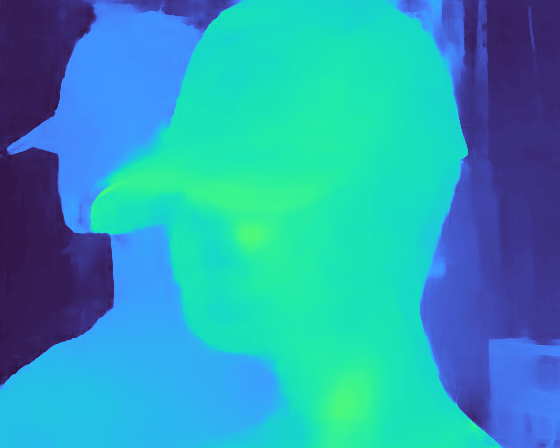} &
\includegraphics[width=\sotaresultswidth]{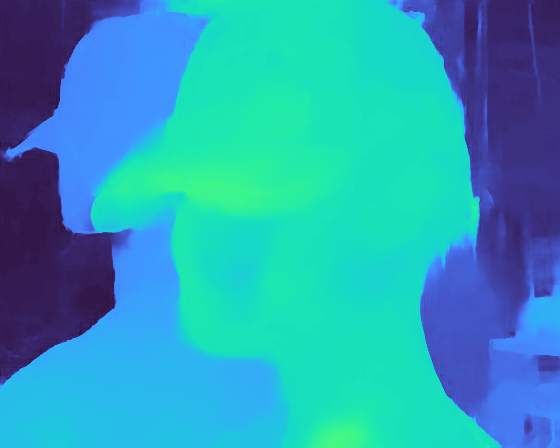} &
\includegraphics[width=\sotaresultswidth]{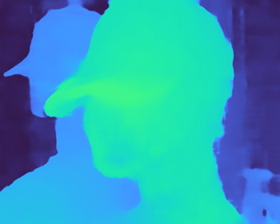} &
\includegraphics[width=\sotaresultswidth]{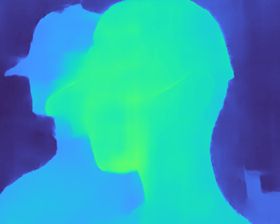}
\\
\includegraphics[width=\sotaresultswidth]{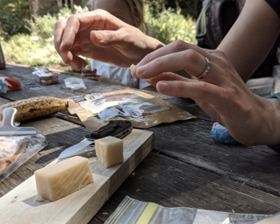} &
\includegraphics[width=\sotaresultswidth]{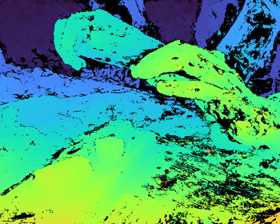} &
\includegraphics[width=\sotaresultswidth]{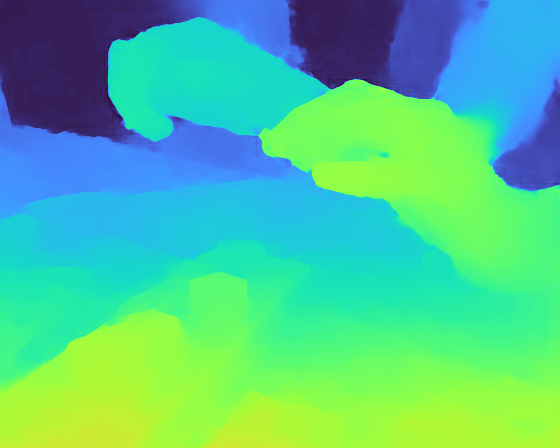} &
\includegraphics[width=\sotaresultswidth]{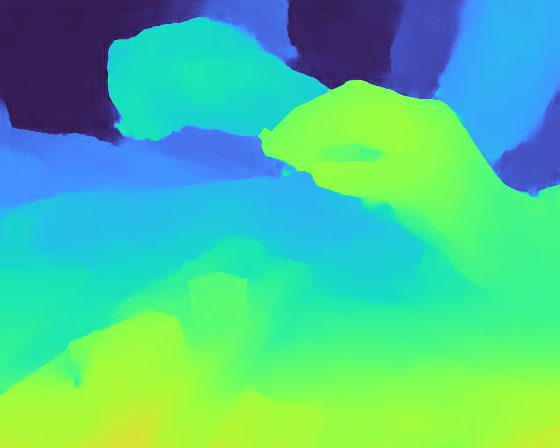} &
\includegraphics[width=\sotaresultswidth]{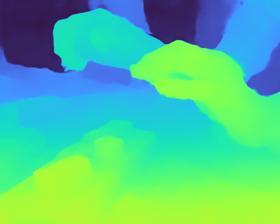} &
\includegraphics[width=\sotaresultswidth]{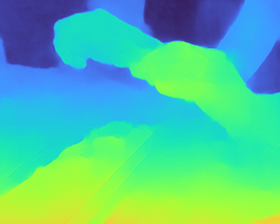}
\\
\includegraphics[width=\sotaresultswidth]{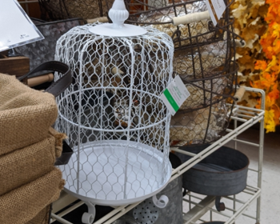} &
\includegraphics[width=\sotaresultswidth]{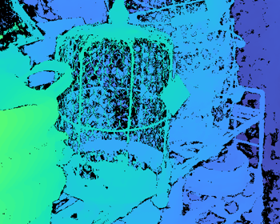} &
\includegraphics[width=\sotaresultswidth]{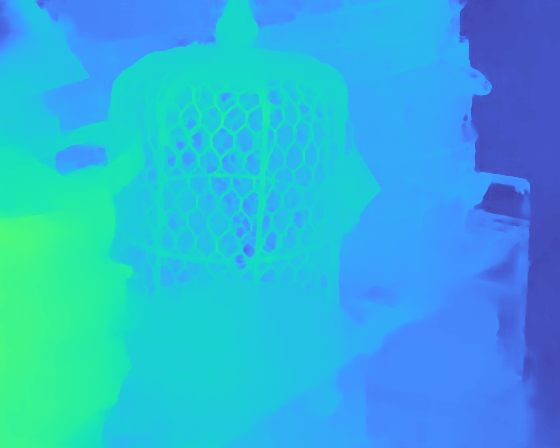} &
\includegraphics[width=\sotaresultswidth]{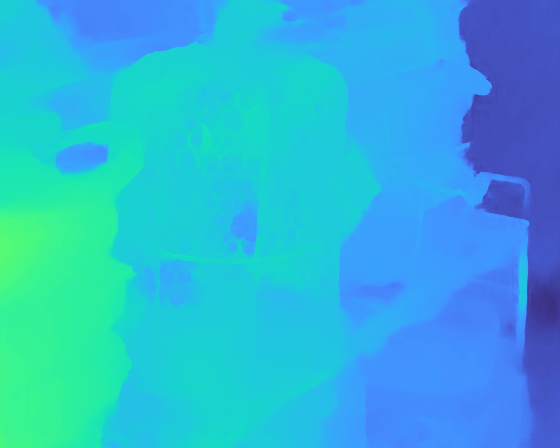} &
\includegraphics[width=\sotaresultswidth]{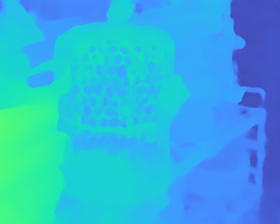} &
\includegraphics[width=\sotaresultswidth]{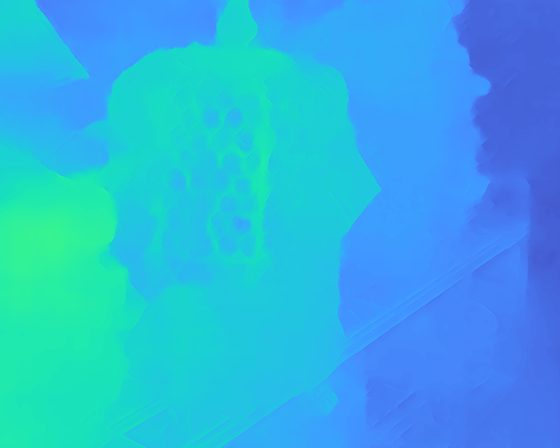}
\\
    \fakecaption{(a) \,Image} & \fakecaption{(b)\,GT} & \fakecaption{(c)\,Ours} & \fakecaption{(d)\,StereoNet} & \fakecaption{(e)\,PSMNet} & \fakecaption{(f)\,DPNet}\\
    \end{tabular}
    \caption{Qualitative comparison to state-of-the-art stereo and DP based methods.}
    \label{fig:gallery_2}
    \vspace{-10pt}
\end{figure}

\begin{figure}[tp]
    \centering
    \begin{tabular}{@{}c@{\,\,}c@{\,\,}c@{\,\,}c@{\,\,}c@{\,\,}c@{}}
\includegraphics[width=\sotaresultswidth]{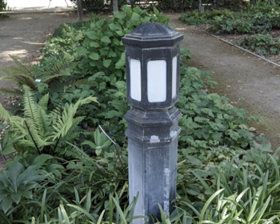} &
\includegraphics[width=\sotaresultswidth]{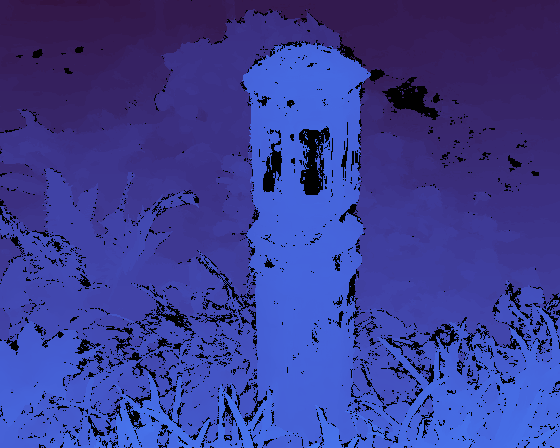} &
\includegraphics[width=\sotaresultswidth]{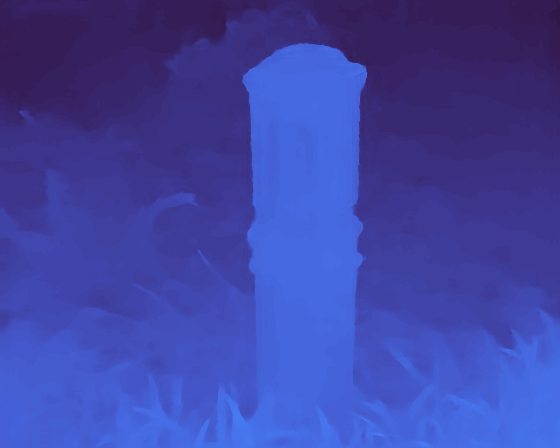} &
\includegraphics[width=\sotaresultswidth]{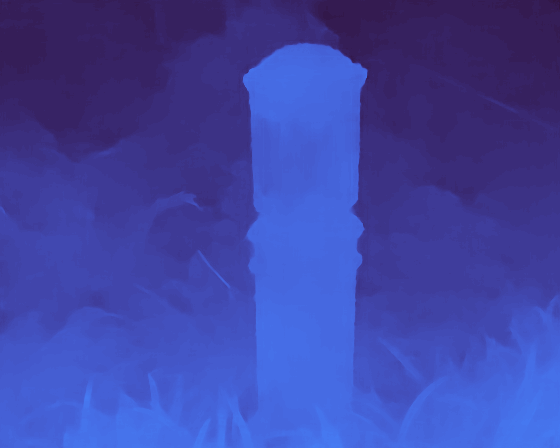} &
\includegraphics[width=\sotaresultswidth]{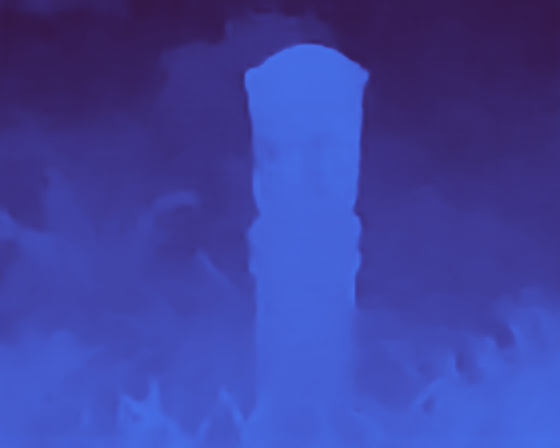} &
\includegraphics[width=\sotaresultswidth]{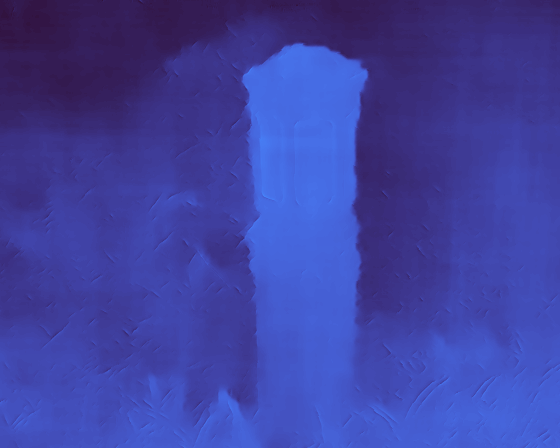}
\\
\includegraphics[width=\sotaresultswidth]{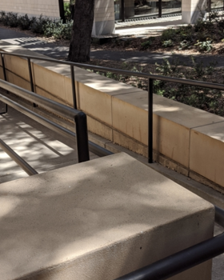} &
\includegraphics[width=\sotaresultswidth]{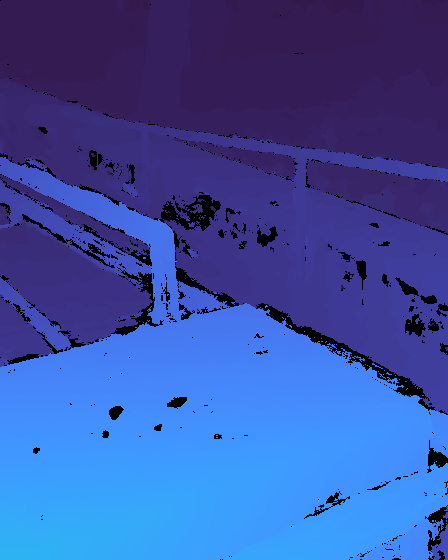} &
\includegraphics[width=\sotaresultswidth]{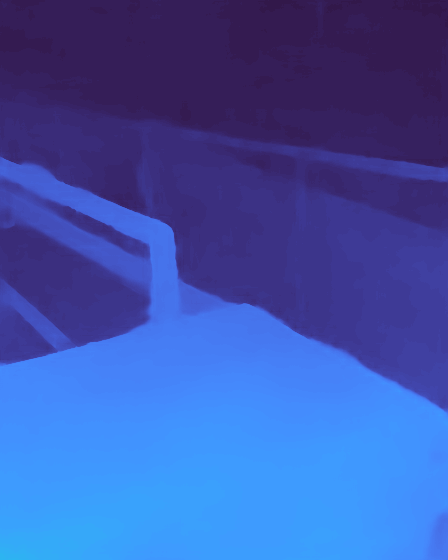} &
\includegraphics[width=\sotaresultswidth]{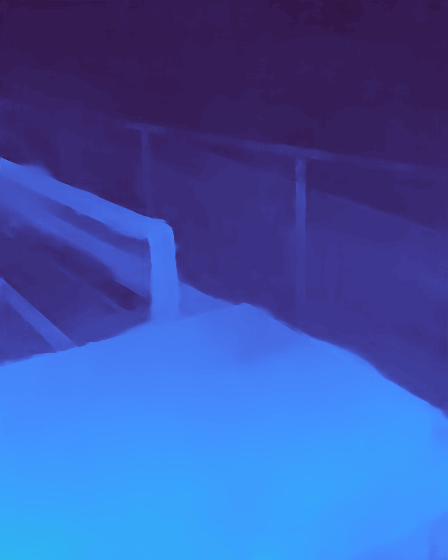} &
\includegraphics[width=\sotaresultswidth]{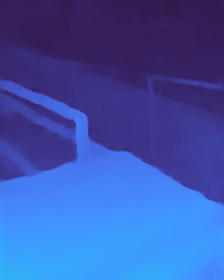} &
\includegraphics[width=\sotaresultswidth]{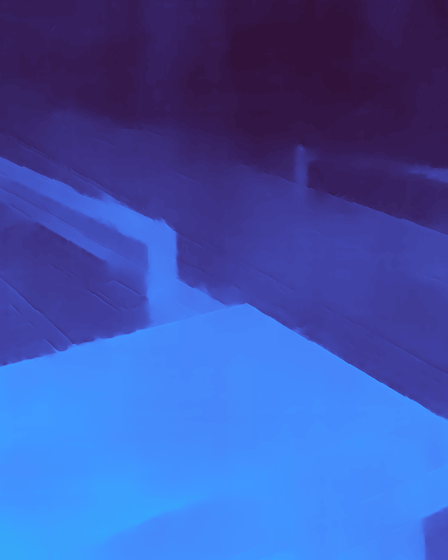}
\\
\includegraphics[width=\sotaresultswidth]{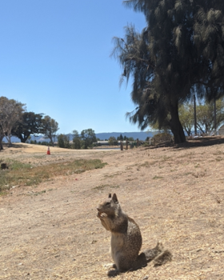} &
\includegraphics[width=\sotaresultswidth]{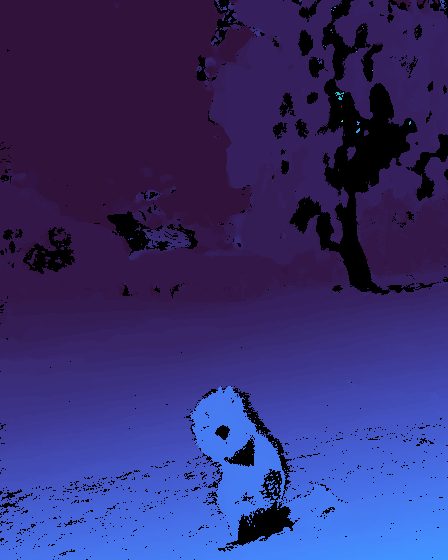} &
\includegraphics[width=\sotaresultswidth]{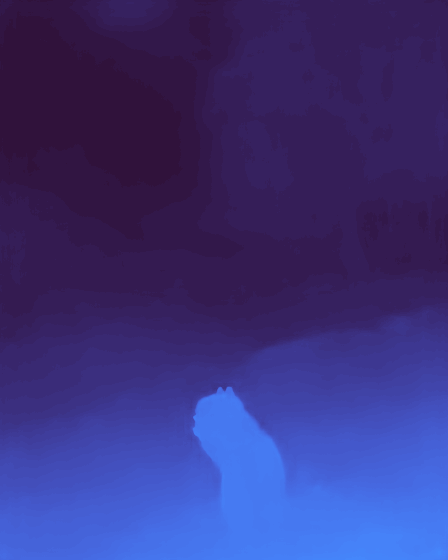} &
\includegraphics[width=\sotaresultswidth]{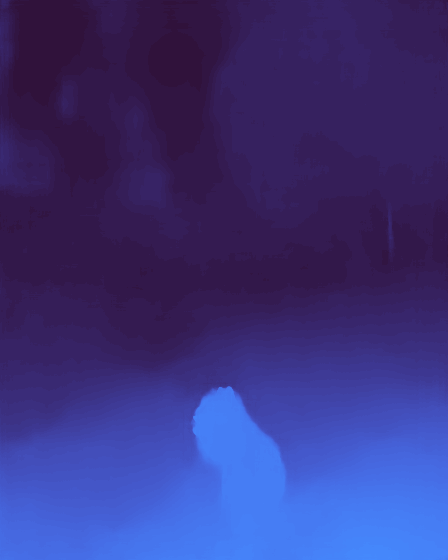} &
\includegraphics[width=\sotaresultswidth]{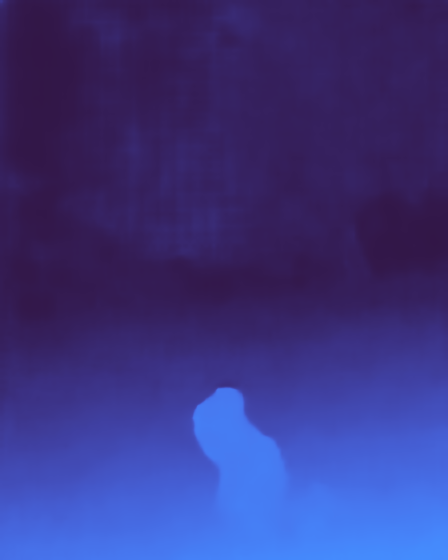} &
\includegraphics[width=\sotaresultswidth]{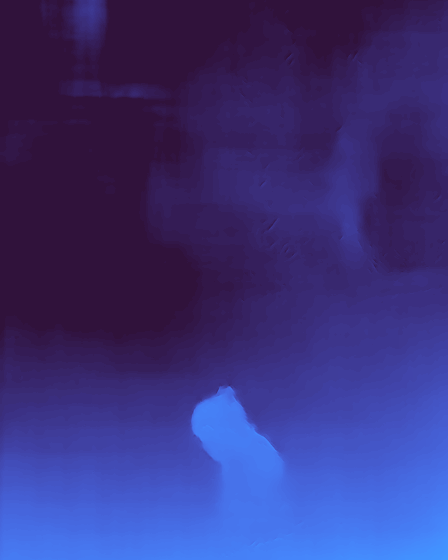}
\\
\includegraphics[width=\sotaresultswidth]{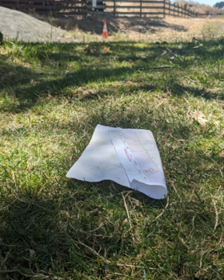} &
\includegraphics[width=\sotaresultswidth]{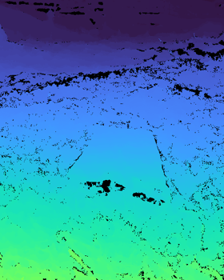} &
\includegraphics[width=\sotaresultswidth]{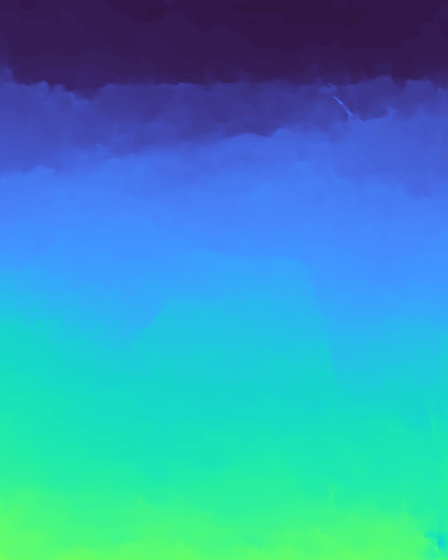} &
\includegraphics[width=\sotaresultswidth]{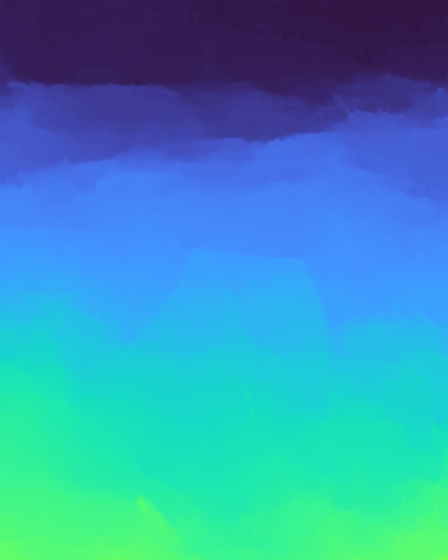} &
\includegraphics[width=\sotaresultswidth]{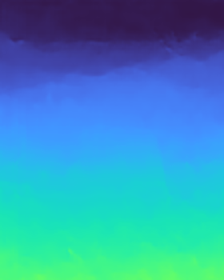} &
\includegraphics[width=\sotaresultswidth]{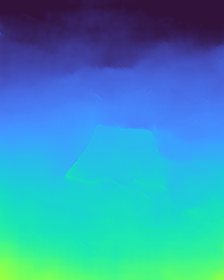}
\\
\includegraphics[width=\sotaresultswidth]{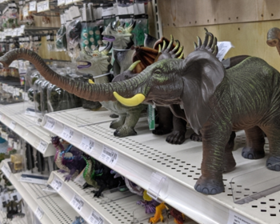} &
\includegraphics[width=\sotaresultswidth]{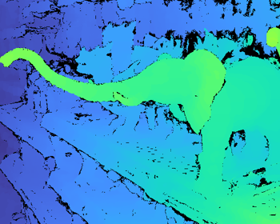} &
\includegraphics[width=\sotaresultswidth]{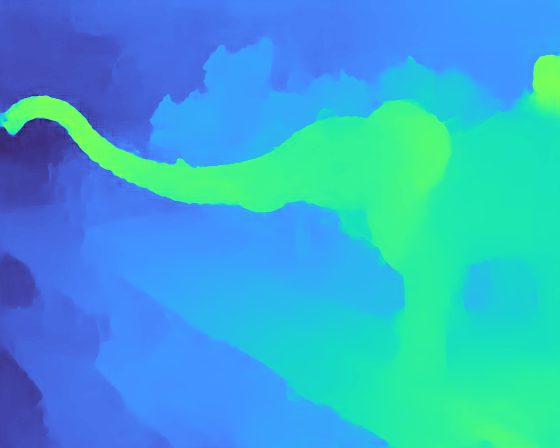} &
\includegraphics[width=\sotaresultswidth]{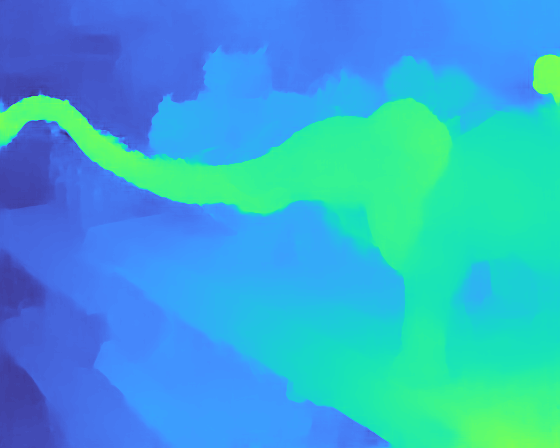} &
\includegraphics[width=\sotaresultswidth]{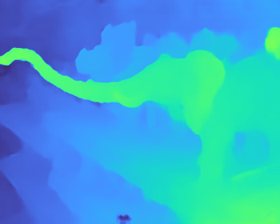} &
\includegraphics[width=\sotaresultswidth]{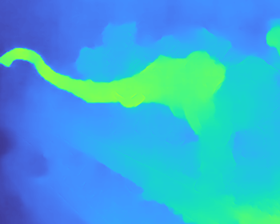}
\\
\includegraphics[width=\sotaresultswidth]{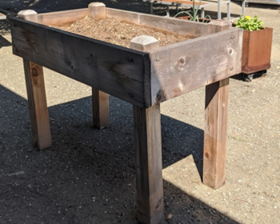} &
\includegraphics[width=\sotaresultswidth]{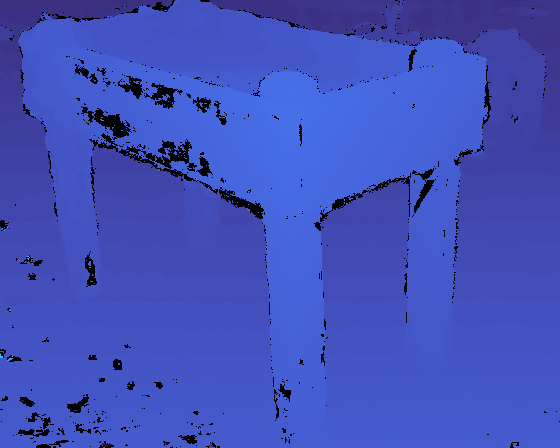} &
\includegraphics[width=\sotaresultswidth]{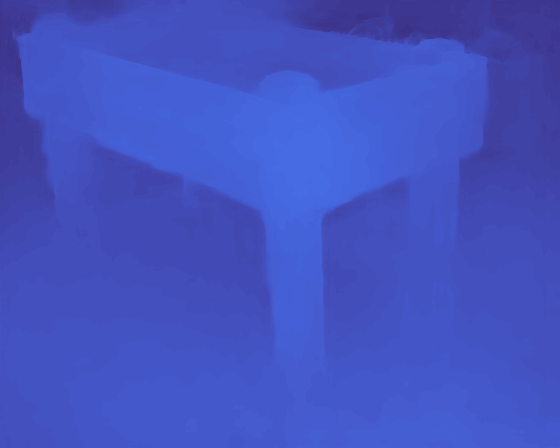} &
\includegraphics[width=\sotaresultswidth]{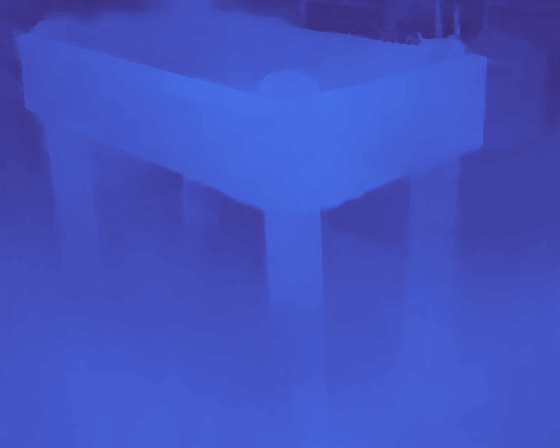} &
\includegraphics[width=\sotaresultswidth]{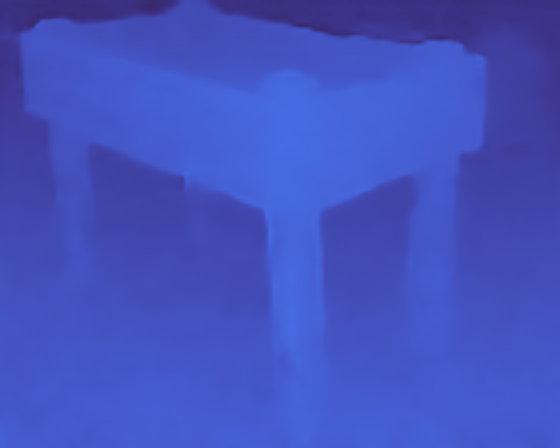} &
\includegraphics[width=\sotaresultswidth]{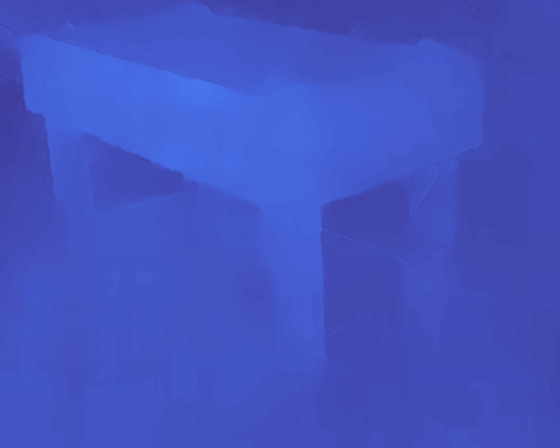}
\\
    \fakecaption{(a) \,Image} & \fakecaption{(b)\,GT} & \fakecaption{(c)\,Ours} & \fakecaption{(d)\,StereoNet} & \fakecaption{(e)\,PSMNet} & \fakecaption{(f)\,DPNet}\\
    \end{tabular}
    \caption{DPNet (f) performs worse in distant areas compared to methods that take DC as input ((c), (d), (e)) due to the small baseline between the two dual-pixel images.
    }
    \label{fig:gallery_dp_worse}
    \vspace{-10pt}
\end{figure}

\subsection{Analysis of Best and Worst Cases}
In Fig.~\ref{fig:failure} we show representative images from the best (top 3 rows) and the worst (bottom 3 rows) results for our method as ranked by the $\operatorname{MAE}$ metric. As expected, the method performs very accurately on images with high frequency details and textured scenes whereas it does worse (along with other methods) in textureless areas.

\newcommand{\goodbadresultswidth}{0.16\textwidth}{}
\begin{figure}[hp]
    \centering
    \begin{tabular}{@{}c@{\,\,}c@{\,\,}c@{\,\,}c@{\,\,}c@{\,\,}c@{}}
\includegraphics[width=\goodbadresultswidth]{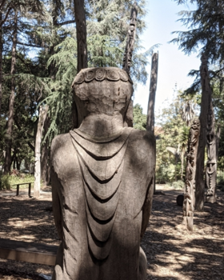} &
\includegraphics[width=\goodbadresultswidth]{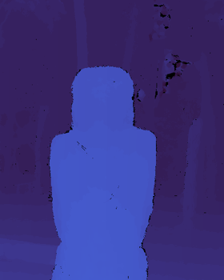} &
\includegraphics[width=\goodbadresultswidth]{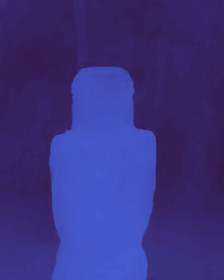} &
\includegraphics[width=\goodbadresultswidth]{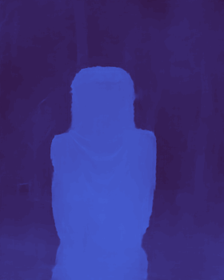} &
\includegraphics[width=\goodbadresultswidth]{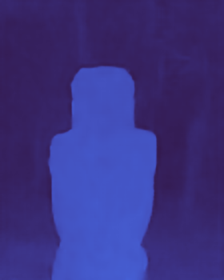} &
\includegraphics[width=\goodbadresultswidth]{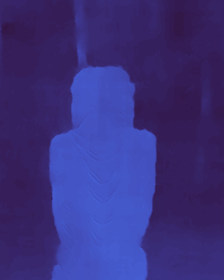}
\\
\includegraphics[width=\goodbadresultswidth]{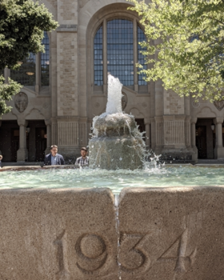} &
\includegraphics[width=\goodbadresultswidth]{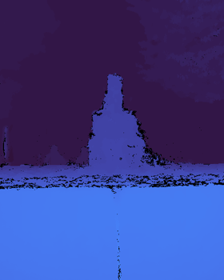} &
\includegraphics[width=\goodbadresultswidth]{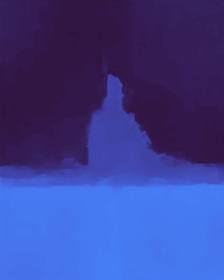} &
\includegraphics[width=\goodbadresultswidth]{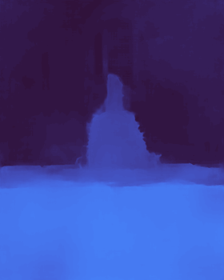} &
\includegraphics[width=\goodbadresultswidth]{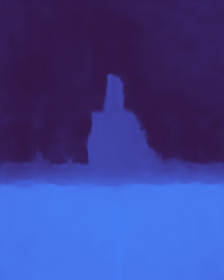} &
\includegraphics[width=\goodbadresultswidth]{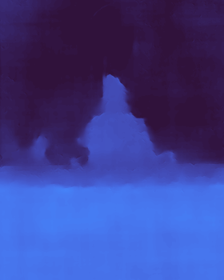}
\\
\includegraphics[width=\goodbadresultswidth]{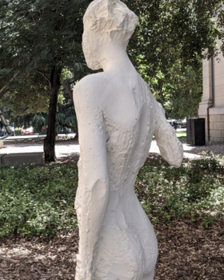} &
\includegraphics[width=\goodbadresultswidth]{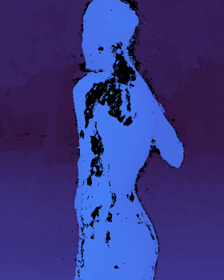} &
\includegraphics[width=\goodbadresultswidth]{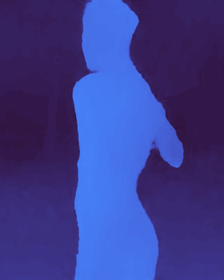} &
\includegraphics[width=\goodbadresultswidth]{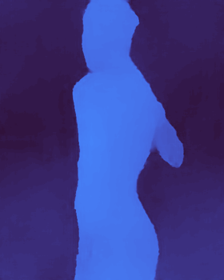} &
\includegraphics[width=\goodbadresultswidth]{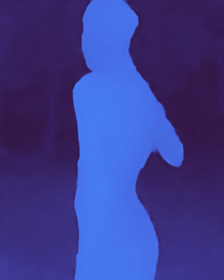} &
\includegraphics[width=\goodbadresultswidth]{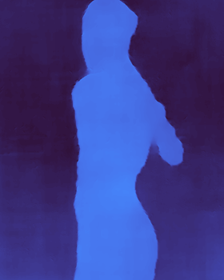}
\\ 
&&&&&\\
\includegraphics[width=\goodbadresultswidth]{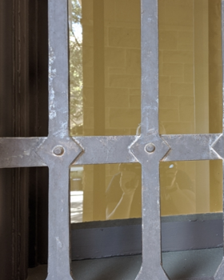} &
\includegraphics[width=\goodbadresultswidth]{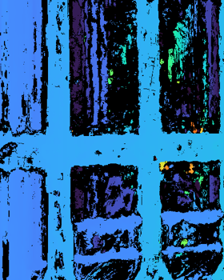} &
\includegraphics[width=\goodbadresultswidth]{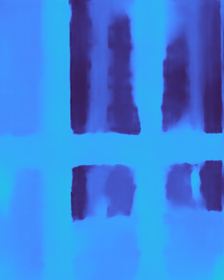} &
\includegraphics[width=\goodbadresultswidth]{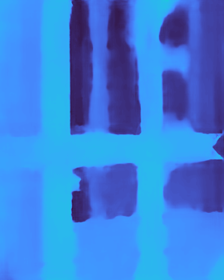} &
\includegraphics[width=\goodbadresultswidth]{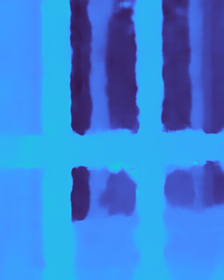} &
\includegraphics[width=\goodbadresultswidth]{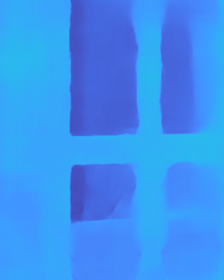}
\\
\includegraphics[width=\goodbadresultswidth]{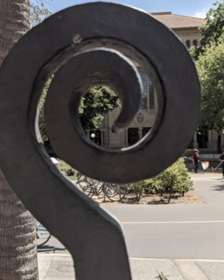} &
\includegraphics[width=\goodbadresultswidth]{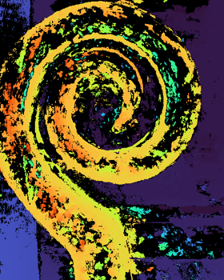} &
\includegraphics[width=\goodbadresultswidth]{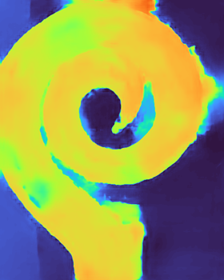} &
\includegraphics[width=\goodbadresultswidth]{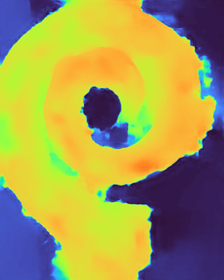} &
\includegraphics[width=\goodbadresultswidth]{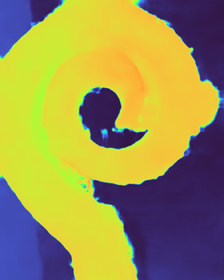} &
\includegraphics[width=\goodbadresultswidth]{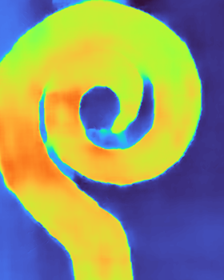}
\\
\includegraphics[width=\goodbadresultswidth]{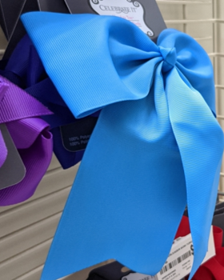} &
\includegraphics[width=\goodbadresultswidth]{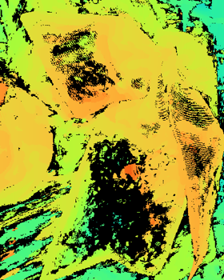} &
\includegraphics[width=\goodbadresultswidth]{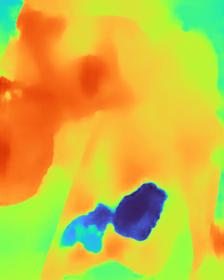} &
\includegraphics[width=\goodbadresultswidth]{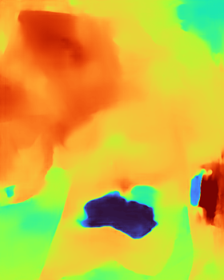} &
\includegraphics[width=\goodbadresultswidth]{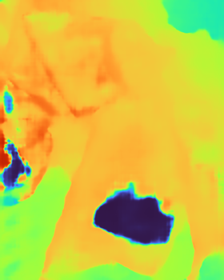} &
\includegraphics[width=\goodbadresultswidth]{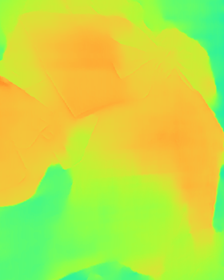}
\\
    \fakecaption{(a) \,Image} & \fakecaption{(b)\,GT} & \fakecaption{(c)\,Ours} & \fakecaption{(d)\,StereoNet} & \fakecaption{(e)\,PSMNet} & \fakecaption{(f)\,DPNet}\\
    \end{tabular}
    \caption{Representative images from our best (top 3 rows) and worst (bottom 3 rows) results, as rated by MAE metric.}
    \label{fig:failure}
    \vspace{-15pt}
\end{figure}

\subsection{More Results for Applications}
We show more examples of computational photography applications.
Fig. \ref{fig:supp_bokeh} shows results of synthetic shallow depth-of-field effect using disparity from different models. Our method produces better details for object boundary and thin structure, which  prevents artifacts near the subject boundary. 

\newcommand{\bokehresultswidth}{0.19\textwidth}{}
\begin{figure}[tp]
    \centering
    \begin{tabular}{@{}c@{\,\,}c@{\,\,}c@{\,\,}c@{\,\,}c@{}}
\includegraphics[width=\bokehresultswidth]{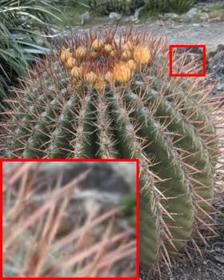} &
\includegraphics[width=\bokehresultswidth]{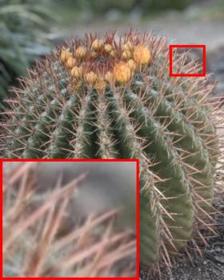} &
\includegraphics[width=\bokehresultswidth]{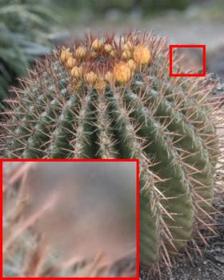} &
\includegraphics[width=\bokehresultswidth]{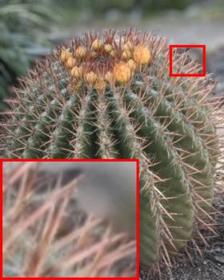} &
\includegraphics[width=\bokehresultswidth]{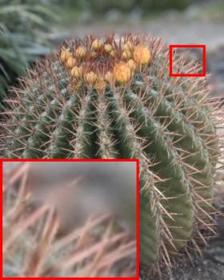}
\\
\includegraphics[width=\bokehresultswidth]{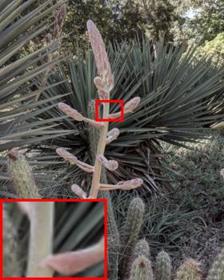} &
\includegraphics[width=\bokehresultswidth]{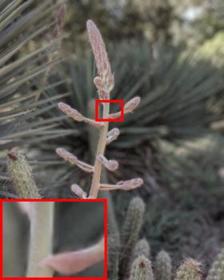} &
\includegraphics[width=\bokehresultswidth]{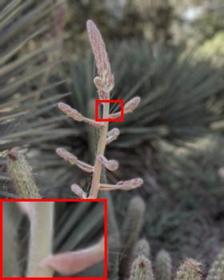} &
\includegraphics[width=\bokehresultswidth]{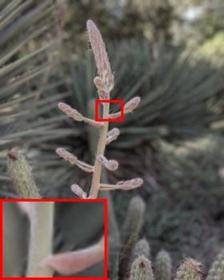} &
\includegraphics[width=\bokehresultswidth]{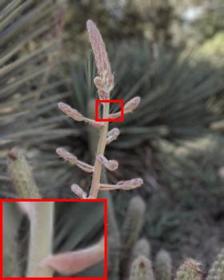}
\\
\includegraphics[width=\bokehresultswidth]{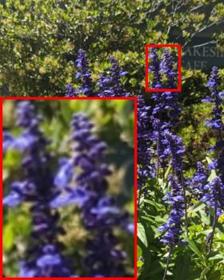} &
\includegraphics[width=\bokehresultswidth]{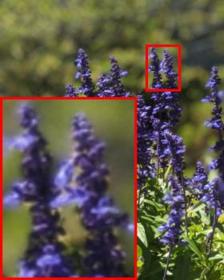} &
\includegraphics[width=\bokehresultswidth]{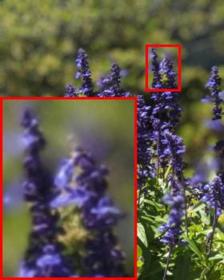} &
\includegraphics[width=\bokehresultswidth]{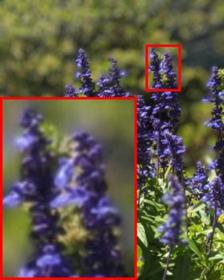} &
\includegraphics[width=\bokehresultswidth]{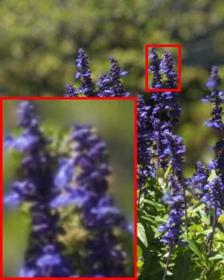}
\\
\includegraphics[width=\bokehresultswidth]{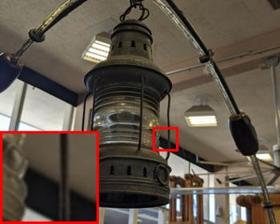} &
\includegraphics[width=\bokehresultswidth]{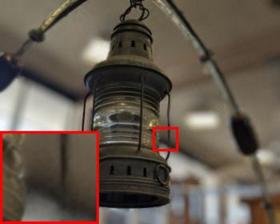} &
\includegraphics[width=\bokehresultswidth]{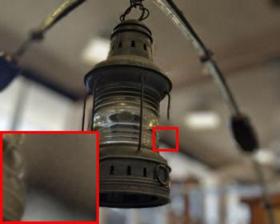} &
\includegraphics[width=\bokehresultswidth]{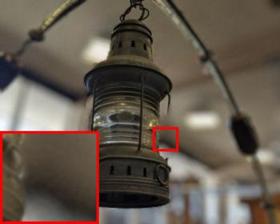} &
\includegraphics[width=\bokehresultswidth]{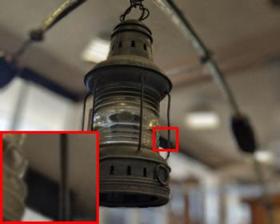}
\\
\includegraphics[width=\bokehresultswidth]{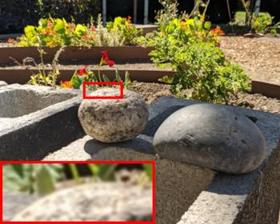} &
\includegraphics[width=\bokehresultswidth]{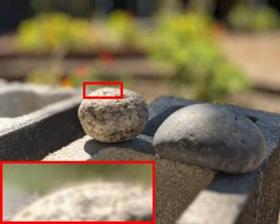} &
\includegraphics[width=\bokehresultswidth]{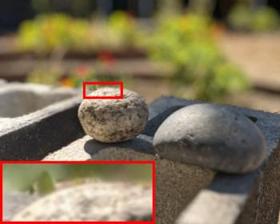} &
\includegraphics[width=\bokehresultswidth]{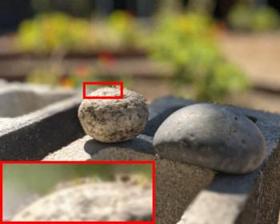} &
\includegraphics[width=\bokehresultswidth]{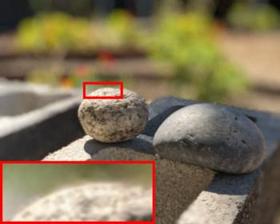}
\\
     \fakecaption{(a) \,Image} & \fakecaption{(b)\,Ours} & \fakecaption{(c)\,StereoNet} & \fakecaption{(d)\,PSMNet} & \fakecaption{(e)\,DPNet}\\
    \end{tabular}
    \caption{Synthetic shallow depth-of-field results for different methods. Accurate depth near occlusion boundaries is critical for avoiding artifacts near the subject boundary.}
    \label{fig:supp_bokeh}
    \vspace{-10pt}
\end{figure}

\end{document}